\documentclass{article}

\usepackage[preprint]{neurips_2019}

\usepackage[utf8]{inputenc} 
\usepackage[T1]{fontenc}    
\usepackage{hyperref}       
\usepackage{url}            
\usepackage{booktabs}       
\usepackage{amsfonts}       
\usepackage{nicefrac}       
\usepackage{microtype}      
\usepackage{subcaption}
\usepackage{amssymb}
\usepackage{wrapfig}
\usepackage{pifont}
\usepackage{graphicx}
\usepackage[dvipsnames]{xcolor}
\usepackage{algorithm, algpseudocode}
\usepackage{amsmath}
\newcommand{\distas}[1]{\mathbin{\overset{#1}{\kern\z@\sim}}}%
\newsavebox{\mybox}\newsavebox{\mysim}

\floatname{algorithm}{Algorithm}

\algtext*{EndFunction}

\usepackage{arydshln}
\usepackage{multirow}
\usepackage[belowskip=-0pt]{caption}

\definecolor{mydarkblue}{rgb}{0,0.08,0.45}
\hypersetup{
    colorlinks=true,
    linkcolor=mydarkblue,
    citecolor=mydarkblue,
    filecolor=mydarkblue,
    urlcolor=mydarkblue}

\title{Latent ODEs for Irregularly-Sampled Time Series}
\author{Yulia Rubanova,\ \  Ricky T. Q. Chen, \ \ David Duvenaud \\
University of Toronto and the Vector Institute \\
\texttt{\{rubanova,rtqichen,duvenaud\}@cs.toronto.edu}}

\begin{document}
\maketitle
\begin{abstract}
Time series with non-uniform intervals occur in many applications, and are difficult to model using standard recurrent neural networks (RNNs).
We generalize RNNs to have continuous-time hidden dynamics defined by ordinary differential equations (ODEs), a model we call ODE-RNNs.
Furthermore, we use ODE-RNNs to replace the recognition network of the recently-proposed Latent ODE model.
Both ODE-RNNs and Latent ODEs can naturally handle arbitrary time gaps between observations, and can explicitly model the probability of observation times using Poisson processes.
We show experimentally that these ODE-based models outperform their RNN-based counterparts on irregularly-sampled data.
%
%
\end{abstract}

\section{Introduction}

\begin{wrapfigure}[29]{r}{0.41\textwidth}
\vspace{-6mm}%
\hspace{-2mm}%
\begin{tabular}{c}
Standard RNN \vspace{-0.5mm} \\ \includegraphics[width=\linewidth, clip, trim=0mm 0mm 0mm 4mm]{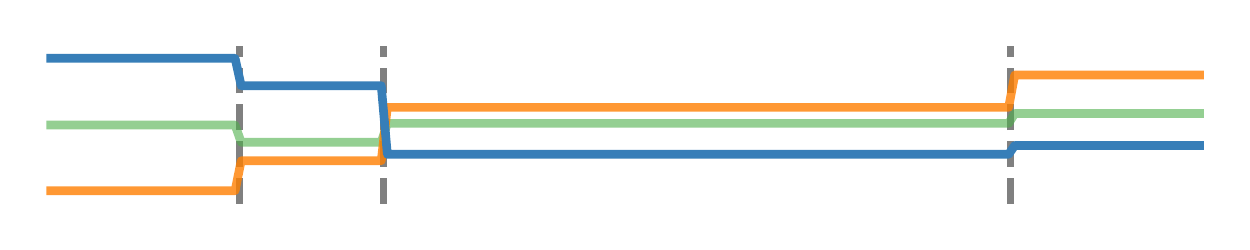}\\
RNN-Decay    \vspace{-0.4mm} \\ \includegraphics[width=\linewidth, clip, trim=0mm 0mm 0mm 4mm]{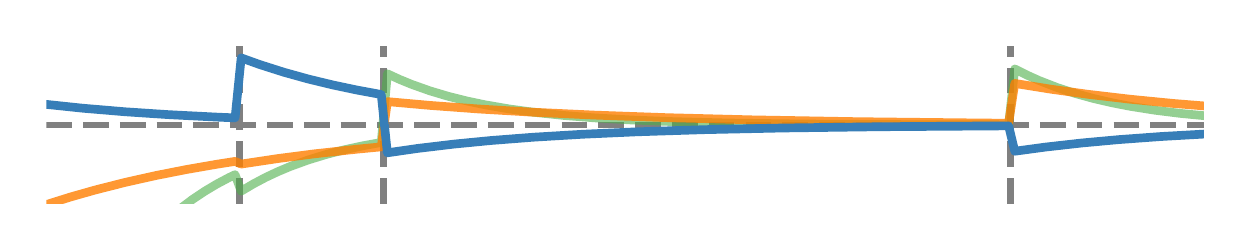} \\
Neural ODE          \vspace{-0.5mm} \\ \includegraphics[width=\linewidth, clip, trim=0mm 0mm 0mm 0mm]{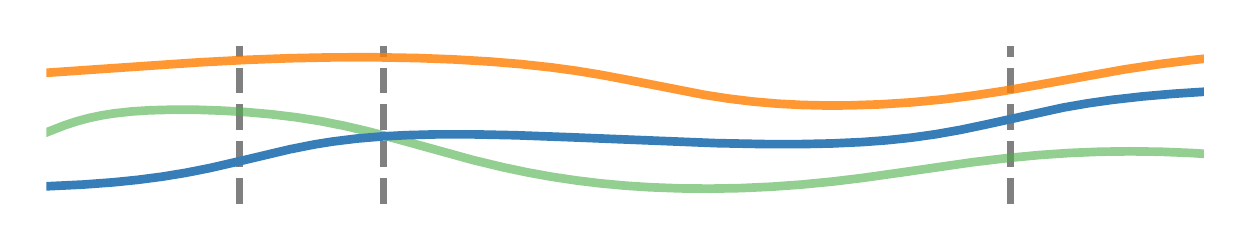} \\
ODE-RNN      \vspace{-0.5mm} \\ \includegraphics[width=\linewidth, clip, trim=0mm 4mm 0mm 4mm]{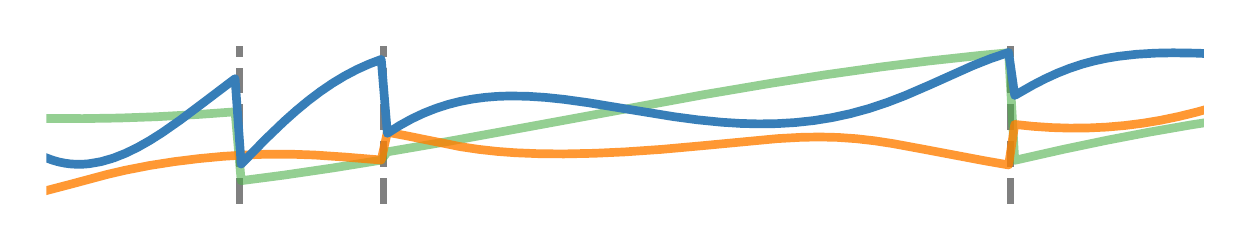}\\ 
{\small Time}
\end{tabular}
\caption{Hidden state trajectories.
Vertical lines show observation times. Lines show different dimensions of the hidden state.
Standard RNNs have constant or undefined hidden states between observations. 
The RNN-Decay model has states which exponentially decay towards zero, and are updated at observations.
States of Neural ODE follow a complex trajectory but are determined by the initial state.
The ODE-RNN model has states which obey an ODE between observations, and are also updated at observations.
}
\label{fig:fig1}
\end{wrapfigure}

Recurrent neural networks (RNNs) are the dominant model class for high-dimensional, regularly-sampled time series data, such as text or speech.
However, they are an awkward fit for irregularly-sampled time series data, common in medical or business settings. 
A standard trick for applying RNNs to irregular time series is to divide the timeline into equally-sized intervals, and impute or aggregate observations using averages.
Such preprocessing destroys information, particularly about the timing of measurements, which can be informative about latent variables~\citep{pmlr_Lipton16, che_sontag_2018}.

An approach which better matches reality is to construct a continuous-time model with a latent state defined at all times.
Recently, steps have been taken in this direction, defining RNNs with continuous dynamics given by a simple exponential decay between observations~\citep{che_sontag_2018, BRITS_2018, google_ehr_2018, neural_hawkes}.

We generalize state transitions in RNNs to continuous-time dynamics specified by a neural network, as in Neural ODEs~\citep{NeuralODE}.
We call this model the ODE-RNN, and use it to contruct two distinct continuous-time models.
First, we use it as a standalone autoregressive model.
Second, we refine the Latent ODE model of \citet{NeuralODE} by using the ODE-RNN as a recognition network.
Latent ODEs define a generative process over time series based on the deterministic evolution of an initial latent state, and can be trained as a variational autoencoder~\citep{kingma2013auto}.
Both models naturally handle time gaps between observations, and remove the need to group observations into equally-timed bins. 
We compare ODE models to several RNN variants and find that ODE-RNNs can perform better when the data is sparse.
Since the absence of observations itself can be informative, we further augment Latent ODEs to jointly model times of observations using a Poisson process.

\section{Background}
\paragraph{Recurrent neural networks}
A simple way to handle irregularly-timed samples is to include the time gap between observations $\Delta_t = t_i - t_{i-1}$ into the update function of the RNN:
\begin{align}\label{eq:rnn_delta_t}
    h_i = \textnormal{RNNCell}(h_{i-1}, \Delta_t, x_i)
\end{align}
However, this approach raises the question of how to define the hidden state $h$ between observations.
A simple alternative introduces an exponential decay of the hidden state towards zero when no observations are made~\citep{che_sontag_2018, BRITS_2018, google_ehr_2018, mozer_2017}:
\begin{align}\label{eq:exp_decay}
    h_i = \textnormal{RNNCell}(h_{i-1} \cdot \exp\{-\tau \Delta_t \}, x_i)
\end{align}
where $\tau$ is a decay rate parameter.
However, \citet{mozer_2017} found that empirically, exponential-decay dynamics did not improve predictive performance over standard RNN approaches.

\paragraph{Neural Ordinary Differential Equations}Neural ODEs~\citep{NeuralODE} are a family of continuous-time models which define a hidden state $h(t)$ to be the solution to an ODE initial-value problem (IVP):
\begin{align}
    \frac{d h(t)}{dt} = f_\theta(h(t), t) \quad \textnormal{where} \quad h(t_0) = h_0
\end{align}
in which the function $f_\theta$ specifies the dynamics of the hidden state, using a neural network with parameters $\theta$. 
The hidden state $h(t)$ is defined at all times, and can be evaluated at any desired times using a numerical ODE solver:
\begin{align}\label{eq:latent_ode}
h_0,\dots,h_N = \text{ODESolve}(f_\theta, h_0, (t_0,\dots,t_N))
\end{align}
\citet{NeuralODE} showed how adjoint sensitivities can be used to compute memory-efficient gradients w.r.t.\ $\theta$, allowing black-box ODE solvers to be used as a building block in larger models.
They also conducted toy experiments in a time-series model in which the latent state follows a Neural ODE.
\citet{NeuralODE} used time-invariant dynamics in their time-series model: $\nicefrac{d h(t)}{dt} = f_\theta(h(t))$
, and we follow the same approach, but adding time-dependence would be straightforward if necessary.




\section{Method}

In this section, we use neural ODEs to define two distinct families of continuous-time models: the autoregressive ODE-RNN, and the variational-autoencoder-based Latent ODE.

\subsection{Constructing an ODE-RNN Hybrid}

Following \citet{mozer_2017}, we note that an RNN with exponentially-decayed hidden state implicitly obeys the following ODE $\frac{dh(t)}{dt} = -\tau h$ with $h(t_0) = h_0$,
where $\tau$ is a parameter of the model.
The solution to this ODE is the pre-update term $h_0 \cdot \exp\{-\tau \Delta_t\}$ in \eqref{eq:exp_decay}.
This differential equation is time-invariant, and assumes that the stationary point (i.e.\ zero-valued state) is special.
We can generalize this approach and model the hidden state using a Neural ODE.
The resulting algorithm is given in Algorithm~\ref{alg:ode_gru}.
We define the state between observations to be the solution to an ODE: $h_{i}' = \textnormal{ODESolve}(f_\theta, h_{i-1}, (t_{i-1}, t_i))$ and then at each observation, update the hidden state using a standard RNN update $h_{i} = \textnormal{RNNCell}(h_{i}', x_{i})$.
Our model does not explicitly depend on $t$ or $\Delta_t$ when updating the hidden state, but does depend on time implicitly through the resulting dynamical system.
Compared to RNNs with exponential decay, our approach allows a more flexible parameterization of the dynamics.
A comparison between the state dynamics of these models is given in table~\ref{tab:state_dynamics}.



\begin{algorithm}
	\begin{algorithmic}
	\State {\bfseries Input:} Data points and their timestamps $\{(x_i, t_{i})\}_{i=1..N}$
	\State $h_0$ = \textbf{0}
	\For{i in 1, 2, \dots, N}
    \State \textcolor{blue}{$h_{i}' = \textnormal{ODESolve}(f_\theta, h_{i-1}, (t_{i-1}, t_i))$} \Comment{Solve ODE to get state at $t_i$}
	\State $h_{i} = \textnormal{RNNCell}(\textcolor{blue}{h_{i}'}, x_{i})$ \Comment{Update hidden state given current observation $x_i$}
	\EndFor
	\State $o_{i} = \textnormal{OutputNN}(h_{i})$ for all $i=1..N$
	\State {\bfseries Return:} $\{o_i\}_{i=1..N}; h_{N}$
	\end{algorithmic}
	\caption{The ODE-RNN.  The only difference, highlighted in \textcolor{blue}{blue}, from standard RNNs is that the pre-activations $h'$ evolve according to an ODE between observations, instead of being fixed.}
	\label{alg:ode_gru}
\end{algorithm}

\paragraph{Autoregressive Modeling with the ODE-RNN}
The ODE-RNN can straightforwardly be used to probabilistically model sequences.
Consider a series of observations $\{x_i\}_{i = 0}^N$ at times $\{t_i\}_{i = 0}^N$. 
Autoregressive models make a one-step-ahead prediction conditioned on the history of observations, i.e.\ they factor the joint density $p(x) = \prod_i p_\theta(x_i|x_{i-1}, \dots, x_0)$.
As in standard RNNs, we can use an ODE-RNN to specify the conditional distributions $p_\theta(x_i|x_{i-1}...x_0)$ (Algorithm~\ref{alg:ode_gru}).

\subsection{Latent ODEs: a Latent-variable Construction}


Autoregressive models such as RNNs and the ODE-RNN presented above are easy to train and allow fast online predictions.
However, autoregressive models can be hard to interpret, since their update function combines both their model of system dynamics, and of conditioning on new observations.
Furthermore, their hidden state does not explicitly encode uncertainty about the state of the true system.
In terms of predictive accuracy, autoregressive models are often sufficient for densely sampled data, but perform worse when observations are sparse.

An alternative to autoregressive models are latent-variable models.
For example, \citet{NeuralODE} proposed a latent-variable time series model, where the generative model is defined by ODE whose initial latent state $z_0$ determines the entire trajectory:
%
\begin{align}
 z_0 & \sim p(z_0) \\
 z_0, z_1, \dots,z_N & = \text{ODESolve}(f_\theta, z_0, (t_0, t_1, \dots,t_N)) \\
 \textnormal{each} \quad {x}_{i} & \stackrel{\footnotesize{indep.}}{\sim} p(x_i | z_{i}) \quad \  i=0, 1, \dots, N
 \label{eq:gen_ode}
\end{align}
%

%




\begin{wraptable}[7]{r}{0.49\textwidth}
\centering
\small
\vspace{-0.5cm}
\begin{tabular}{@{}lrr@{}}\toprule
    \textbf{\small Encoder-decoder models} & Encoder & Decoder \\ \midrule
    Latent ODE (ODE enc.) & ODE-RNN & ODE  \\
    Latent ODE (RNN enc.) & RNN & ODE \\
    RNN-VAE & RNN & RNN\\
    \bottomrule
\end{tabular}
\vspace{0.4mm}
\caption{Different encoder-decoder architectures.}
\label{tab:model_comparison}
\end{wraptable}

We follow \citet{NeuralODE} in using a variational autoencoder framework for both training and prediction.
This requires estimating the approximate posterior $q(z_0| \{x_i,t_i\}_{i=0}^N)$.
Inference and prediction in this model is effectively an encoder-decoder or sequence-to-sequence architecture, in which a variable-length sequence is encoded into a fixed-dimensional embedding, which is then decoded into another variable-length sequence, as in \citet{sutskever2014sequence}. 

\begin{table}[b]
\centering
\vspace{-4mm}
\begin{minipage}[c]{0.48\linewidth}
    \caption{Definition of hidden state $h(t)$ between observation times $t_{i-1}$ and $t_i$ in autoregressive models.
    In standard~RNNs, the hidden state does not change between updates.
    In ODE-RNNs, the hidden state is defined by an ODE, and is additionally updated by another network at each observation. 
    } \label{tab:state_dynamics}
 \end{minipage}%
 \hfill%
 \begin{minipage}[c]{0.48\linewidth}
 \resizebox{\linewidth}{!}{%
   	\begin{tabular}{@{}lr@{}}\toprule
    \textbf{Model} & State $h(t_i)$ between observations\\ \midrule
    Standard RNN &  $h_{t_{i-1}}$ \\
    RNN-Decay & $h_{t_{i-1}} e^{-\tau \Delta_t}$\\
    GRU-D & $h_{t_{i-1}} e^{-\tau \Delta_t}$ \\
    ODE-RNN & $\text{ODESolve}(f_\theta, {h_{i-1}}, ({t_{i-1}}, t))$ \\
    \bottomrule
    \end{tabular}}
 \end{minipage}
\end{table}

\citet{NeuralODE} used an RNN as a recognition network to compute this approximate posterior.
We conjecture that using an ODE-RNN as defined above for the recognition network would be a more effective parameterization when the datapoints are irregularly sampled.
Thus, we propose using an ODE-RNN as the encoder for a latent ODE model, resulting in a fully ODE-based sequence-to-sequence model.
In our approach, the mean and standard deviation of the approximate posterior $q(z_0| \{x_i,t_i\}_{i=0}^N)$ are a function of the final hidden state of an ODE-RNN:
%
\begin{align}
q(z_0| \{x_i,t_i\}_{i=0}^N) = \mathcal{N} (\mu_{z_0}, \sigma_{z_0}) \quad \textnormal{where} \quad \mu_{z_0}, \sigma_{z_0} = g( \text{ODE-RNN}_{\phi}(\{x_i,t_i\}_{i=0}^N))
\end{align}
Where $g$ is a neural network translating the final hidden state of the ODE-RNN encoder into the mean and variance of $z_0$.
To get the approximate posterior at time point $t_0$, we run the ODE-RNN encoder backwards-in-time from $t_N$ to $t_0$.
We jointly train both the encoder and decoder by maximizing the evidence lower bound (ELBO):
\begin{align}
    \textnormal{ELBO}(\theta, \phi) = \mathbb{E}_{z_0 \sim q_\phi(z_0| \{x_i,t_i\}_{i=0}^N)} \left[ \log p_\theta(x_{0},\dots,x_{N})) \right] -\textnormal{KL}\left[q_\phi(z_0| \{x_i,t_i\}_{i=0}^N) || p(z_0) \right]
    \label{eq:elbo}
\end{align}
%
%
%
%
%
\begin{figure}
	\centering
	\includegraphics[width=0.85\linewidth]{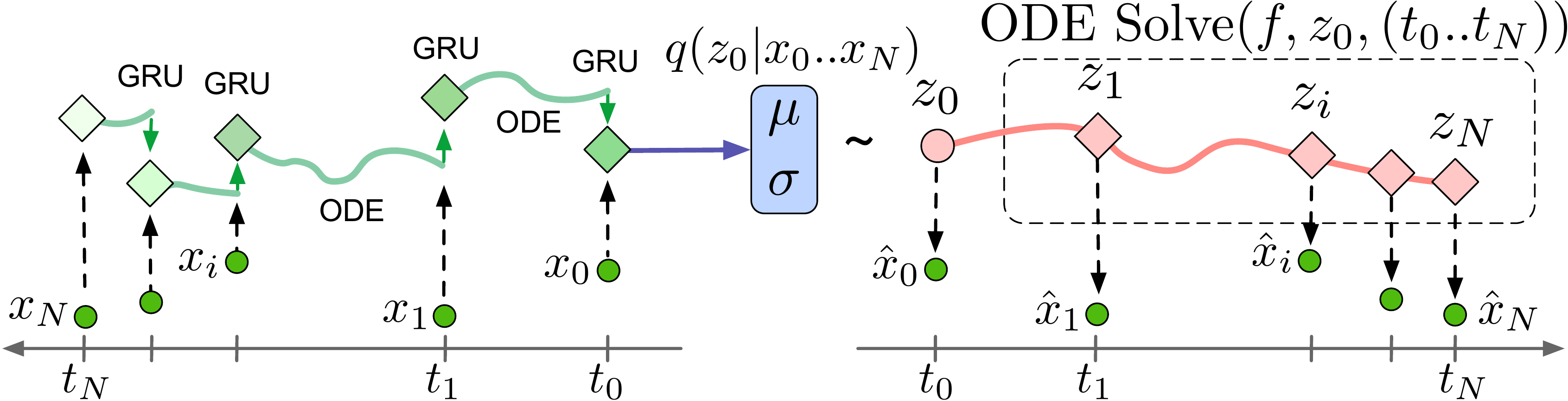}
	\caption{The Latent ODE model with an ODE-RNN encoder.
	To make predictions in this model, the ODE-RNN encoder is run backwards in time to produce an approximate posterior over the initial state: $q(z_0| \{x_i,t_i\}_{i=0}^N)$.
	Given a sample of $z_0$, we can find the latent state at any point of interest by solving an ODE initial-value problem.
	Figure adapted from \citet{NeuralODE}.}
	\label{fig:latent-ode}
	\vspace{-3mm}
\end{figure}
%
%
%

This latent variable framework comes with several benefits:
First, it explicitly decouples the dynamics of the system (ODE), the likelihood of observations, and the recognition model, allowing each to be examined or specified on its own.
Second, the posterior distribution over latent states provides an explicit measure of uncertainty, which is not available in standard RNNs and ODE-RNNs.
Finally, it becomes easier to answer non-standard queries, such as making predictions backwards in time, or conditioning on a subset of observations.

\subsection{Poisson process likelihoods}

\begin{wrapfigure}[22]{r}{0.30\textwidth}
    \vspace{-6mm}
	\centering
	{\footnotesize Diastolic arterial blood pressure}
    \includegraphics[width=0.99\linewidth, clip, trim=0 0 0 15]{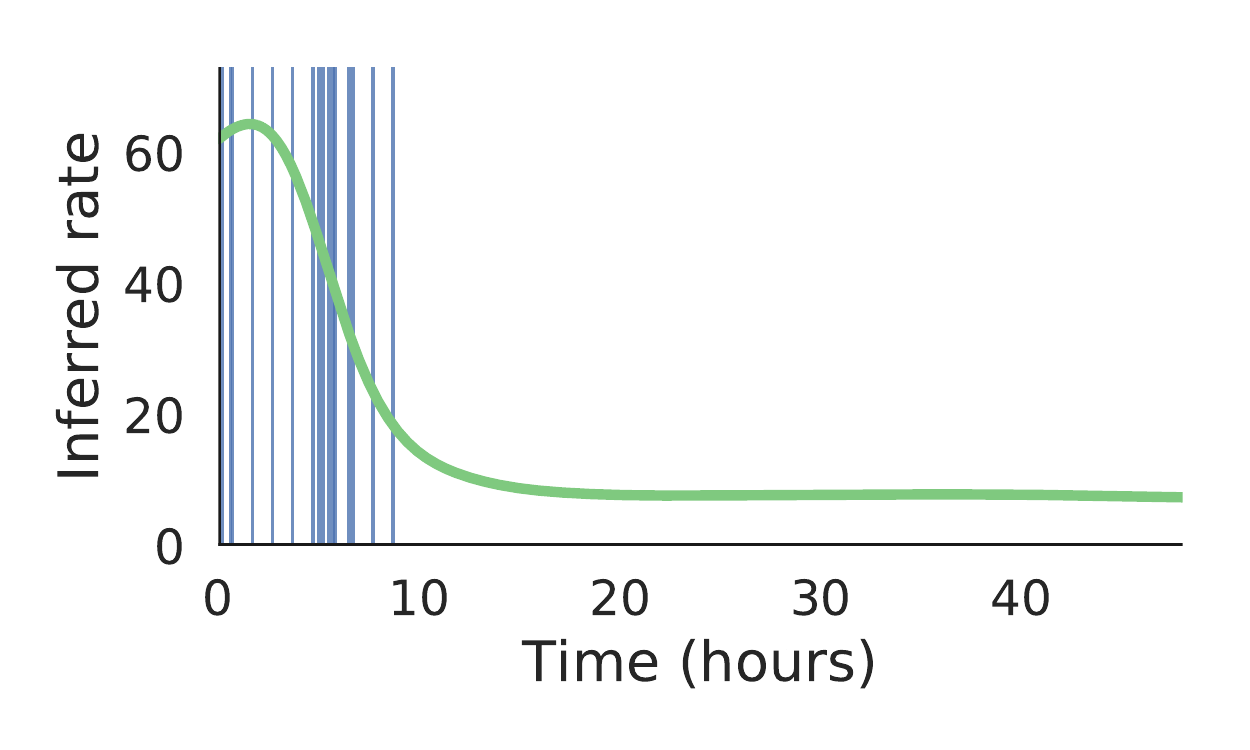}
    {\footnotesize Partial pressure of arterial O2}
     \includegraphics[width=0.99\linewidth, clip, trim=0 0 0 15]{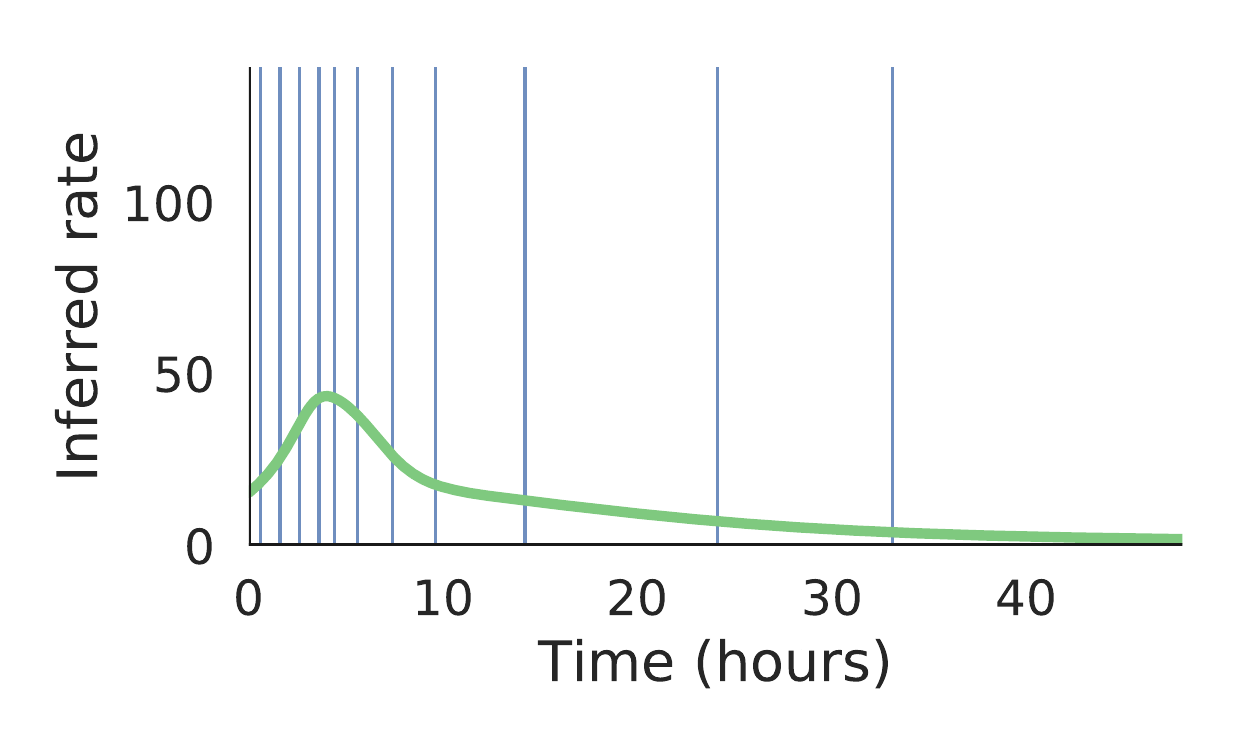}
    \caption{Visualization of the inferred Poisson rate $\lambda(t)$ (green line) for two selected features of different patients from the Physionet dataset. Vertical lines mark observation times.} 
    \label{fig:poisson}
\end{wrapfigure}
The fact that a measurement was made at a particular time is often informative about the state of the system~\citep{che_sontag_2018}.
In the ODE framework, we can use the continuous latent state to parameterize the intensity of events using an\emph{inhomogeneous} Poisson point process \citep{palm1943intensitatsschwankungen} where the event rate $\lambda(t)$ changes over time.
Poisson point processes have the following log-likelihood:
\begin{equation}
\log p(t_1, \dots, t_N | t_\textnormal{start}, t_\textnormal{end}, \lambda(\cdot)) = \sum_{i=1}^{N} \log \lambda(t_i) - \int_{t_\textnormal{start}}^{t_\textnormal{end}} \lambda(t) dt \nonumber 
\label{eq:poisson_process}
\end{equation}
%
Where $t_\textnormal{start}$ and $t_\textnormal{end}$ are the times at which observations started and stopped being recorded.

We augment the Latent ODE framework with a Poisson process over the observation times, where we parameterize $\lambda(t)$ as a function of $z(t)$.
This means that instead of specifying and maximizing the conditional marginal likelihood $p(x_1, \dots, x_N | t_1, \dots, t_N, \theta)$, we can instead specify and maximizing the joint marginal likelihood $p(x_1, \dots, x_N , t_1, \dots, t_N, | \theta)$.
%
%
%
To compute the joint likelihood, we can evaluate the Poisson intensity $\lambda(t)$, precisely estimate its integral, and the compute latent states at all required time points, using a single call to an ODE solver.

\citet{neural_hawkes} used a similar approach, but relied on a fixed time discretization to estimate the Poisson intensity. 
\citet{NeuralODE} showed a toy example of using Latent ODEs with a Poisson process likelihood to fit latent dynamics from observation times alone.
In section~\ref{sec:phys}, we incorporate a Poisson process likelihood into a latent ODE to model observation rates in medical data.


\subsection{Batching and computational complexity}
One computational difficulty that arises from irregularly-sampled data is that observation times can be different for each time series in a minibatch.
In order to solve all ODEs in a minibatch in sync, we must we must output the solution of the combined ODE at the union of all time points in the batch.

Taking the union of time points does not substantially hurt the runtime of the ODE solver, as the adaptive time stepping in ODE solvers is not sensitive to the number of time points $(t_1 ... t_N)$ at which the solver outputs the state.
Instead, it depends on the length on the time interval $[t_1, t_N]$ and the complexity of the dynamics. (see suppl. figure 3).
Thus, ODE-RNNs and Latent ODEs have a similar asymptotic time complexity to standard RNN models. 
However, as the ODE must be continuously solved even when no observations occur, the compute cost does not scale with the sparsity of the data, as it does in decay-RNNs.
In our experiments, we found that the ODE-RNN takes 60\% more time than the standard GRU to evaluate, and the Latent ODE required roughly twice the amount of time to evaluate than the ODE-RNN.

\subsection{When should you use an ODE-based model over a standard RNN?}


Standard RNNs are ignore the time gaps between points.
As such, standard RNNs work well on regularly spaced data, with few missing values, or when the time intervals between points are short.

Models with continuous-time latent state, such as the ODE-RNN or RNN-Decay, can be evaluated at any desired time point, and therefore are suitable for interpolation tasks.
In these models, the future hidden states depend on the time since the last observation, also making them better suited for sparse and/or irregular data than standard RNNs.
RNN-Decay enforces that the hidden state converges monontically to a fixed point over time.
In ODE-RNNs the form of the dynamics between the observations is learned rather than pre-defined.
Thus, ODE-RNNs can be used on sparse and/or irregular data without making strong assumptions about the dynamics of the time series.

\paragraph{Latent variable models versus autoregressive models}
We refer to models which iteratively compute the joint distribution ${p(x) = \prod_i p_\theta(x_i|x_{i-1}, \dots, x_0)}$ as autoregressive models (e.g.\ RNNs and ODE-RNNs).
We call models of the form ${p(x) = \int \prod_i p(x_i| z_0) p(z_0) dz_0}$ latent-variable models (e.g.\ Latent ODEs and RNN-VAEs). 

In autoregressive models, both the dynamics and the conditioning on data are encoded implicitly through the hidden state updates, which makes them hard to interpret.
In contrast, encoder-decoder models (Latent ODE and RNN-VAE) represent state explicitly through a vector $z_t$, and represent dynamics explicitly through a generative model. 
Latent states in these models can be used to compare different time series, for e.g.\ clustering or classification tasks, and their dynamics functions can be examined to identify the types of dynamics present in the dataset.

\section{Experiments}

\subsection{Toy dataset}
We tested our model on a toy dataset of 1,000 periodic trajectories with variable frequency and the same amplitude.
We sampled the initial point from a standard Gaussian, and added Gaussian noise to the observations. 
Each trajectory has 100 irregularly-sampled time points.
During training, we subsample a fixed number of points at random, and attempt to reconstruct the full set of 100 points.

\begin{figure}[b]
	\centering
	%
	\hspace{0cm}%
	\begin{subfigure}[b]{0.70\columnwidth}
	\centering
	\includegraphics[width=\columnwidth]{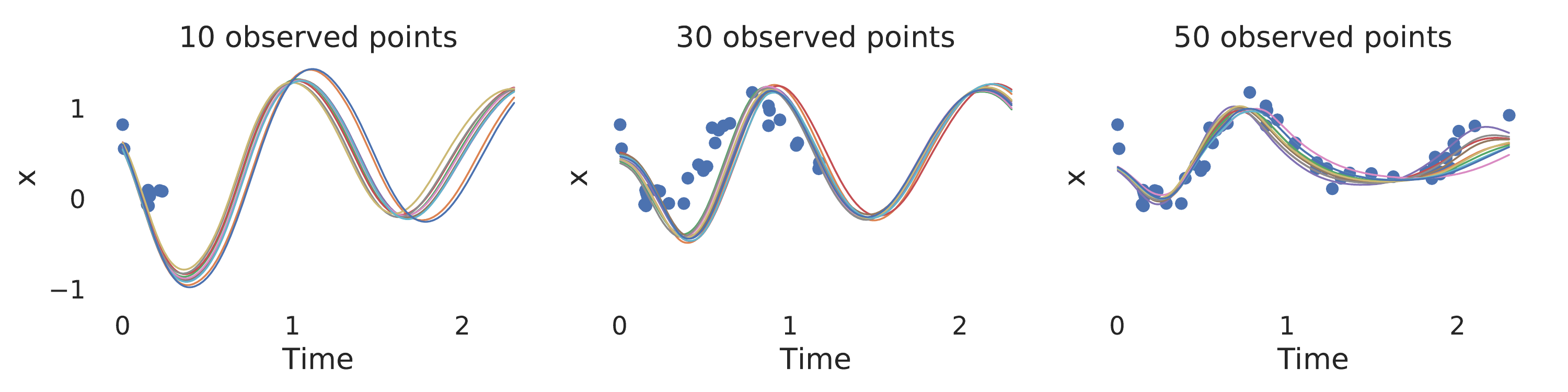}
	\caption{Conditioning on increasing number of observations}
	\end{subfigure}
	\begin{subfigure}[b]{0.29\columnwidth}
	\centering
	\vspace{8mm}
	\includegraphics[width=\columnwidth]{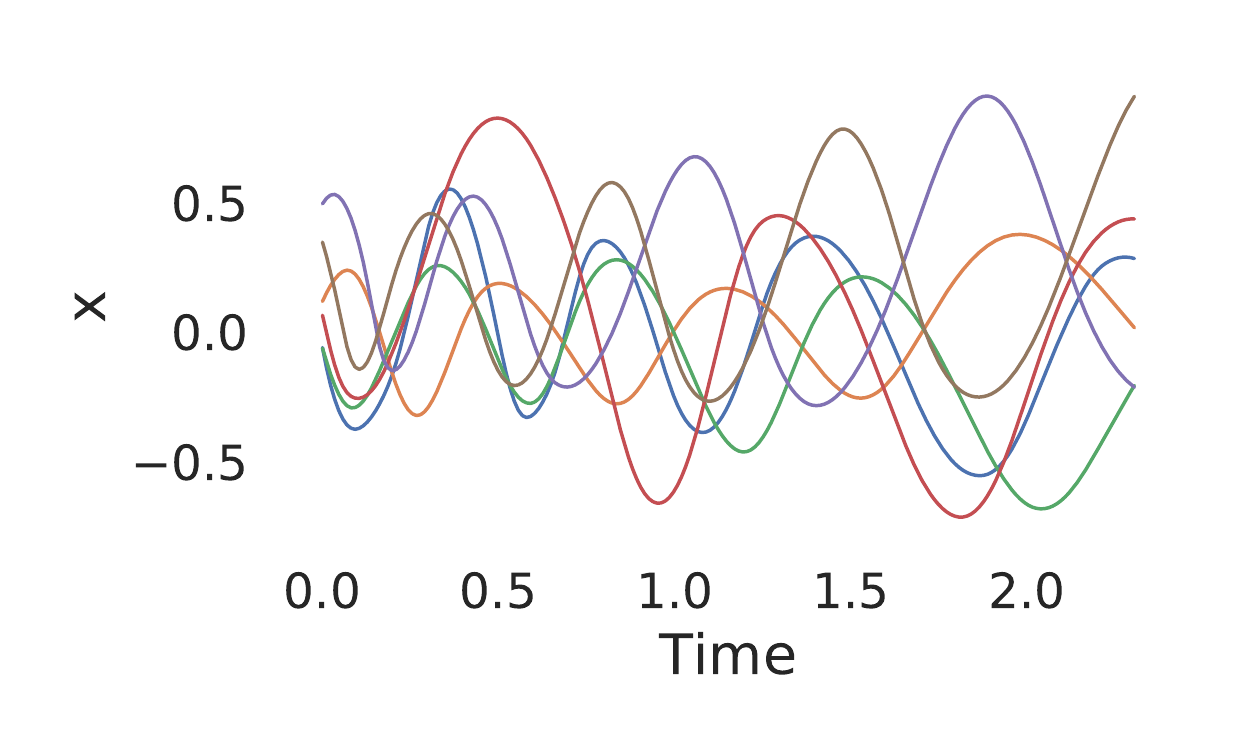}
	\caption{Prior samples}
	\end{subfigure}
	%
	\caption{(a) 
	A Latent ODE model conditioned on a small subset of points. 
	This model, trained on exactly 30 observations per time series, still correctly extrapolates when more observations are provided.
	(b) Trajectories sampled from the prior $p(z_0) \sim \textnormal{Normal}\big(z_0; 0, I\big)$ of the trained model, then decoded into observation space.}
	\label{fig:cond_on_subset}
\end{figure}

\paragraph{Conditioning on sparse data}
Latent ODEs can often reconstruct trajectories reasonably well given a small subset of points, and provide an estimate of uncertainty over both the latent trajectories and predicted observations.
To demonstrate this, we trained a Latent ODE model to reconstruct the full trajectory (100 points) from a subset of 30 points.
At test time, we conditioned this model on a subset of 10, 30 or 50 points.
Conditioning on more points results in a better fit as well as smaller variance across the generated trajectories (fig. \ref{fig:cond_on_subset}).
Figure~\ref{fig:cond_on_subset}(b) demonstrates that the trajectories sampled from the prior of the trained model are also periodic.



\begin{figure}[h]
	\centering
    \begin{subfigure}[b]{0.48\columnwidth}
    	\centering
    	\small
    	\caption{Latent ODE with RNN encoder}
        \includegraphics[width=\textwidth,trim={20 20 10 15},clip]{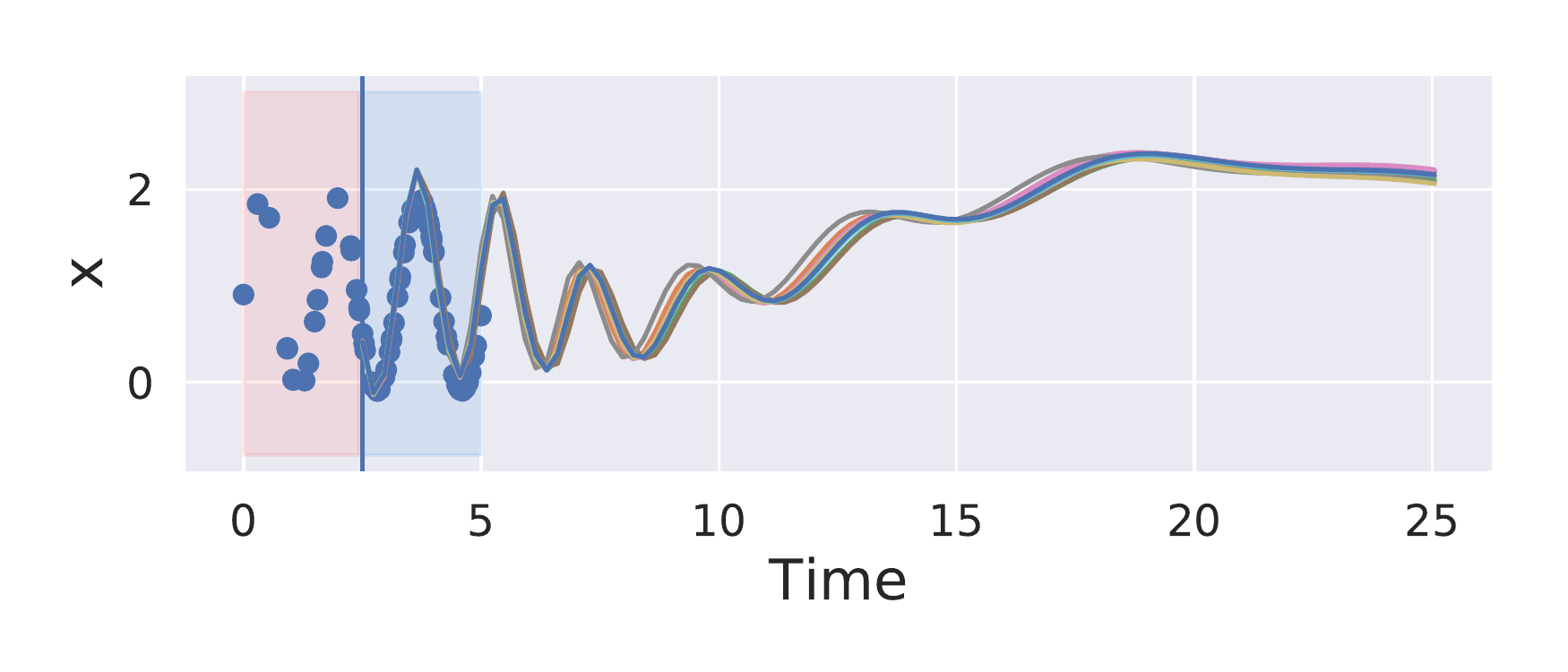}
    \end{subfigure}
    \begin{subfigure}[b]{0.48\columnwidth}
    	\centering
    	\small
    	\caption{Latent ODE with ODE-RNN encoder}
        \includegraphics[width=\textwidth,trim={5 20 25 15},clip]{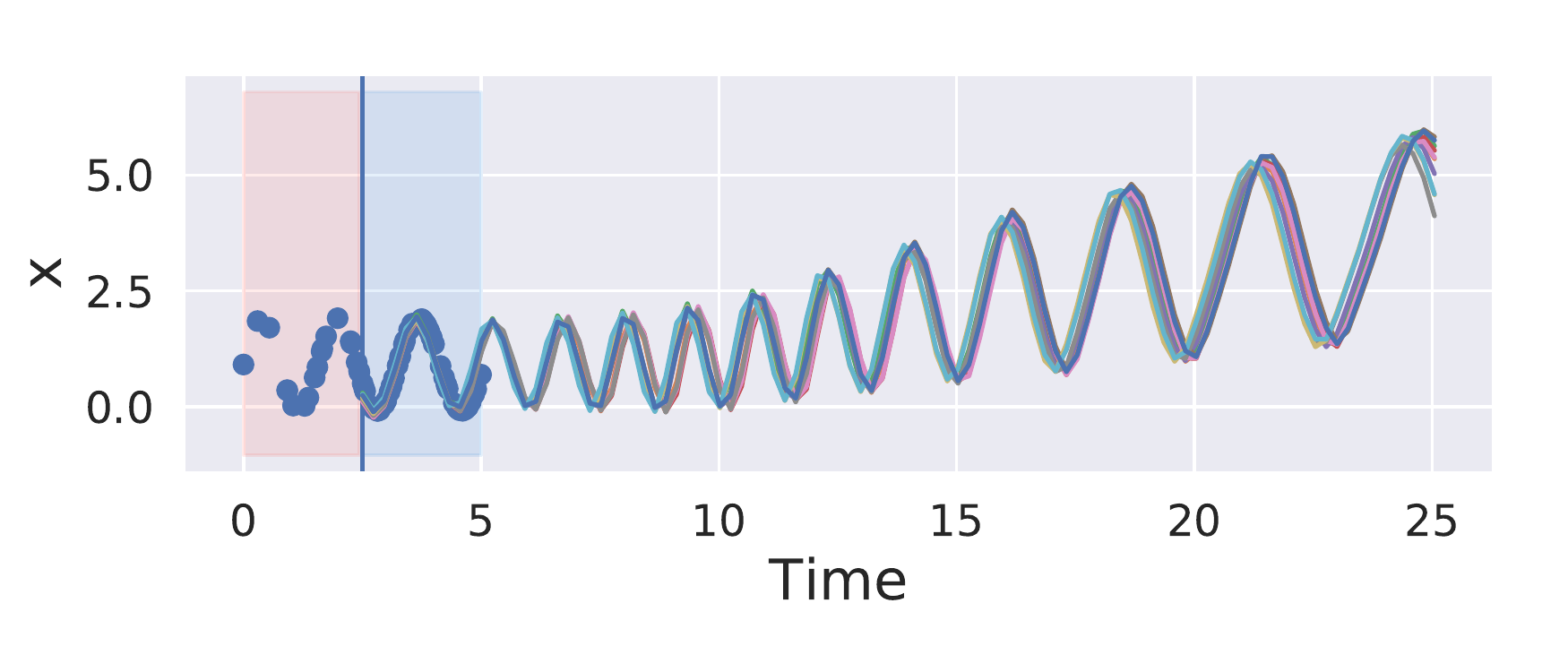}
    \end{subfigure}
    \caption{\textbf{(a)} Approximate posterior samples from a Latent ODE trained with an RNN recognition network, as in \cite{NeuralODE}.
    \textbf{(b)} Approximate posterior samples from a Latent ODE trained with an ODE-RNN recognition network (ours).
    At training time, the Latent ODE conditions on points in red area, and reconstruct points in blue area.
    At test time, we condition the model on 20 points in red area, and solve the generative ODE on a larger time interval.}
    \label{fig:periodic_extrap}
\end{figure}

\paragraph{Extrapolation}
Next, we show that a time-invariant ODE can recover stationary periodic dynamics from data automatically.
Figure~\ref{fig:periodic_extrap} shows a Latent ODE trained to condition on 20 points in the $[0; 2.5]$ interval (red area) and predict points on $[2.5; 5]$ interval (blue area).
A Latent ODE with an ODE-RNN encoder was able to extrapolate the time series far beyond the training interval and maintain periodic dynamics.
In contrast, a Latent ODE trained with RNN encoder as in \citet{NeuralODE} did not extrapolate the periodic dynamics well.

\subsection{Quantitative Evaluation}

We evaluate the models quantitavely on two tasks: interpolation and extrapolation.
On each dataset, we used 80\% for training and 20\% for test.
See the supplement a detailed description.

\paragraph{Baselines}
In the class of autoregressive models, we compare ODE-RNNs to standard RNNs.
We compared the following autoregressive models:
(1) ODE-RNN (proposed)
(2) A classic RNN where $\Delta_t$ is concatenated to the input (RNN-$\Delta_t$)
(3) An RNN with exponential decay on the hidden states $h \cdot e^{-\tau \Delta_t}$ (RNN-Decay)
(4) An RNN with missing values imputed by a weighted average of previous value and empirical mean (RNN-Impute), and
(5) GRU-D~\citep{che_sontag_2018} which combines exponential decay and the above imputation strategy.
Among encoder-decoder models, we compare the Latent ODE to a variational autoencoder in which both the encoder and decoder are recurrent neural nets (RNN-VAE). 
The ODE-RNN can use any hidden state update formula for the RNNCell function in Algorithm~\ref{alg:ode_gru}.
Throughout our experiments, we use the Gated Recurrent Unit (GRU)~\citep{gru_cho_2014}. See the supplement for the architecture details.

\paragraph{Interpolation}
The standard RNN and the ODE-RNN are straightforward to apply to the interpolation task.
To perform interpolation with a Latent ODE, we encode the time series backwards in time, compute the approximate posterior $q(z_0| \{x_i,t_i\}_{i=0}^N)$ at the first time point $t_0$, sample the initial state of ODE $z_0$, and generate mean observations at each observation time.

\paragraph{Extrapolation}
In the extrapolation setting, we use the standard RNN or ODE-RNN trained on the interpolation task, and then extrapolate the sequence by re-feeding previous predictions.
To encourage extrapolation, we used scheduled sampling~\citep{scheduled_sampling}, feeding previous predictions instead of observed data with probability 0.5 during training.
One might expect that directly optimizing for extrapolation would perform best at extrapolation.
Such a model would resemble an encoder-decoder model, which we consider separately below (the RNN-VAE).
For extrapolation in encoder-decoder models, including the Latent ODE, we split the timeline in half.
We encode the observations in the first half forward in time and reconstruct the second half.

\subsection{MuJoCo Physics Simulation}
\label{sec:mujoco}

Next, we demonstrated that ODE-based models can learn an approximation to simple Newtonian physics.
To show this, we created a physical simulation using the ``Hopper'' model from the Deepmind Control Suite~\citep{DeepMindControlSuite}.
We randomly sampled the initial position of the hopper and initial velocities such that hopper rotates in the air and falls on the ground (figure~\ref{fig:mujoco}).
These trajectories are deterministic functions of their initial states, which matches the assumptions made by the Latent ODE.
The dataset is 14-dimensional, and we model it with a 15-dimensional latent state.
We generated 10,000 sequences of 100 regularly-sampled time points each.

We perform both interpolation and extrapolation tasks on the MuJoCo dataset.
During training, we subsampled a small percentage of time points to simulate sparse observation times.
For evaluation, we measured the mean squared error (MSE) on the full time series.

\begin{table*}[h]
    \small
    \centering
	\caption{Test Mean Squared Error (MSE) ($\times 10^{-2}$) on the MuJoCo dataset.}
	\label{tab:mujoco}
	\centering
	\begin{tabular}{@{}cl|cccc|cccc@{}}
		 & & \multicolumn{4}{c|}{Interpolation (\% Observed Pts.)} & \multicolumn{4}{c}{Extrapolation (\% Observed Pts.)} \\
        & Model & 10\% & 20\% & 30\% & 50\% & 10\% & 20\% & 30\% & 50\% \\
         \midrule
         \multirow{3}{*}{ \rotatebox{90}{Autoreg}} & RNN $\Delta_t$ & 2.454 & 1.714 & 1.250 & 0.785 & \textbf{7.259} & \textbf{6.792} & \textbf{6.594} & 30.571 \\
         & RNN GRU-D & 1.968 & 1.421 & 1.134 & 0.748 & 38.130 & 20.041 & 13.049 & \textbf{5.833} \\
         & ODE-RNN {\scriptsize(Ours)} & \textbf{1.647} & \textbf{1.209} & \textbf{0.986} & \textbf{0.665} & 13.508 & 31.950 & 15.465 & 26.463 \\
         \midrule
        \multirow{3}{*}{\rotatebox{90}{Enc-Dec}} & RNN-VAE   & 6.514 & 6.408 & 6.305 & 6.100 & 2.378 & 2.135 & 2.021 & 1.782 \\
        & Latent ODE {\scriptsize(RNN enc.)} & 2.477 & 0.578 & 2.768 & 0.447 & 1.663 & 1.653 & 1.485 & 1.377 \\
        & Latent ODE {\scriptsize(ODE enc, ours)} & \textbf{0.360} & \textbf{0.295} & \textbf{0.300} & \textbf{0.285} & \textbf{1.441} & \textbf{1.400} & \textbf{1.175} & \textbf{1.258} 
	\end{tabular}
\end{table*}

Table~\ref{tab:mujoco} shows mean squared error for models trained on different percentages of observed points.
Latent ODEs outperformed standard RNN-VAEs on both interpolation and extrapolation.
Our ODE-RNN model also outperforms standard RNNs on the interpolation task.
The gap in performance between RNN and ODE-RNN increases with sparser data.
Notably, the Latent ODE (an encoder-decoder model) shows better performance than the ODE-RNN (an autoregressive model).

All autoregressive models performed poorly at extrapolation.
This is expected, as they were only trained for one-step-ahead prediction, although standard RNNs performed better than ODE-RNNs.
Latent ODEs outperformed RNN-VAEs on the extrapolation task.







\paragraph{Interpretability of the latent state}
Figure~\ref{fig:mujoco} shows how the norm of the latent state time-derivative $f_\theta(z)$ changes with time for two reconstructed MuJoCo trajectories.
When the hopper hits the ground, there is a spike in the norm of the ODE function.
In contrast, when the hopper is lying on the ground, the norm of the dynamics is small. 

\newcommand{\nfwidth}{0.09\textwidth}%
\newcommand{\trajid}{9}%
\newcommand{\anothertrajid}{5}%
\begin{figure*}[b]
	\begin{subfigure}[b]{0.07\linewidth}
	    \small
		Truth
		\vspace{3mm}
	\end{subfigure}
	\begin{subfigure}[b]{0.9\linewidth}
		\centering
		\includegraphics[width=\nfwidth,trim={2.5cm 2.5cm 2.5cm 2.5cm},clip]{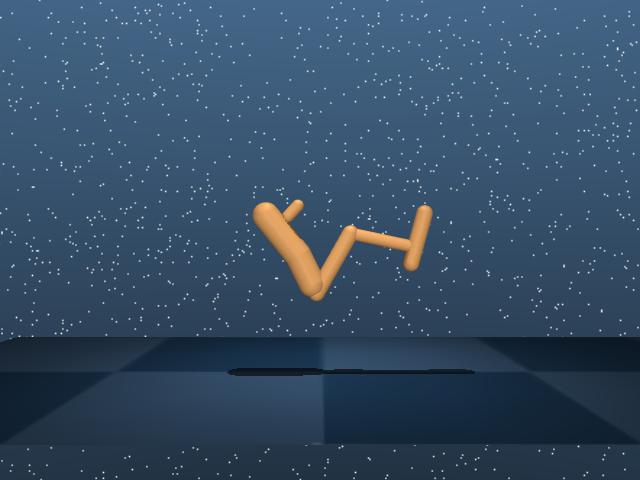}
		\includegraphics[width=\nfwidth,trim={2.5cm 2.5cm 2.5cm 2.5cm},clip]{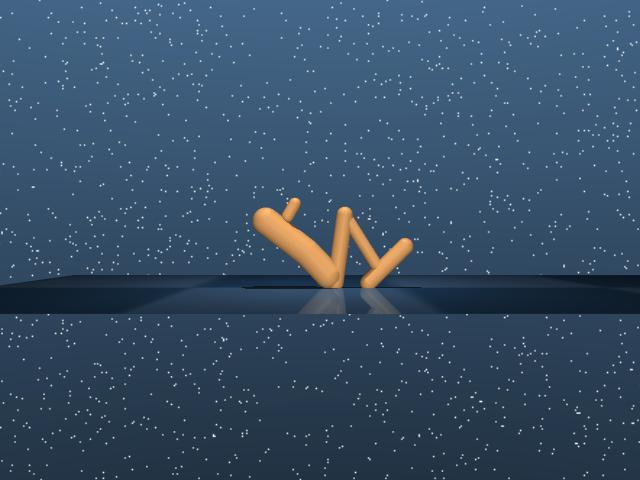}
		\includegraphics[width=\nfwidth,trim={2.5cm 2.5cm 2.5cm 2.5cm},clip]{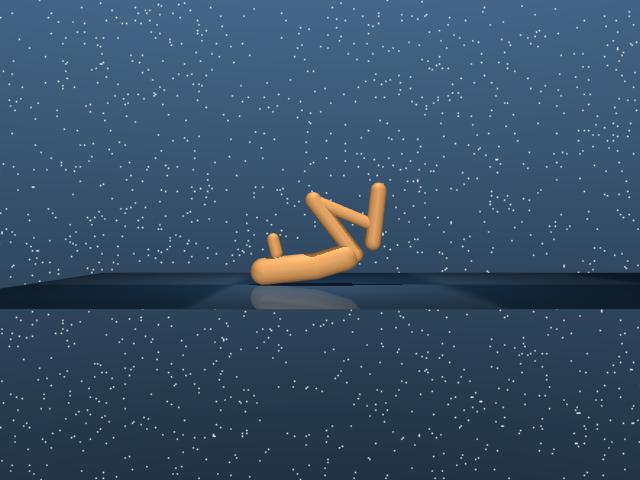}
		\includegraphics[width=\nfwidth,trim={2.5cm 2.5cm 2.5cm 2.5cm},clip]{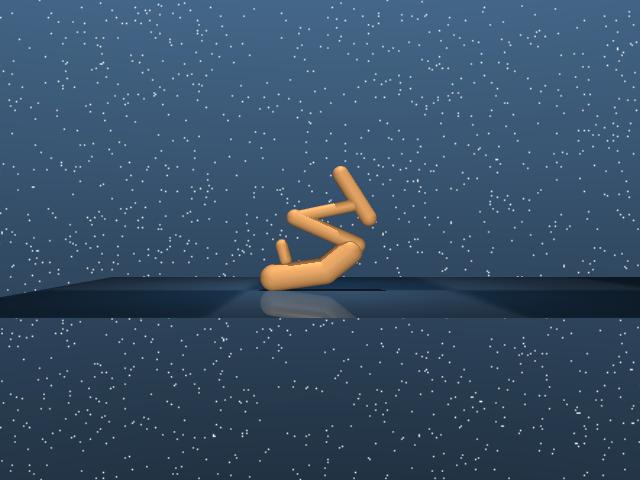}
		\includegraphics[width=\nfwidth,trim={2.5cm 2.5cm 2.5cm 2.5cm},clip]{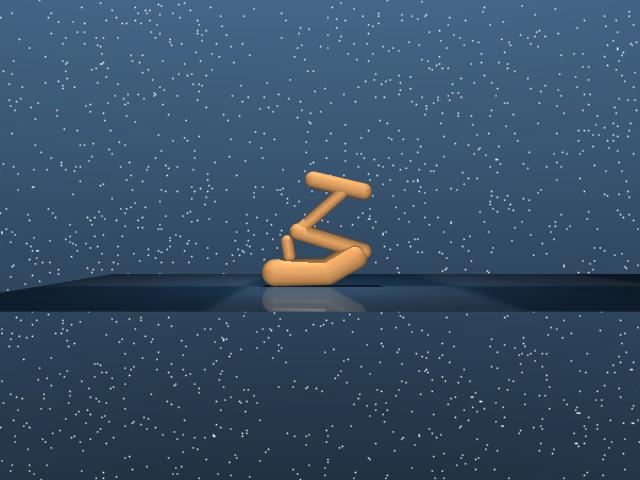}
		\hfill
		\includegraphics[width=\nfwidth,trim={2.5cm 2.5cm 2.5cm 2.5cm},clip]{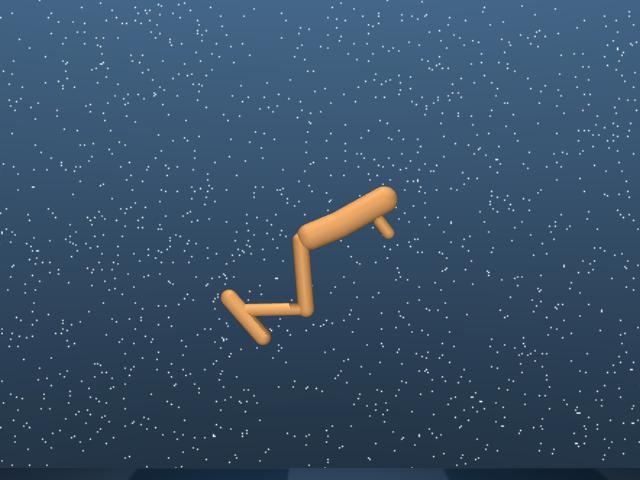}
		\includegraphics[width=\nfwidth,trim={2.5cm 2.5cm 2.5cm 2.5cm},clip]{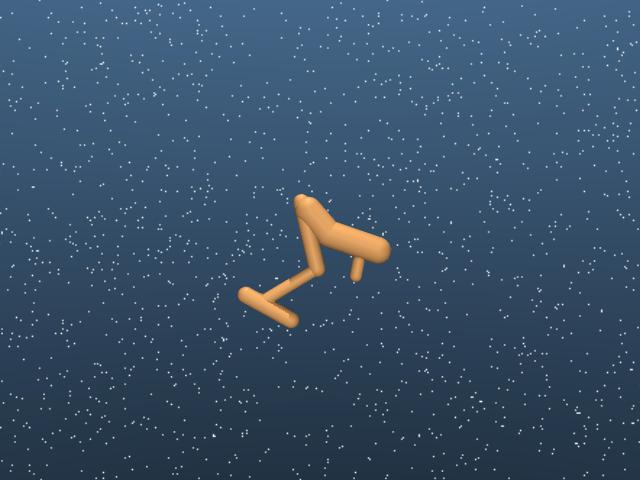}
		\includegraphics[width=\nfwidth,trim={2.5cm 2.5cm 2.5cm 2.5cm},clip]{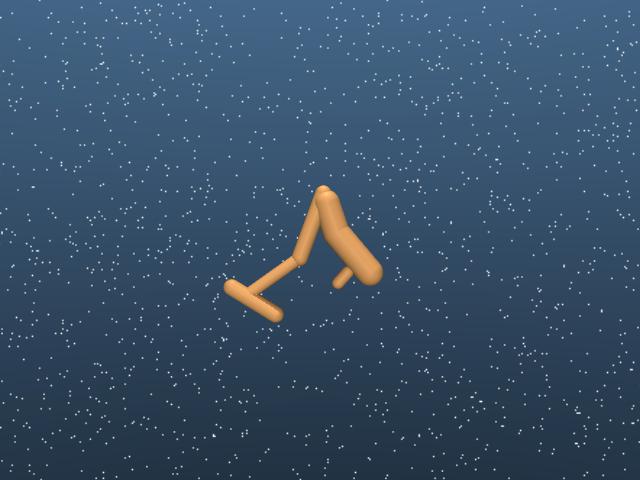}
		\includegraphics[width=\nfwidth,trim={2.5cm 2.5cm 2.5cm 2.5cm},clip]{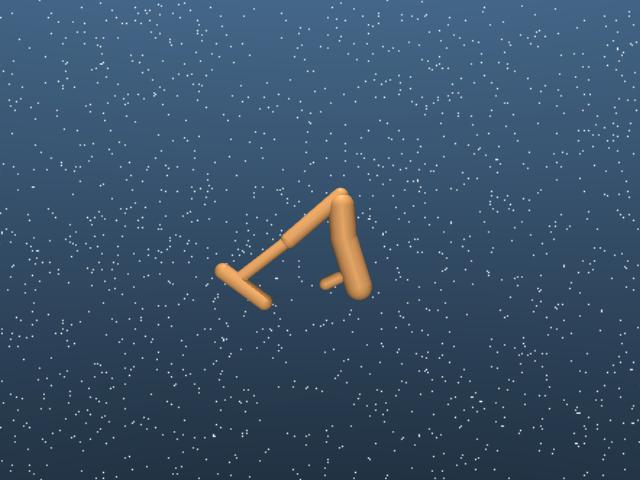}
		\includegraphics[width=\nfwidth,trim={2.5cm 2.5cm 2.5cm 2.5cm},clip]{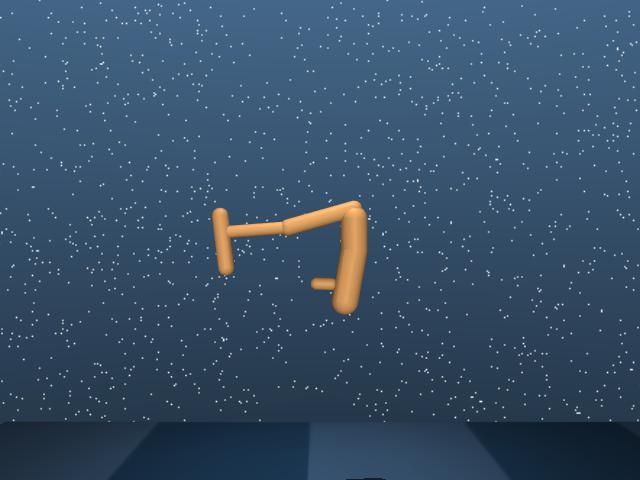}
	\end{subfigure}
	%
	\begin{subfigure}[b]{0.07\linewidth}
		\small
    	 Latent\\ODE
    	 \vspace{2mm}
	\end{subfigure}
	\begin{subfigure}[b]{0.9\linewidth}
		\centering
		\includegraphics[width=\nfwidth,trim={2.5cm 2.5cm 2.5cm 2.5cm},clip]{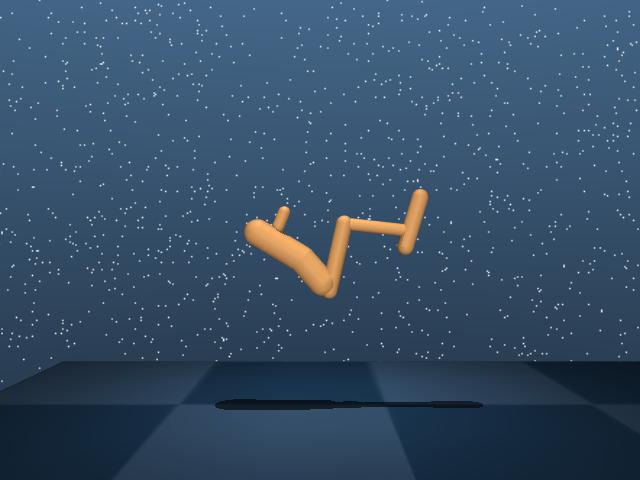}
		\includegraphics[width=\nfwidth,trim={2.5cm 2.5cm 2.5cm 2.5cm},clip]{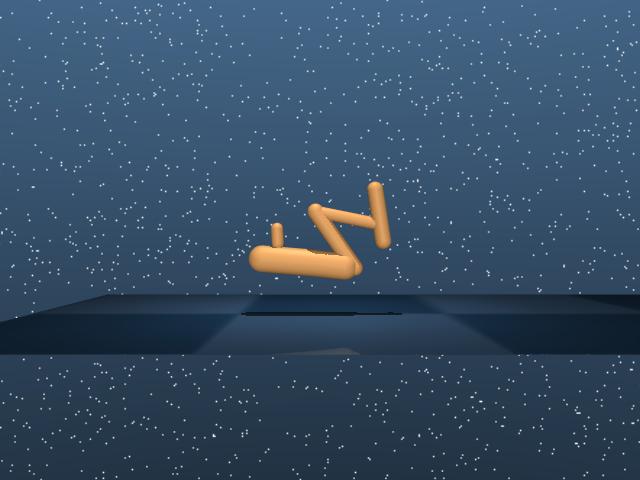}
		\includegraphics[width=\nfwidth,trim={2.5cm 2.5cm 2.5cm 2.5cm},clip]{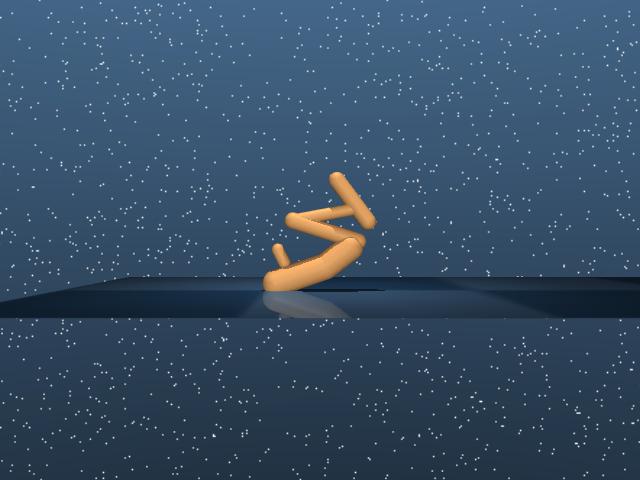}
		\includegraphics[width=\nfwidth,trim={2.5cm 2.5cm 2.5cm 2.5cm},clip]{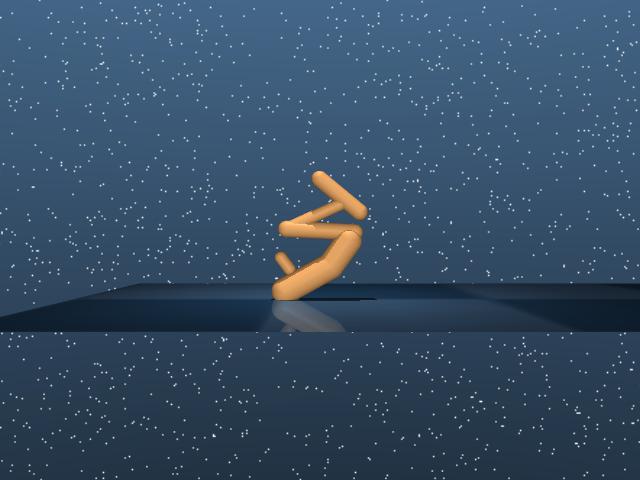}
		\includegraphics[width=\nfwidth,trim={2.5cm 2.5cm 2.5cm 2.5cm},clip]{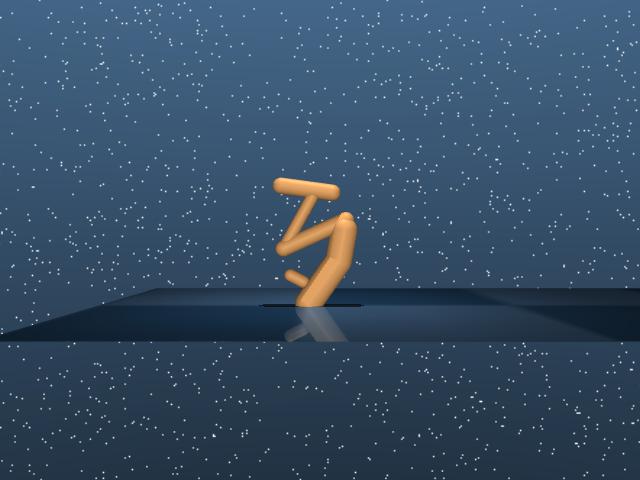}
		\hfill
		\includegraphics[width=\nfwidth,trim={2.5cm 2.5cm 2.5cm 2.5cm},clip]{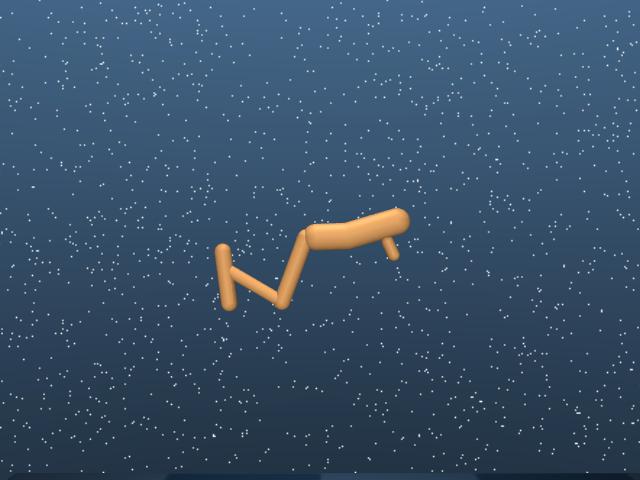}
		\includegraphics[width=\nfwidth,trim={2.5cm 2.5cm 2.5cm 2.5cm},clip]{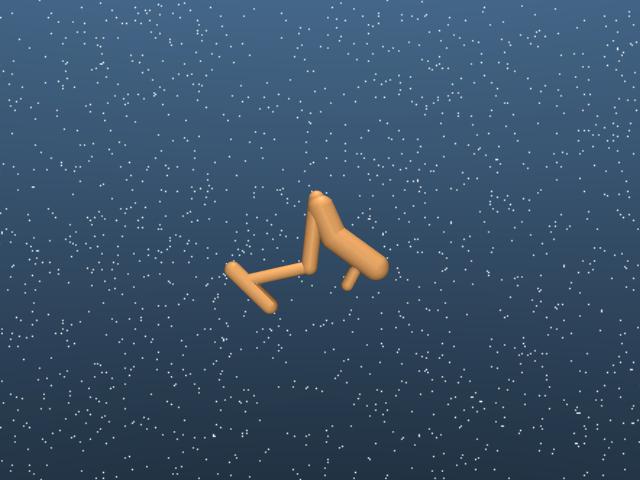}
		\includegraphics[width=\nfwidth,trim={2.5cm 2.5cm 2.5cm 2.5cm},clip]{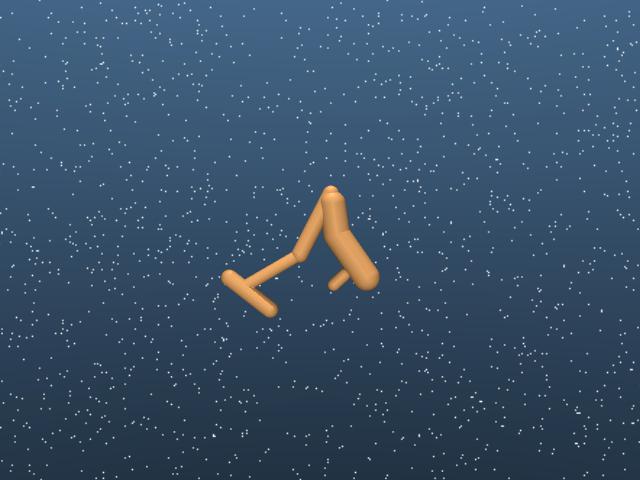}
		\includegraphics[width=\nfwidth,trim={2.5cm 2.5cm 2.5cm 2.5cm},clip]{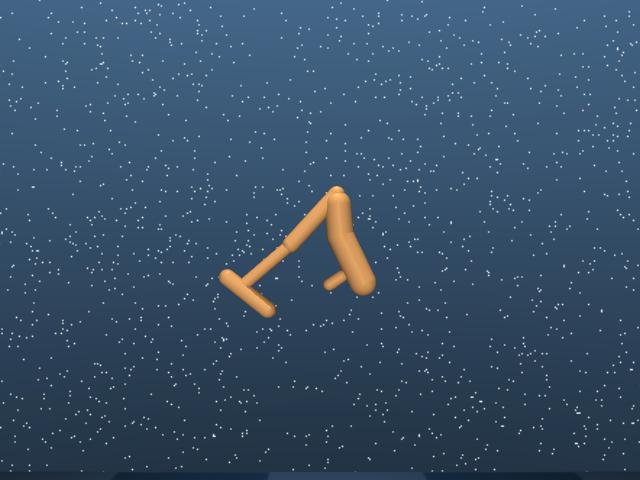}
		\includegraphics[width=\nfwidth,trim={2.5cm 2.5cm 2.5cm 2.5cm},clip]{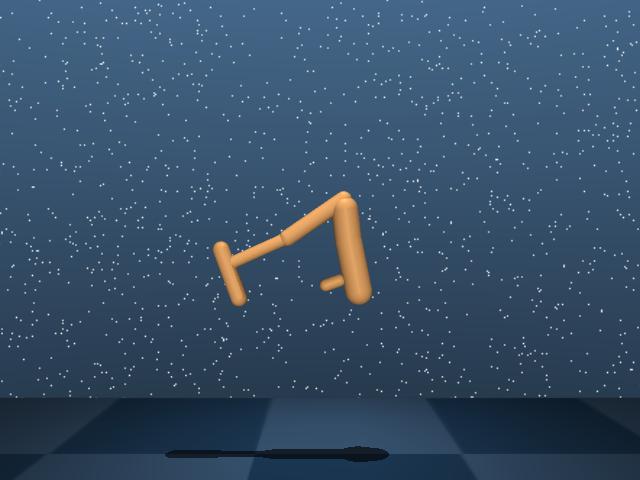}
	\end{subfigure}
	%
	\begin{subfigure}[b]{0.06\linewidth}
	\footnotesize
	$||f(z)||$\\(ODE)
	\vspace{1mm}
	\end{subfigure}
	\begin{subfigure}[b]{0.93\linewidth}
		\centering
		\includegraphics[width=0.49\linewidth,trim={0 5mm 0 0},clip]{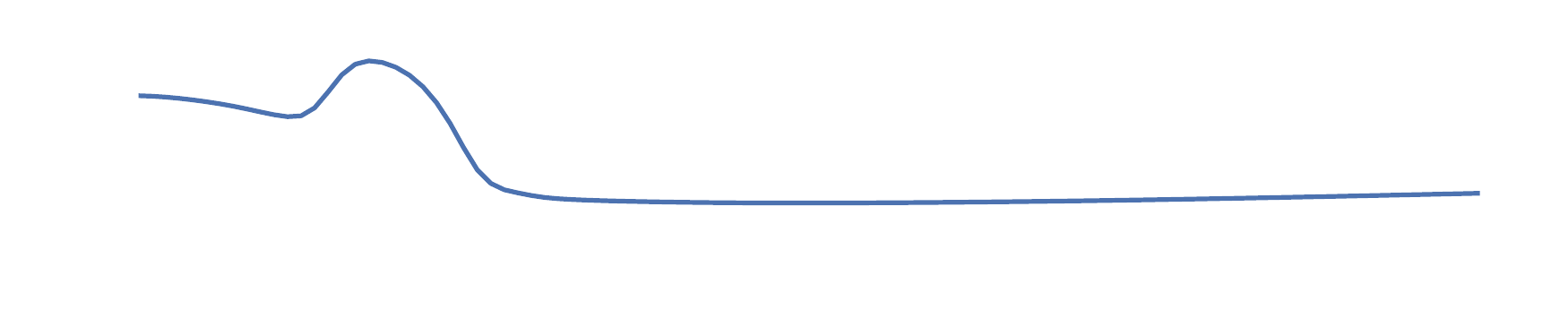}
		\includegraphics[width=0.49\linewidth,trim={0 5mm 0 0},clip]{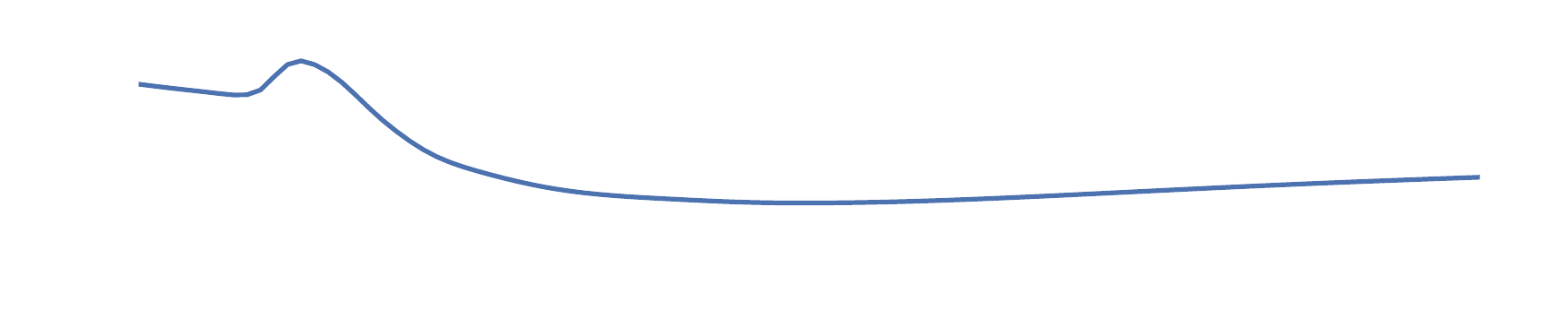}
	\end{subfigure}
	\vspace{-5mm}
	\begin{subfigure}[b]{0.06\linewidth}
	{\footnotesize $||\Delta h||$\\(RNN)}
	\vspace{2mm}
	\end{subfigure}
	\begin{subfigure}[b]{0.93\linewidth}
		\centering
		\includegraphics[width=0.49\linewidth,trim={0 0 0 5mm},clip]{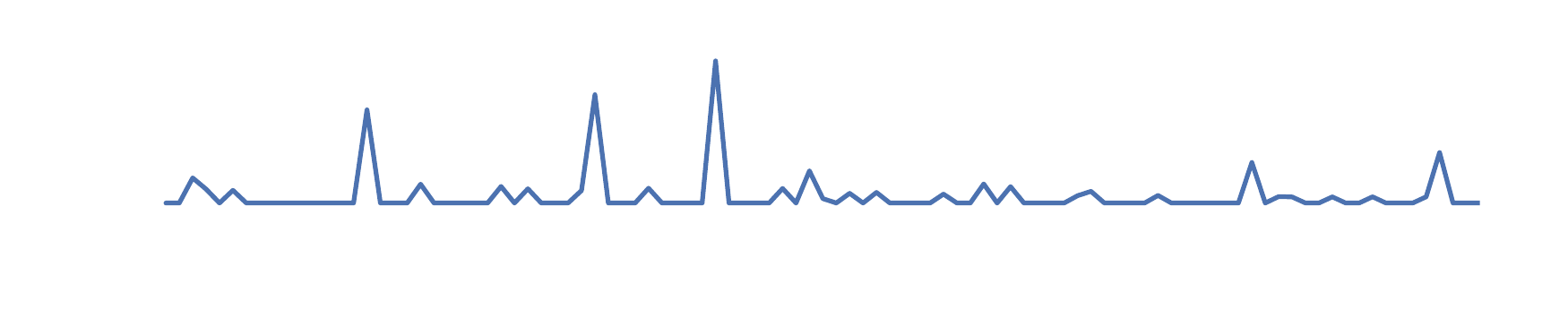}
		\includegraphics[width=0.49\linewidth,trim={0 0 0 5mm},clip]{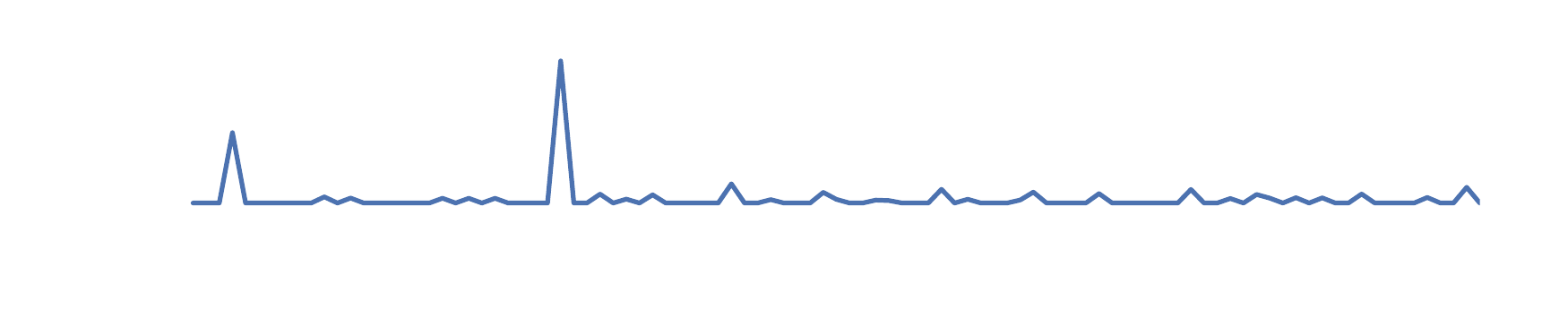}
	\end{subfigure}
	\begin{subfigure}[b]{0.05\linewidth}
	\hfill
	\end{subfigure}
	\begin{subfigure}[b]{0.465\linewidth}
		\centering
		\hspace{3mm}
		\small Time
	\end{subfigure}
	\begin{subfigure}[b]{0.465\linewidth}
		\centering
		\hspace{3mm}
		\small Time
	\end{subfigure}
	%
	\caption{\emph{Top row:} True trajectories from MuJoCo dataset.
	\emph{Second row:} Trajectories reconstructed by a latent ODE model.
	\emph{Third row:} Norm of the dynamics function $f_\theta$ in the latent space of the latent ODE model.
	\emph{Fourth row:} Norm of the hidden state of a RNN trained on the same dataset.
	}
	\vspace{-3mm}
\label{fig:mujoco}
\end{figure*}

\begin{figure}[h]
\centering
\begin{minipage}{.35\textwidth}
\centering
\includegraphics[width=\columnwidth,trim={12 20 12 10},clip]{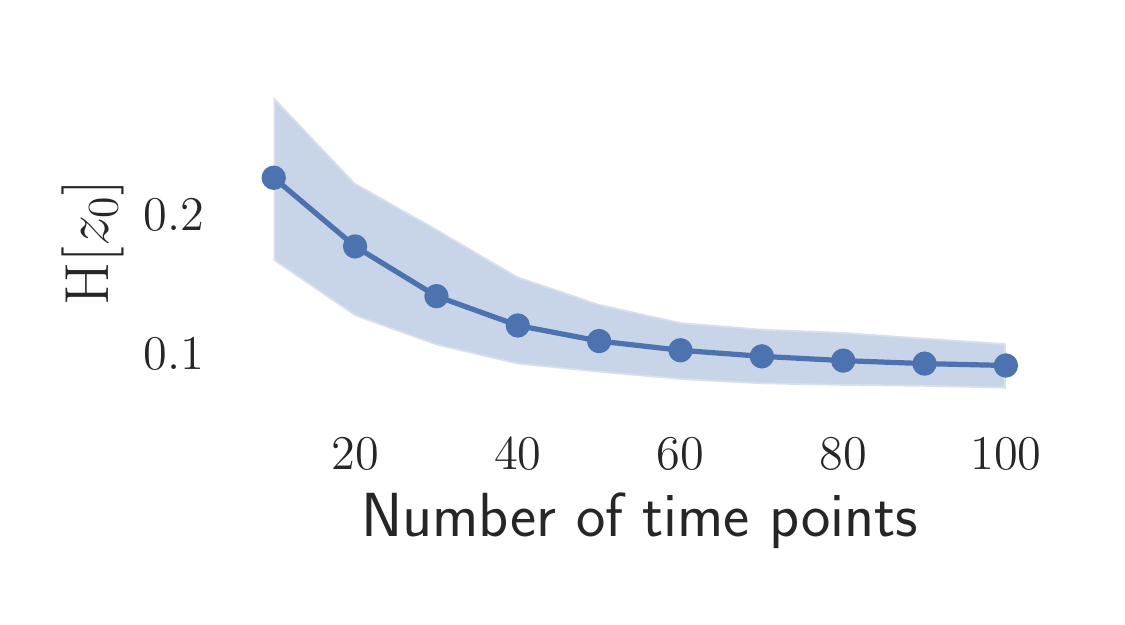}
\caption{Entropy of the approximate posterior over $z_0$ versus number of observed time points. The line shows the mean; shaded area shows 10\% and 90\% percentiles estimated over 1000 trajectories}
\label{fig:mujoco_h0_std}
\end{minipage}
\hfill
\begin{minipage}{.62\textwidth}
	\begin{subfigure}[b]{0.32\linewidth}
	\centering
    \includegraphics[width=0.9\columnwidth, trim=70 65 70 20, clip]{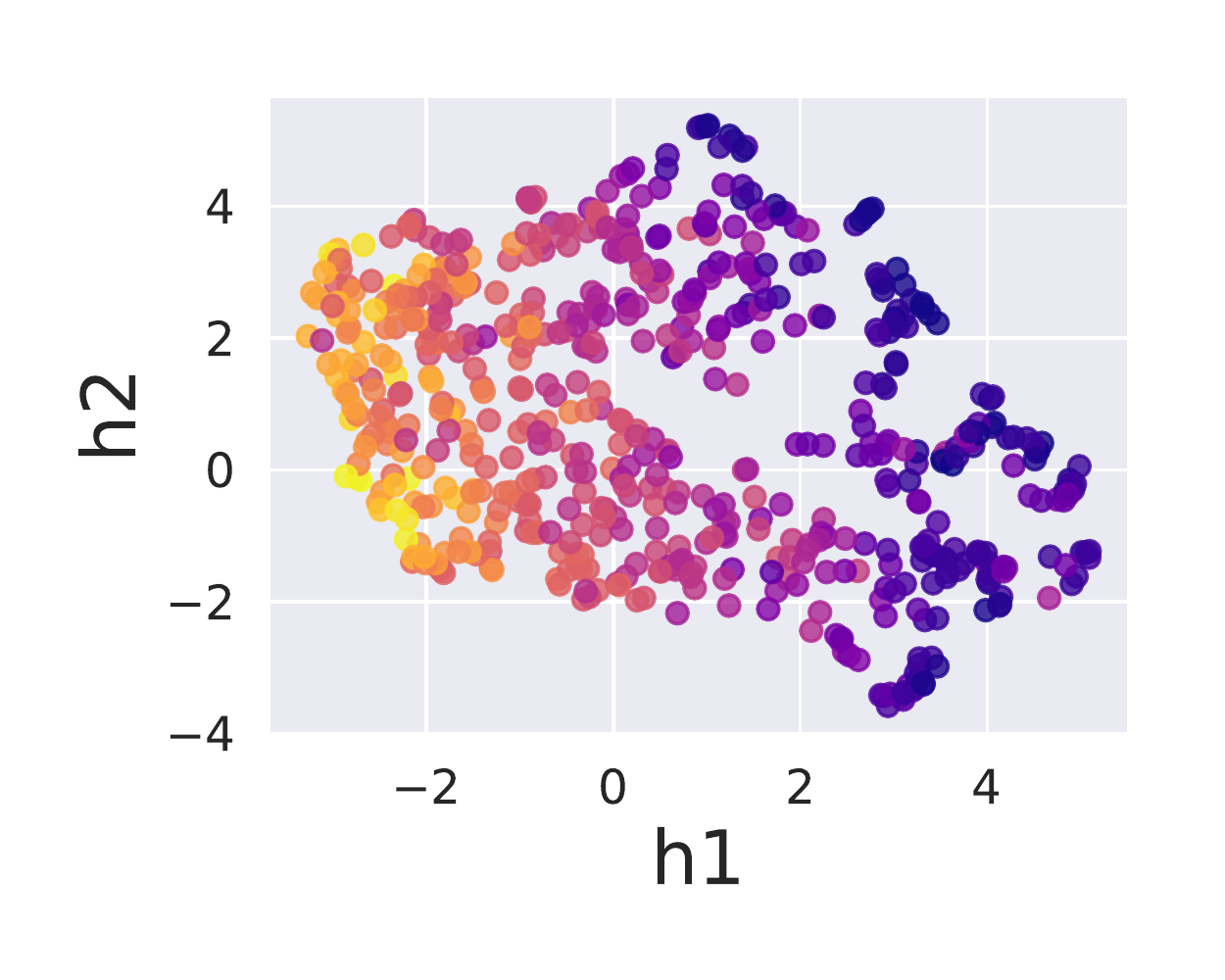}
    \caption{Height}
    \end{subfigure}%
    \begin{subfigure}[b]{0.32\linewidth}
	\centering
    \includegraphics[width=0.9\columnwidth, trim=70 65 70 20, clip]{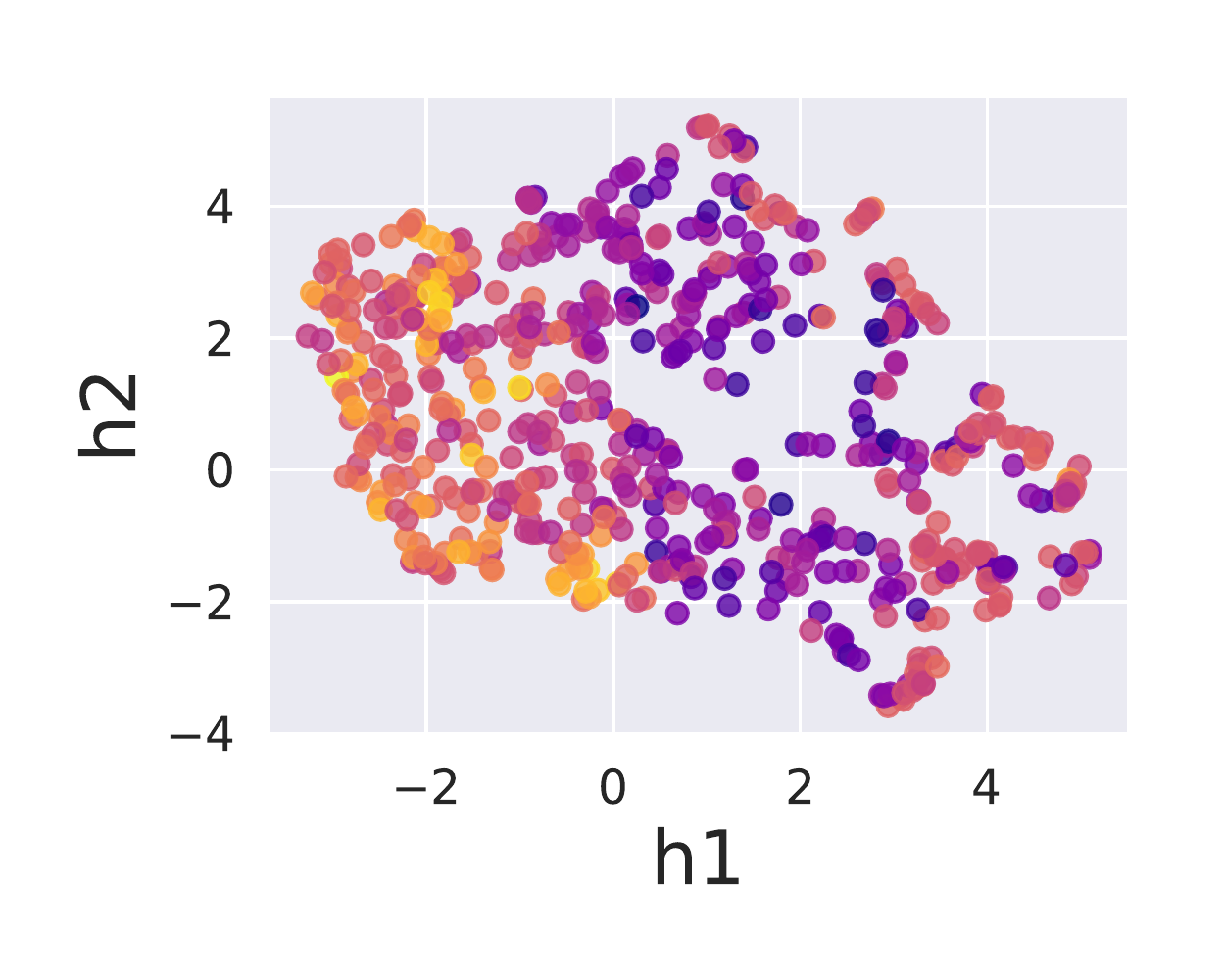}
    \caption{Velocity}
	\end{subfigure}%
	\begin{subfigure}[b]{0.32\linewidth}
	\centering
    \includegraphics[width=0.9\columnwidth, trim=70 65 70 20, clip]{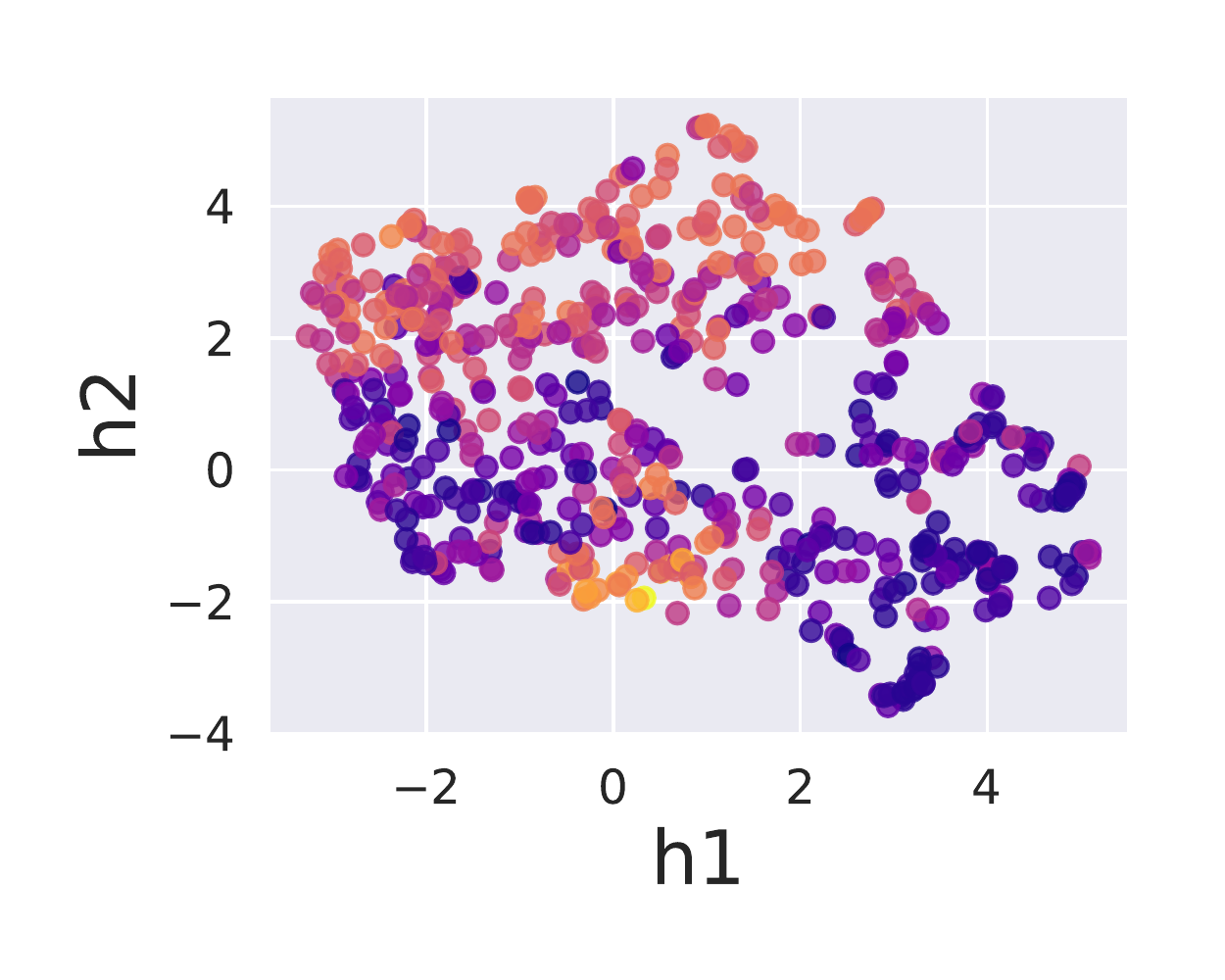}
    \caption{Hip Position}
	\end{subfigure}
\caption{Nonlinear projection of latent space of $z_0$ from a Latent ODE model trained on the MuJoCo dataset).
Each point is the encoding of one time series.
The points are colored by the (a) initial height (distance from the ground) (b) initial velocity in z-axis (c) relative initial position of the hip of the hopper.
The latent state corresponds closely to the physical parameters of the true simulation.}
\label{fig:mujoco_h0space}
\end{minipage}
\vspace{-5mm}
\end{figure}

Figure~\ref{fig:mujoco_h0_std} shows the entropy of the approximate posterior $q(z_0| \{x_i,t_i\}_{i=0}^N)$ of a trained model conditioned on different numbers of observations. 
The average entropy (uncertainty) monotonically decreases as more points are observed.
Figure~\ref{fig:mujoco_h0space} shows the latent state $z_0$ projected to 2D using UMAP~\citep{umap}.
The latent state corresponds closely to the physical parameters of the true simulation that most strongly determine the future trajectory of the hopper: distance from the ground, initial velocity on z-axis, and relative position of the leg of the hopper.

\subsection{Physionet}
\label{sec:phys}


We evaluated our model on the PhysioNet Challenge 2012 dataset~\citep{Silva2012}, which contains 8000 time series, each containing measurements from the first 48 hours of a different patient's admission to ICU.
Measurements were made at irregular times, and of varying sparse subsets of the 37 possible features.

\begin{figure}[h]
	\centering
    \includegraphics[width=0.9\columnwidth, clip, trim=5 15 5 8]{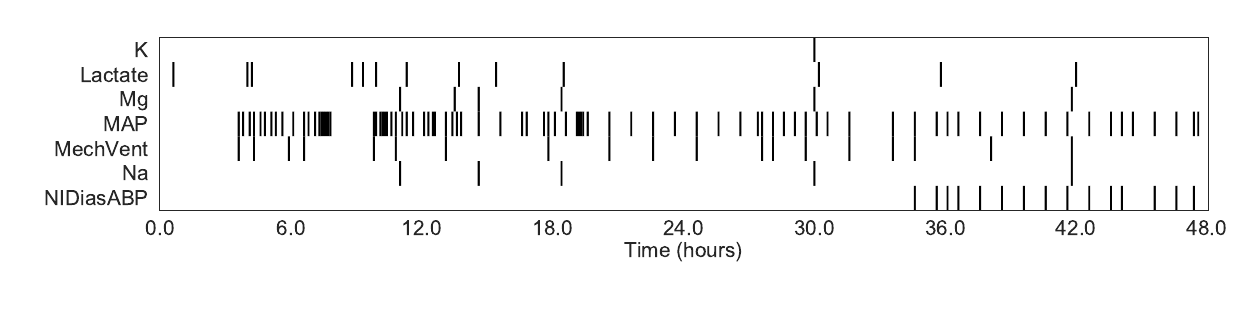}
    \caption{Observation times of a subset of features for one patient in the Physionet dataset.
    Black lines indicate observation times, whose number and timing vary across patients.}
    \label{fig:phys}
\end{figure}

Most existing approaches to modeling this data use a coarse discretization of the aggregated measurements per hour~\citep{che_sontag_2018}, which forces the model to train on only one-twentieth of measurements.
In contrast, our approach, in principle, does not require any discretization or aggregation of measurements.
To speed up training, we rounded the observation times to the nearest minute, reducing the number of measurements only 2-fold.
Hence, there are still 2880 (60*48) possible measurement times per time series under our model's preprocessing, while the previous standard was to used only 48 possible measurement times.
We used 20 latent dimensions in the latent ODE generative model. See supplement for more details on hyperparameters.

Tables~\ref{tab:phys_autoreg} and \ref{tab:phys_autoencoder} report mean squared error averaged over runs with different random seeds, and their standard deviations. We run one-sided t-test to establish a statistical significance. Best models are marked in bold. ODE-based models have smaller mean squared error than RNN baselines on this dataset.

Finally, we constructed binary classifiers based on each model type to predict in-hospital mortality.
We passed the hidden state at the last measured time point into a two-layer binary classifier.
Due to class imbalance (13.75\% samples with positive label), we report test area under curve (AUC) instead of accuracy.
Table~\ref{tab:phys_classif} shows that the ODE-RNN, Latent ODE and GRU-D achieved the similar classification AUC.
A possible explanation is that modelling dynamics between time points does not make a difference for binary classification of the full time series.

We also included a Poisson Process likelihood on observation times, jointly trained with the Latent ODE model.
Figure~\ref{fig:poisson} shows the inferred measurement rate on a patient from the dataset.
Although the Poisson process was able to model observation times reasonably well, including this likelihood term did not improve classification accuracy.

\begin{table}
    \parbox{0.36\linewidth}{
        \centering
        \captionsetup{justification=centering}
        \caption{Test MSE (mean $\pm$ std) on PhysioNet. \textbf{Autoregressive} models.}
    	\label{tab:phys_autoreg}
        \begin{tabular}{@{}lc@{}}
        \toprule
         Model & {\small Interp ($\times 10^{-3}$)}\\
        \midrule
        RNN $\Delta_t$ &   3.520 $\pm$ 0.276 \\
        RNN-Impute & 3.243 $\pm$ 0.275 \\
        RNN-Decay &  3.215 $\pm$ 0.276 \\
        RNN GRU-D & 3.384 $\pm$ 0.274 \\
        \addlinespace[2pt]
        \hdashline
        \addlinespace[2pt]
        ODE-RNN (Ours) & \textbf{2.361 $\pm$ 0.086} \\
        \bottomrule
        \end{tabular}
    }
    \hfill
    \parbox{0.6\linewidth}{ 
        \centering
        \captionsetup{justification=centering}
        \caption{Test MSE (mean $\pm$ std) on PhysioNet.\\\textbf{Encoder-decoder} models.}
    	\label{tab:phys_autoencoder}
        \begin{tabular}{@{}lcc@{}}
        \toprule
        Model & {\small Interp ($\times 10^{-3}$)} & {\small Extrap ($\times 10^{-3}$)}\\
        \midrule
        RNN-VAE & 5.930 $\pm$ 0.249 & 3.055 $\pm$ 0.145 \\
        Latent ODE (RNN enc.) & 3.907 $\pm$ 0.252 & 3.162 $\pm$ 0.052\\
        \addlinespace[2pt]
        \hdashline
        \addlinespace[2pt]
        Latent ODE (ODE enc) & \textbf{2.118 $\pm$ 0.271} & \textbf{2.231 $\pm$ 0.029} \\
        Latent ODE + Poisson & 2.789 $\pm$ 0.771 & \textbf{2.208 $\pm$ 0.050}\\
        \bottomrule
        \end{tabular}
    }
\end{table}

\subsection{Human Activity dataset}

We trained the same classifier models as above on the Human Activity dataset, which contains time series from five individuals performing various activities: walking, sitting, lying, etc.
The data consists of 3d positions of tags attached to their belt, chest and ankles (12 features in total).
After preprocessing, the dataset has 6554 sequences of 211 time points (details in supplement).
The task is to classify each time point into one of seven types of activities (walking, sitting, etc.).
We used a 15-dimensional latent state (more details in the supplement).
Table~\ref{tab:human_activity_classif} shows that the Latent ODE-based classifier had higher accuracy than the ODE-RNN classifier on this task.

\begin{table}
    \vspace{-4mm}
    \parbox{0.45\linewidth}{
    	\centering
    	\captionsetup{justification=centering}
        \caption{\textbf{Per-sequence classification.}\\ AUC on Physionet.}
    	\label{tab:phys_classif}
        \begin{tabular}{@{}lc@{}}
        \toprule
        Method & AUC\\
        \midrule
        RNN $\Delta_t$ & 0.787 $\pm$ 0.014  \\
        RNN-Impute & 0.764 $\pm$ 0.016\\
        RNN-Decay & 0.807 $\pm$ 0.003  \\
        RNN GRU-D & \textbf{0.818 $\pm$ 0.008}  \\
        RNN-VAE & 0.515 $\pm$ 0.040\\
        Latent ODE (RNN enc.) & 0.781 $\pm$ 0.018 \\
        \addlinespace[2pt]
        \hdashline
        \addlinespace[2pt]
        ODE-RNN & \textbf{0.833 $\pm$ 0.009} \\
        Latent ODE (ODE enc) & \textbf{0.829 $\pm$ 0.004}\\
        Latent ODE + Poisson & \textbf{0.826 $\pm$ 0.007}\\
        \bottomrule
        \end{tabular}
  }
    \hfill
    \parbox{0.45\linewidth}{ 
    	\centering
    	\captionsetup{justification=centering}
        \caption{\textbf{Per-time-point classification.}\\ Accuracy on Human Activity.}
    	\label{tab:human_activity_classif}
        \begin{tabular}{@{}lc@{}}
        \toprule
        Method &  Accuracy\\
        \midrule
        RNN $\Delta_t$ &  0.797 $\pm$ 0.003 \\
        RNN-Impute &  0.795 $\pm$ 0.008 \\
        RNN-Decay  & 0.800 $\pm$ 0.010 \\
        RNN GRU-D &  0.806 $\pm$ 0.007 \\
        RNN-VAE &   0.343 $\pm$ 0.040\\
        Latent ODE (RNN enc.) & 0.835 $\pm$ 0.010 \\
        \addlinespace[2pt]
        \hdashline
        \addlinespace[2pt]
        ODE-RNN &  0.829 $\pm$ 0.016 \\
        Latent ODE (ODE enc) &  \textbf{0.846 $\pm$ 0.013} \\
        \bottomrule
        \end{tabular}
    }
    \vspace{-4mm}
\end{table}

\section{Related work}

Standard RNNs treat observations as a sequence of tokens, not accounting for variable gaps between observations.
One way to accommodate this is to discretize the timeline into equal intervals, impute missing data, and then run an RNN on the imputed inputs.
To perform imputation, \citet{che_sontag_2018} used a weighted average between the empirical mean and the previous observation.
Others have used a separate interpolation network~\citep{shukla2018interpolationprediction}, Gaussian processes~\citep{pmlrv70futoma17a}, or generative adversarial networks~\citep{gan_imputation} to perform interpolation and imputation prior to running an RNN on time-discretized inputs.
In contrast, \citet{pmlr_Lipton16} used a binary mask to indicate the missing measurements and reported that RNNs performs better with zero-filling than with imputed values.
They note that such methods can be sensitive to the discretization granularity.

Another approach is to directly incorporate the time gaps between observations into RNN.
The simplest approach is to append the time gap $\Delta_t$ to the RNN input. 
However, \citet{mozer_2017} suggested that appending $\Delta_t$ makes the model prone to overfitting, and found empirically that it did not improve predictive performance.
Another solution is to introduce the hidden states that decay exponentially over time~\citep{che_sontag_2018, BRITS_2018, google_ehr_2018}.

\citet{neural_hawkes} used hidden states with exponential decay to parametrize neural Hawkes processes, and explicitly modeled observation intensities.
Hawkes processes are self-exciting processes whose latent state changes at each observation event.
This architecture is similar to our ODE-RNN.
In contrast, the Latent ODE model assumes that observations do not affect the latent state, but only affect the model's posterior over latent states, and is more appropriate when observations (such as taking a patient's temperature) do not substantially alter their state. 

\section{Discussion and conclusion}

We introduced a family of time series models, ODE-RNNs, whose hidden state dynamics are specified by neural ordinary differential equations (Neural ODEs).
We first investigated this model as a standalone refinement of RNNs.
We also used this model to improve the recognition networks of a variational autoencoder model known as Latent ODEs.
Latent ODEs provide relatively interpretable latent states, as well explicit uncertainty estimates about latent states.
Neither model requires discretizing observation times, or imputing data as a preprocessing step, making them suitable for the irregularly-sampled time series data common in many applications. 
Finally, we demonstrate that continuous-time latent states can be combined with Poisson process likelihoods to model the rates at which observations are made.

\subsubsection*{Acknowledgments}

We thank Chun-Hao Chang, Chris Cremer, Quaid Morris, and Ladislav Rampasek for helpful discussions and feedback.
We thank the Vector Institute for providing computational resources.

\bibliography{bibfile}
\small {\bibliographystyle{plainnat}}

\end{document}


\date{}
\maketitle

\section{Experiment setup}

We test the ODE models for two tasks: interpolation and extrapolation. 

\paragraph{Interpolation}
Consider time series with time points ($t_0$...$t_N$). In interpolation task we condition on the subset of points from ($t_0$...$t_N$) and reconstruct the full set of points in the same time interval. We subsample the points by setting the value to zero in the data tensor and in the mask. On MUJoCo dataset we perform experiments with different proportion of subsampled points (ranging from 10\% to 50\%, results shown in table 3 of main manuscript). In Physionet and Human Activity datasets, we do not perform subsampling since the data is already sparse.

\paragraph{Extrapolation}

In extrapolation task we split the time series into two parts: ($t_0$...$t_{N_2}$) and ($t_{N_2}$..$t_N$). We encode the first half of the time series and reconstruct the second half.

Similarly to interpolation task, we randomly sample a subset of time points from the time series and run a recognition network on this subset of points. We evaluate the model by the ability to reconstruct the full time series (without subsampling).

Autoencoder models are straightforward to use for extrapolation task. We run the encoder on the first half of the sequence and decode the second half. For autoregressive models, we train the model to perform interpolation first, and then perform extrapolation at test time by re-feeding previous predictions of the model. During training, we feed in either the previous observed value or predicted value with probability $0.5$, a common regularization method (Goodfellow et al., 2016). For all experiments, we report the mean squared error (MSE) on a test set of held-out sequences.

\paragraph{Choice of first time point of ODESolve} 
The generative model has a special time point $t_0$ where we put a prior on the latent state. ODE can be solved both forward or backward in time, and we are free to choose the time point $t_0$ depending on the task. As such, for interpolation task, we choose $t_0$ to be the time point of the first observation, as shown in eq. (5-7), and run ODE-RNN backwards in time to obtain the approximate posterior. For extrapolation task, we choose the initial point to the the last observed time point $t_N$. In this case, we run ODE-RNN encoder forward in time.



\begin{table}[h]
\caption{}
\label{tab:experim_setup}
\centering
\begin{tabular}{llcl}
\hline
Task & Encoder & Approx. posterior & Decoder \\
\hline
Interpolation  & Backwards in time $t_N \rightarrow t_0$ & $q(z(t_0)|x_0 .. x_N)$ & Forward in time in $[t_0, t_N]$ interval \\
Extrapolation & Forward in time $t_0 \rightarrow t_{N/2}$ & $q(z(t_{N/2})|x_0 .. x_{N/2})$ & Forward in time in $[t_{N/2}, t_N]$ interval \\
\hline
\end{tabular}
\end{table}

\section{Model details}
\subsection{GRU update}

\begin{algorithm}[H]
	\caption{GRU}
	\label{alg:gru}
	\begin{algorithmic}
    	\State {\bfseries Input:} Observations $x$, previous hidden state $h_{prev}$
        \State $z = \sigma(f_z([h_{prev}; x]))$ \Comment{Update coefficient}
    	\State $r = \sigma(f_r([h_{prev}; x]))$ \Comment{Reset coefficient}
    	\State $h' = g([r * h_{prev}; x])$  \Comment{Proposed new state}
    	\State $h = (1-z) * h' + z * h_{prev}$
    	\State {\bfseries Return:} new hidden state $h$
	\end{algorithmic}
\end{algorithm}

\subsection{Latent ODE}

\begin{algorithm}[H]
	\caption{Latent ODE}
	\label{alg:ode_vae}
	\begin{algorithmic}
	\State {\bfseries Input:} Data points $\{x_{i}\}_{i=1..N}$ and corresponding times $\{t_{i}\}_{i=1..N}$
	\State $z'_{0} = \text{ODE-RNN}(\{x_{i}\}_{i=1..N})$
	\State $\mu_{z_0}, \sigma_{z_0} = g_\mu(z'_{0}), g_\sigma(z'_{0})$  \Comment{$g_\mu$ and $g_\sigma$ are feed-forward NN}
	\State $z_0 \sim \mathcal{N} (\mu_{z_0}, \sigma_{z_0})$
	\State $\{z_{i}\} = \text{ODESolve}(f, z_0, (t_0...t_N))$
	\State $\tilde{x}_{i} = OutputNN(z_{i})$ for all $i=1..N$
	\State {\bfseries Return:} $\{\tilde{x}_{i}\}_{i=1..N}$
	\end{algorithmic}
\end{algorithm}




\subsection{Modelling poisson process likelihood}

To model poisson process on Physionet dataset, we augment generative ODE $\frac{d}{dt} z = f(z)$ by adding extra latent dimensions $z_{\lambda}$. Intensity function $\lambda$ is a function of latents $z_{\lambda}$:  $\lambda= g_{\lambda}(z_{\lambda})$, where $g_{\lambda}$ is two-layer feed-forward neural net. Dimensionality of $\lambda$ has to be equal to dimensionality of the data (37 in Physionet). We used 20 latent dimensions for $z$ and 20 dimensions for $z_{\lambda}$. We further augment the ODE with the integral over $\lambda$. Notice that the derivative $\int_0^{t} \lambda(\tau) d\tau$ is $\lambda(t)$. Thus, the augmented ODE is defined as follows:
 
\[    
\frac{d}{dt} \begin{bmatrix}
    z \\
    z_\lambda \\
    \int_0^{t} \lambda(\tau) d\tau
 \end{bmatrix} = \begin{bmatrix}
    f(z) \\
    f_{z_\lambda} \\
    \lambda
 \end{bmatrix}; \ \ \textnormal{where} \ \ \lambda = g(z_{\lambda})
\]

We set the initial value for $\int_0^{t} \lambda(\tau) d\tau$ to zero (notice that $\int_0^{0} \lambda(\tau) d\tau = 0$. Initial values of $z$ and $z_\lambda$ are sampled from approximate posterior. The augmented ODE is solved using a call to ODESolve.
\\ \\
\textbf{Generative model with Poisson Process} The joint generative model is specified as \\ $p(z_0)p(t_0,\dots,t_N|z_0)\prod_{i=0}^N p(x_i|z_0)$, where the distributions are specified below.

\begin{align}
p(z_0) &= \;\textnormal{Normal}\big(z_0; 0, I\big)\\
\{ z(t_i) \}_{i=0}^N &= \text{ODESolve}\big(f_\theta, z_0, (t_0, \dots, t_N)\big) \label{eq:single_call} \\
p(t_0 ... t_N | z_0) &= \;\text{PoissonProcess} \big(t_1 ... t_N ; \lambda(z(t))\big)\\
p(x_i | z_0) &= \;\textnormal{Normal}\big(x_i ; \mu(z(t_i)), \sigma(z(t_i))\big)
\end{align}

\clearpage

The Latent ODE framework specifies a generative model for time series.

\begin{align}
y_0 & \sim p(y_0) \\
\lambda(t) & = \text{ ODESolve}(y_0 , f_{\lambda}, \theta_{f_{\lambda}}, t) \\
t_1 ... t_N & \sim \text{ PoissonProcess}(\lambda(t))\\
y_1, ... y_N & = \text{ ODESolve}(y_0 , f_{y}, \theta_{f_y} , t_1 ... t_N)\\
x_i & \sim p(x|y_i , \theta_x), \ \forall i \in \{1..N\}
\end{align}

\section{Data generation and preprocessing}

\paragraph{Toy dataset}
We generate 1000 one-dimentional trajectories with 100 time points in each on the $[0,5]$ interval. We use sinusoid with fixed amplitude of 1 and sample frequency from $[0.5, 1]$ interval. We sample the starting point from $\mathcal{N}(\mu = 1, \sigma = 0.1)$.


\paragraph{MuJoCo}

We generate 10,000 simulations of the "Hopper" model from Deep Mind Control Suite and MuJoCo simulator. We sample the position of the body in 2d space uniformly from $[0, 0.5]$. We sample the relative position of limbs from $[-2, 2]$ and initial velocities from $[-5, 5]$ interval. We generate trajectories with 200 time steps for extrapolation tasks (100 points to condition on and 100 for extrapolation). We use 100 time points to perform interpolation task.

\paragraph{Physionet}

PhysioNet contains the data from the first 48 hours of patients in ICU. We have excluded four time-invariant features: Age, Gender, Height and ICUType. Hence, each patients has a set of up to 37 features (most patients have only a subset of those features). We round up the time stamps to one minute. Therefore, each time series can contain up to 2880 (60*48) points.

The original Physionet challenge has Train set (labeled) and Test set (unlabeled). We combine them into a single dataset (8000 patients, 37 features in each) and randomly split into 80\% train and 20\% test set, similarly to other datasets. We normalize each feature across all patients in the dataset to be in $[0,1]$ interval.

Each feature in each patients was measured at different time points. In order to perform batching of the features and patients during the training, we take the union of all time points across all features in a batch. We use a mask in ODE-GRU and in the recognition model of ODE-VAE to annotate the features that are present at the particular time point in order to update the latent state. Note that generative part of ODE-VAE does not require any masking. If the training example does not have any observations at the particular time point, we don't update the its hidden state at this time point.





\paragraph{Human Activity}

The original dataset contains 25 sequences from five people (6600 points on average in each sequence). For the sake of reducing the size of union of time points, we round up the time stamps of the measurement to 100 ms -- such discretization does not change the overall number of points. We join the data from four tags (belt, chest, ankles) into a single time series and then split the each sequence into partially overlapping intervals of 50 time points (with overlap of 25 points). We combine sequence from all individuals into a single dataset. After taking the union of all time points in the dataset, we get the dataset of 6554 sequences with 211 time points in each. We do not perform normalization on this dataset.

The labels are provided for each observation and denote the type of activity that the person is performing, such as walking, sitting, lying, etc. Original dataset has 11 classes. Some classes correspond to very similar activities, which are hard to distinguish. We decided to combine the classes within the following groups:  \big("lying", "lying down"\big), \big("sitting", "sitting down"\big), \big("standing up from lying", "standing up from sitting", "standing up from sitting on the ground"\big). The resulting set of classes describes seven activity types: "walking", "falling", "lying", "sitting", "standing up",  "on all fours", "sitting on the ground". We run all experiments and report results using these seven classes.

\paragraph{General notes}
For all datasets, we take the union of all time points in the dataset and run all models on the union. We randomly split the dataset into 80\% train and 20\% test. In Physionet dataset we rescale each feature to be between 0 and 1. We do not normalize other datasets. We also rescale the timeline to be in $[0,1]$.

\section{Architecture}

\paragraph{ODE function}

We use a feed-forward neural net for ODE function $f$. See section \ref{sec:hyper} for sizes of the network in each experiment.

We used Tanh activations in ODE function. Tanh activation constrains the output and prevents the ODE gradients from taking large values. If values of ODE gradient are too big, it might be hard to solve an ODE with the specified tolerance. For this reason, we do not recommend using ReLU.

\paragraph{ODE Solver}
We used ODE Solvers from torchdiffeq python package. We used fifth-order ''dopri5'' solver with adaptive step for generative model of Latent ODE.  We used relative tolerance of 1e-3 and absolute tolerance of 1e-4. Adjoint method described in Chen et al. (2018) can be used to reduce the memory use, at a cost of a longer computation time.


\paragraph{Loss} To compute data log-likelihood, We use negative gaussian log-likelihood with fixed variance as a reconstruction loss. We used fixed variance of 0.001 for MuJoCo and 0.01 for other datasets. We report MSE on the time series in the test set. For classification task, we use cross-entropy loss. To compute ELBO for encoder-decoder models, we used three samples from distribution $\mathcal{N} (\mu_{z_0}, \sigma_{z_0})$.

On Physionet dataset, we got best performance by training reconstruction loss and cross-entropy loss together, with multiplier 100 on cross-entropy loss. We train on the whole dataset of 8000 patients and compute the CE loss only on labeled samples (4000 patients).

On Human Activity dataset, we trained solely with cross-entropy loss. Note that on Physionet labels are provided per time series, and reconstruction loss prevents overfitting on classification task. On Human Activity dataset, the labels are provided per time point and training with CE loss only does not lead to overfitting.

\paragraph{RNN Baselines}

We follow the available implementation of RNN GRU-D:
\url{https://github.com/zhiyongc/GRU-D/blob/master/GRUD.py}. We use exponential decay between hidden states and imputation technique from GRU-D as separate baselines.

\section{Hyperparameters}
\label{sec:hyper}

\paragraph{Choosing hyperparameters}
First, we choose the hyperparameters such that delivered the best performance for the \emph{RNN baselines} in our experiments. Then we run corresponding ODE models with the same hyperparameters. In autoregressive models we use the same size of hidden state, number of layers and units in encoder and decoder networks, etc. in both ODE-RNN and RNN baselines.


\paragraph{Hyperparameters for Latent ODE model}
We empirically found that ODE of the same or slightly smaller dimensionality as the data works best for generative model in Latent ODE. We also found that recognition model has to have bigger dimensionality than generative model. As such, we used 15-dimensional latent state in generative model and 30-dim recognition model for 14-dimensional Mujoco dataset. For 37-dimensional physionet dataset we used 20 latent dimensions in the generative model and 40 in recognition model. For Human activity classification task (12 data dimentions), we used 15 latent dimensions in generative model and 100 in recognition model.

\paragraph{Toy dataset}

To model 1-dimensional toy dataset, we used 10 latent dimensions in the generative model and 20 in recognition model, batch size of 50. ODE function for both generative and recognition models consist of 1 hidden layer with 100 units.

\paragraph{MuJoCo}

In encoder-decoder models, we used 15 latent dimensions in generative model, 30 dimensions in recognition model, batch size of 50. ODE functions had 3 layers and 500 units. We used 15-dimensional hidden state in autoregressive models.

\paragraph{Physionet (interpolation and extrapolation)}

In encoder-decoder models, we used 20 latent dimensions in the generative model, 40 dimensions in recognition model and batch size of 50. ODE function have 3 layers with 50 units. We used 20-dimensional hidden state in autoregressive models.

\paragraph{Fitting Poisson process on Physionet dataset}

For fitting Poisson process along with the likelihood together with reconstruction likelihood, our generative model had the following dimensions: 20 for modelling $z(t)$ (for reconstruction), 20 for modelling $z_\lambda$ (for poisson process) and 37 for $\int \lambda_0^t(\tau) d\tau$ ( equals data dimensionality). Hence, 77 dimensions in total. Other hyperparameters are the same as above.


\paragraph{Classification on Physionet}

In classification task, the goal is to classify each patient with a binary label. We used "in-hospital mortality" label from the original dataset. To prevent overfitting, we train the model together with reconstruction loss with coefficient 100 on cross-entropy loss. We compute reconstruction loss on all 8000 patients, and cross-entropy loss on 4000 labeled patients. We used a 2-layer classifier with 300 units and ReLU activations.

\textbf{Autoregressive models} We used 10-dimensional hidden states, batch size 50 and learning rate 0.01. For ODE function in ODE-RNN, we used a 3-layer neural net with 50 units and Tanh activation.

\textbf{Encoder-decoder models}

We used 20-dimensional latent state in generative model, 40-dimensional hidden state in recognition model, batch size 50 and learning rate 0.01. For ODE function, we used a 3-layer neural net with 30 units and Tanh activation.

\paragraph{Classification on Human Activity} On Human Activity dataset, the task is to classify each time point by the type of activity. We used a linear classifier on each hidden state $h_i$.

\textbf{Autoregressive models} We used 30-dimensional hidden state, batch size 100 and learning rate 0.01. For ODE function, we used a neural net with four hidden layers and 1000 units. 

\textbf{Encoder-decoder models} In the generative model, we used 15-dimensional latent state. Generative ODE is a 2-layer neural nets with 500 units. In recognition model, we used 100-dimensional hidden state and 4-layer neural net with 500 units for ODE function. We used batch size of 100 and learning rate 0.01.

\section{Training details}

We used Adamax optimizer with the learning rate of 0.01. We use a small learning rate decay rate of 0.999. We use KL annealing for VAE models with coefficient 0.99.

\subsection{Computing infrastructure}

All experiments were run on one Nvidia P100 with 2 physical Intel Xeon(R) Silver 4110 CPU.

\subsection{Source code and datasets}

The code for generating and pre-processing of the datasets is included with the submission. The model is implemented in PyTorch 1.0. 

We used the ODE solvers from torchdiffeq package (\url{https://github.com/rtqichen/torchdiffeq})
\ \\

\textbf{Datasets}:

Human Activity: \url{https://archive.ics.uci.edu/ml/datasets/Localization+Data+for+Person+Activity}

Physionet: 
\url{https://physionet.org/physiobank/database/challenge/2012/}

MuJoCo: \url{http://www.cs.toronto.edu/~rtqichen/datasets/HopperPhysics/training.pt}. 

MuJoCo dataset was created using DeepMind Control Suite: \url{https://github.com/deepmind/dm_control}.

\clearpage

\section*{Supplementary figures and tables}

\begin{figure}[h]
	\centering
    \begin{subfigure}[b]{\columnwidth}
    	\centering
        \includegraphics[width=\textwidth]{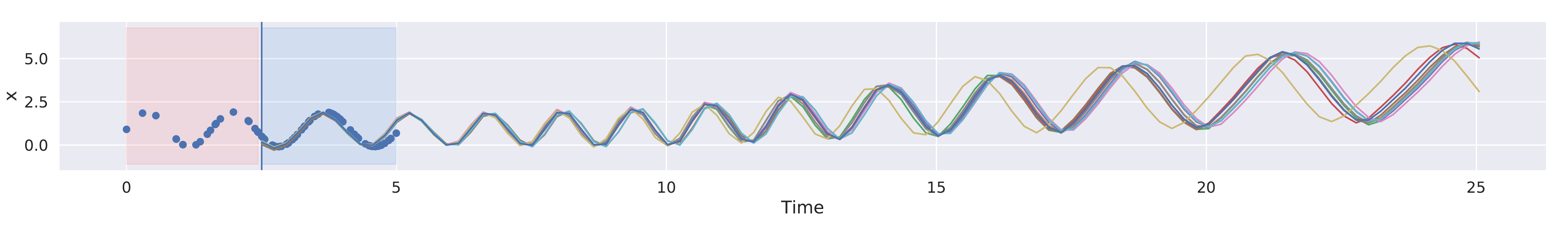}
        \caption{Latent ODE with ODE encoder (ours): conditioned on 20 points}
    \end{subfigure}
    \begin{subfigure}[b]{\columnwidth}
    	\centering
        \includegraphics[width=\textwidth]{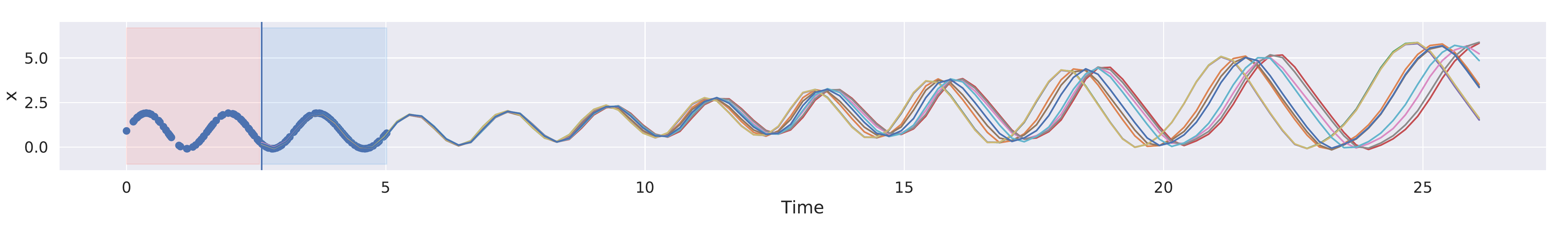}
        \caption{Latent ODE with ODE encoder (ours): conditioned on 80 points}
    \end{subfigure}
    \caption{}
\end{figure}

\begin{figure}[h]
	\centering
    \begin{subfigure}[b]{\columnwidth}
    	\centering
        \includegraphics[width=\textwidth]{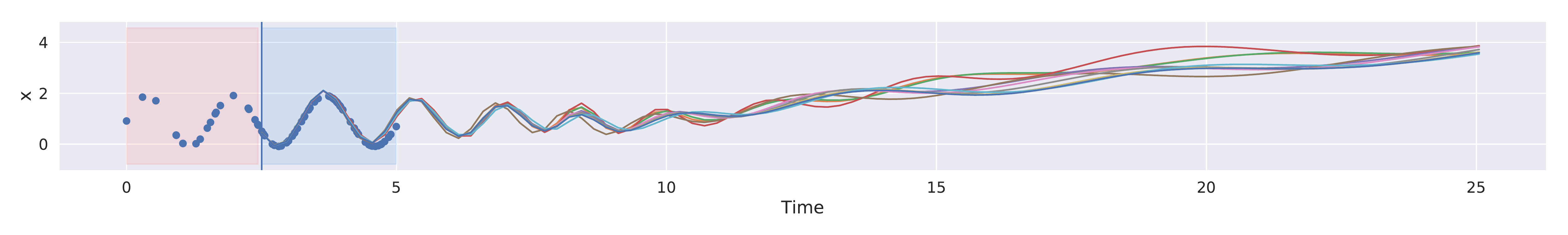}
        \caption{Latent ODE with RNN encoder (Chen et al. 2018): conditioned on 20 points}
    \end{subfigure}
    \begin{subfigure}[b]{\columnwidth}
    	\centering
        \includegraphics[width=\textwidth]{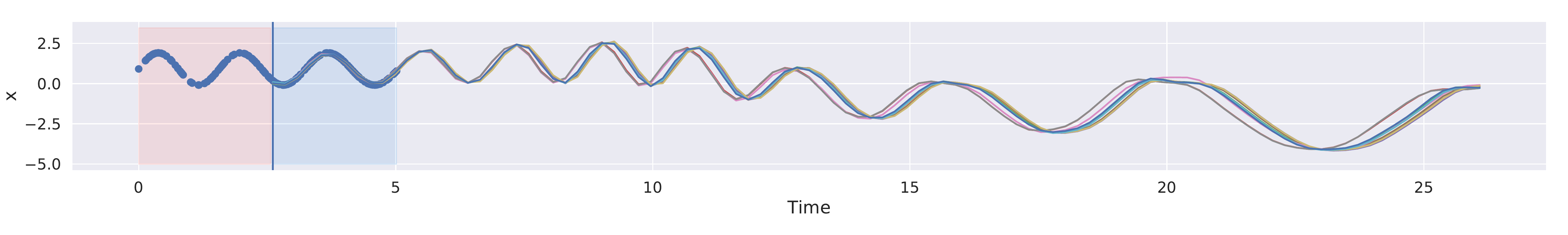}
        \caption{Latent ODE with RNN encoder (Chen et al. 2018): conditioned on 80 points}
    \end{subfigure}
    \caption{}
\end{figure}

\begin{figure}
\centering
    \begin{subfigure}[b]{0.45\columnwidth}
    	\centering
        \includegraphics[width=0.7\columnwidth]{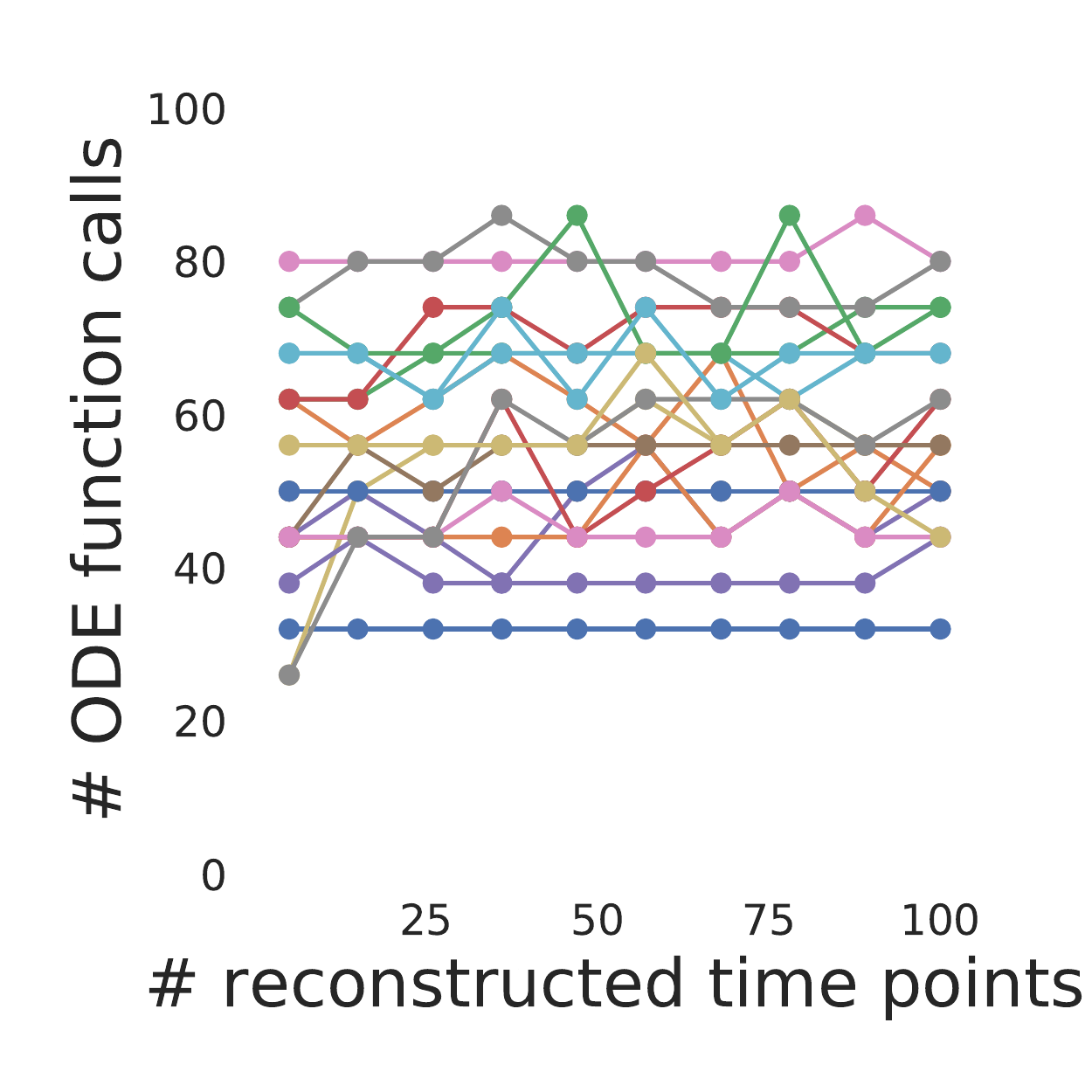}
        \caption{}
    \end{subfigure}
    \begin{subfigure}[b]{0.45\columnwidth}
    	\centering
        \includegraphics[width=0.7\columnwidth]{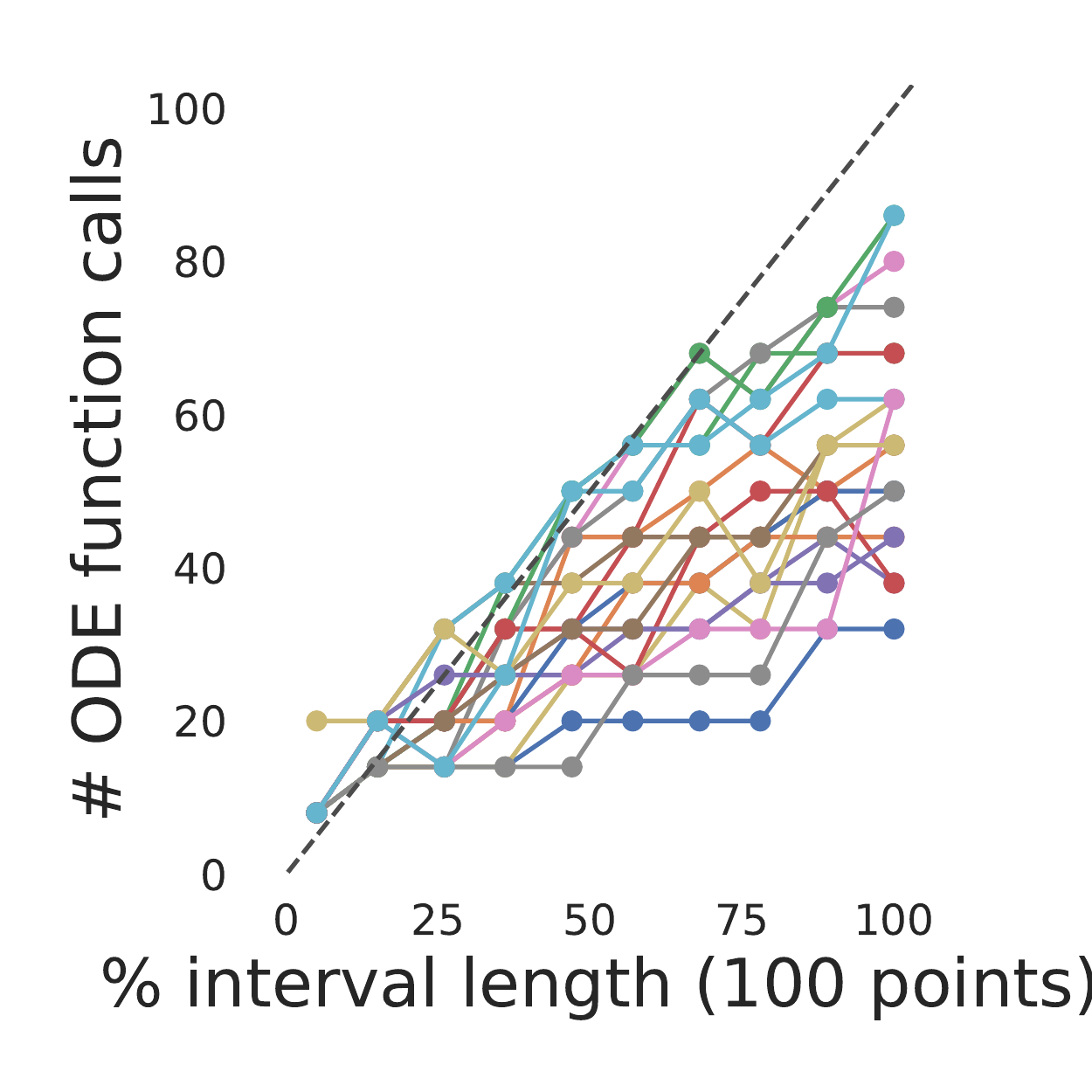}
        \caption{}
    \end{subfigure}
	\caption{(a) Number of evaluations of ODE function $f$ does not depend on the number of time points ($t_0$..$t_N$) where we evaluate ODE solution (b) Instead, number of ODE function evaluations depends on the length of the time interval $[t_0..t_N]$ where ODE is solved. \\
	For plot (a) we randomly subsampled time series in MuJoCo dataset by variable number of points and computed number of function evaluations required by ODESolve in each case. For plot (b) we truncated the time series to ($t_0$..$t_i$) and computed number of ODE function evaluations for different $i$. Each line shows a number of ODE func evals for a single time series from MuJoCo dataset.}
	\label{fig:func_evals}
\end{figure}

\begin{figure*}
	\centering
	\begin{subfigure}[b]{0.24\linewidth}
		\centering
	\includegraphics[width=\textwidth]{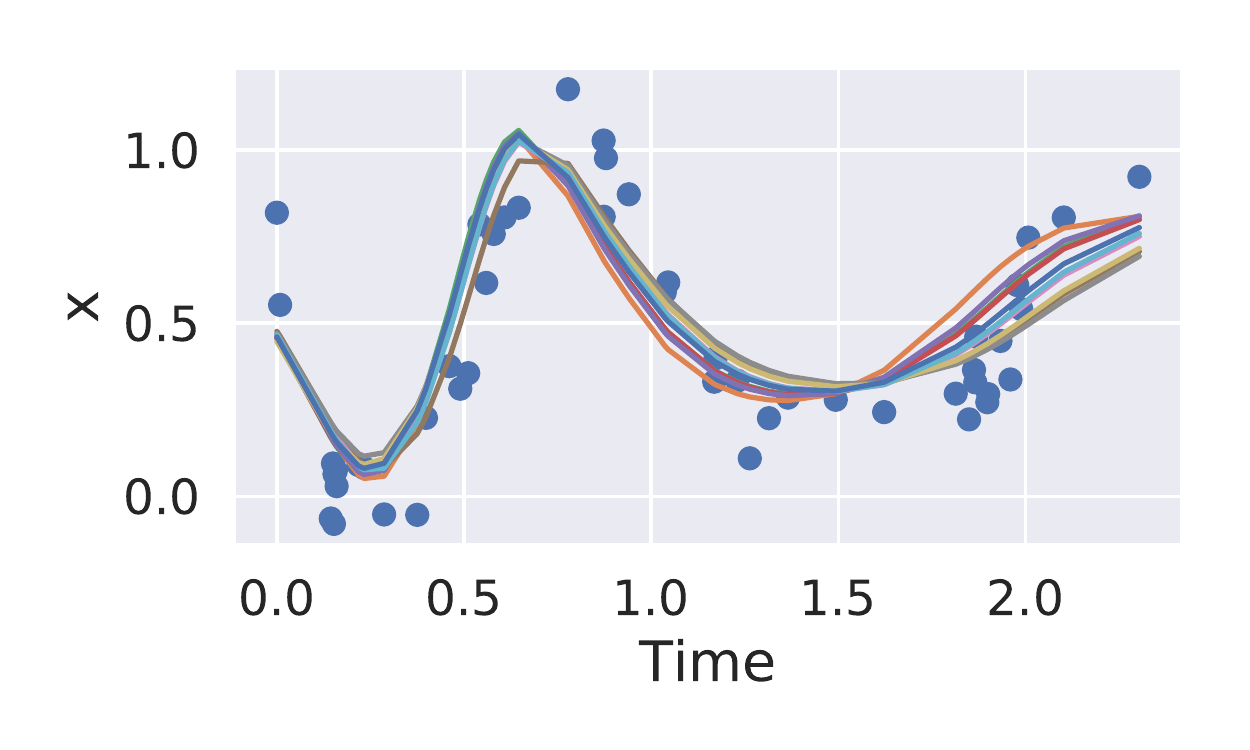}
    \end{subfigure}
    \begin{subfigure}[b]{0.24\linewidth}
    	\centering
        \includegraphics[width=\textwidth]{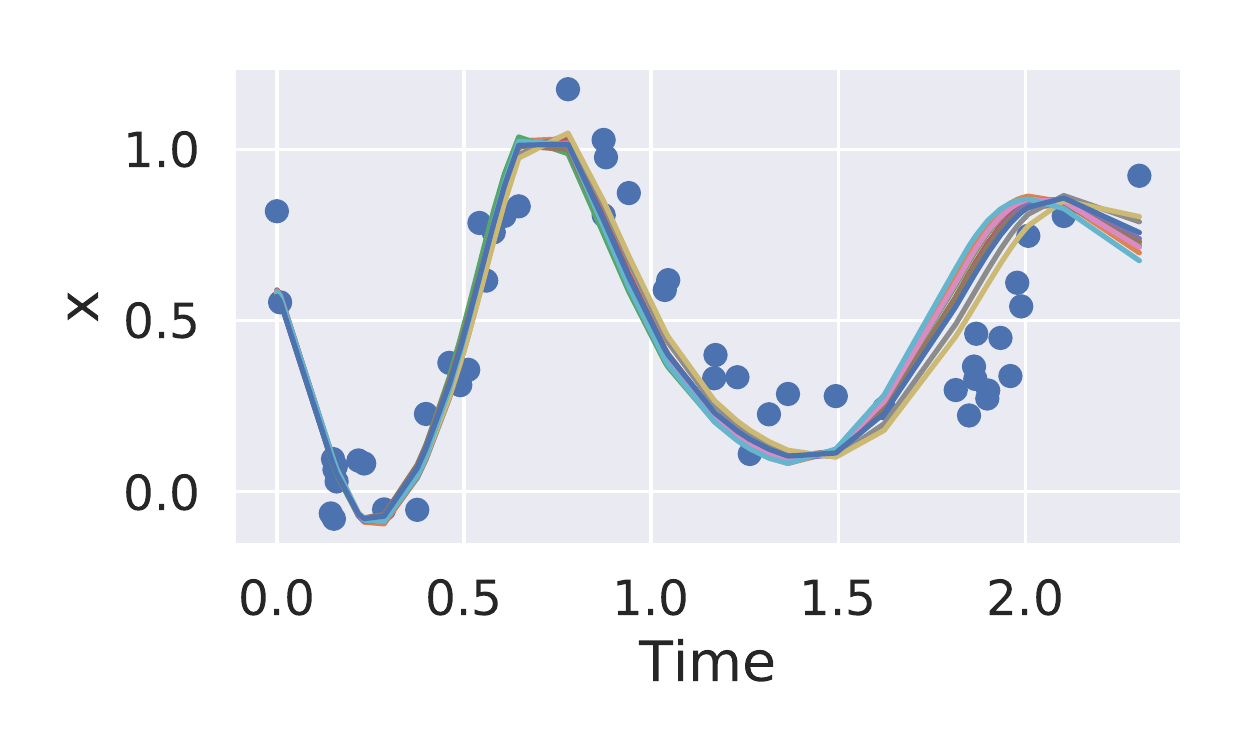}
    \end{subfigure}
    \begin{subfigure}[b]{0.24\linewidth}
    	\centering
        \includegraphics[width=\textwidth]{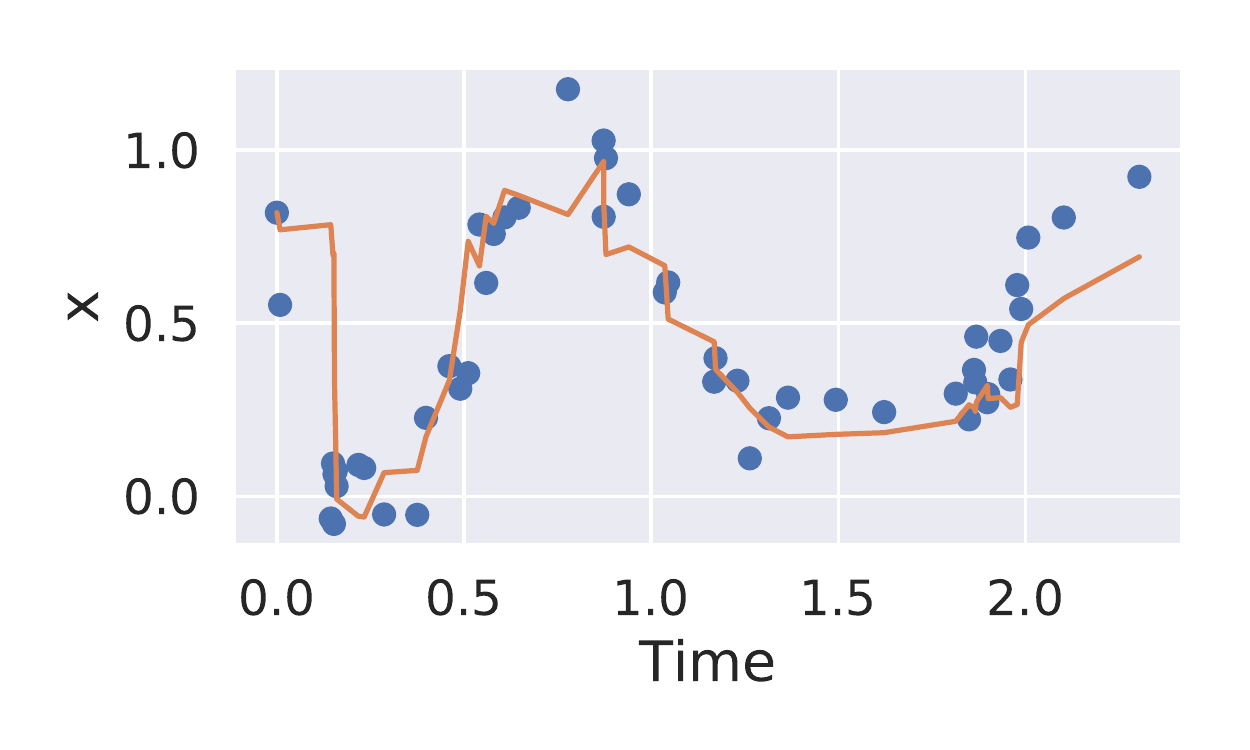}
    \end{subfigure}
    \begin{subfigure}[b]{0.24\linewidth}
    	\centering
        \includegraphics[width=\textwidth]{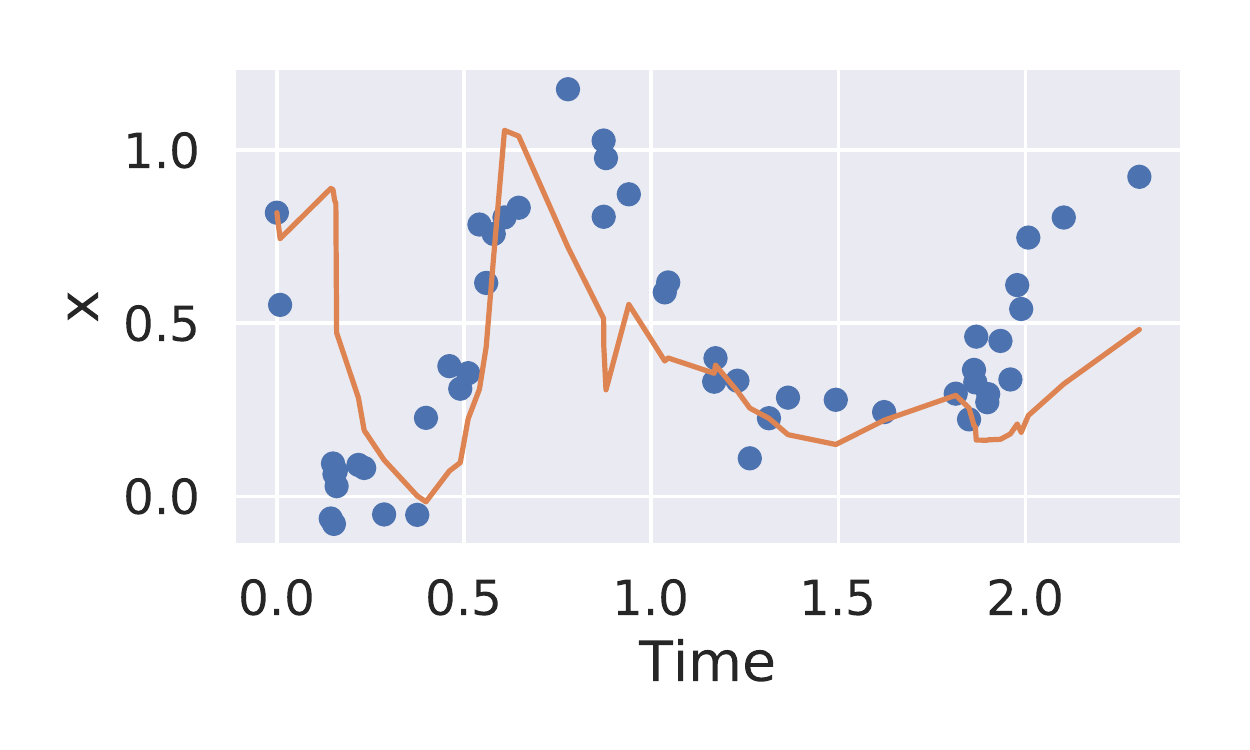}
    \end{subfigure}
    \begin{subfigure}[b]{0.24\linewidth}
		\centering
	\includegraphics[width=\textwidth]{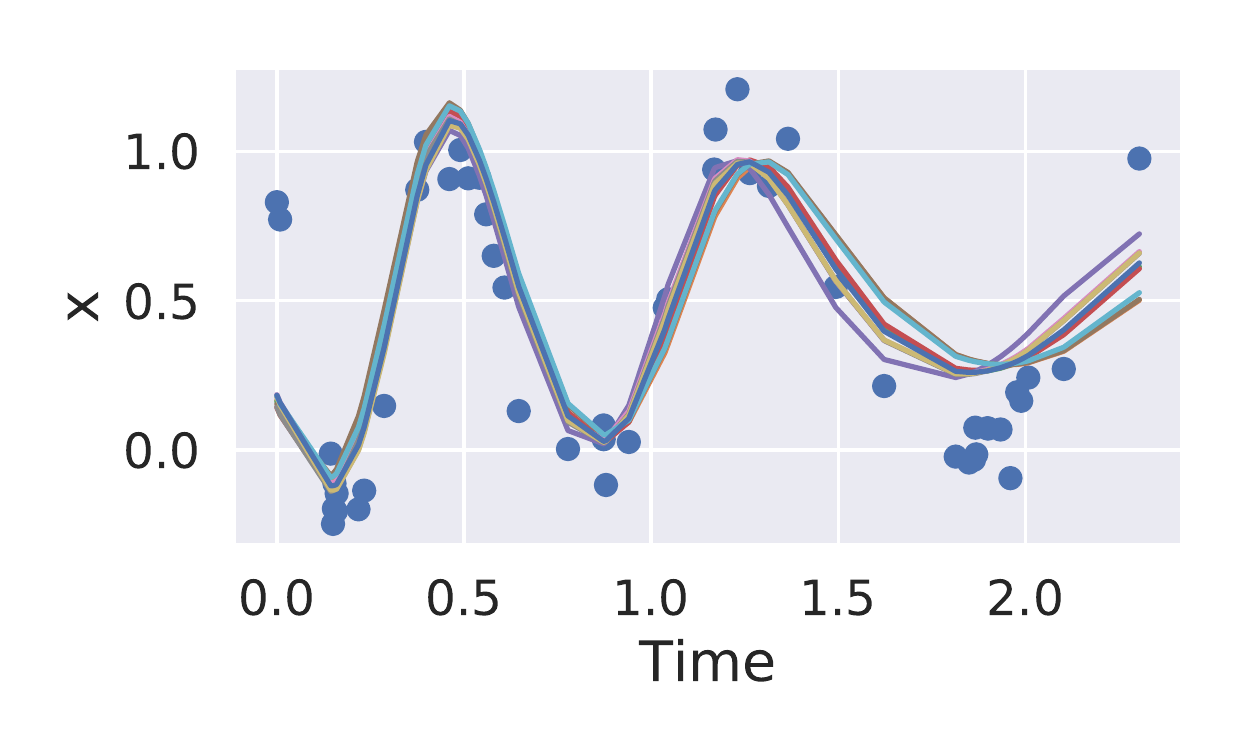}
    \caption{Latent ODE (ODE enc)}
    \end{subfigure}
    \begin{subfigure}[b]{0.24\linewidth}
    	\centering
        \includegraphics[width=\textwidth]{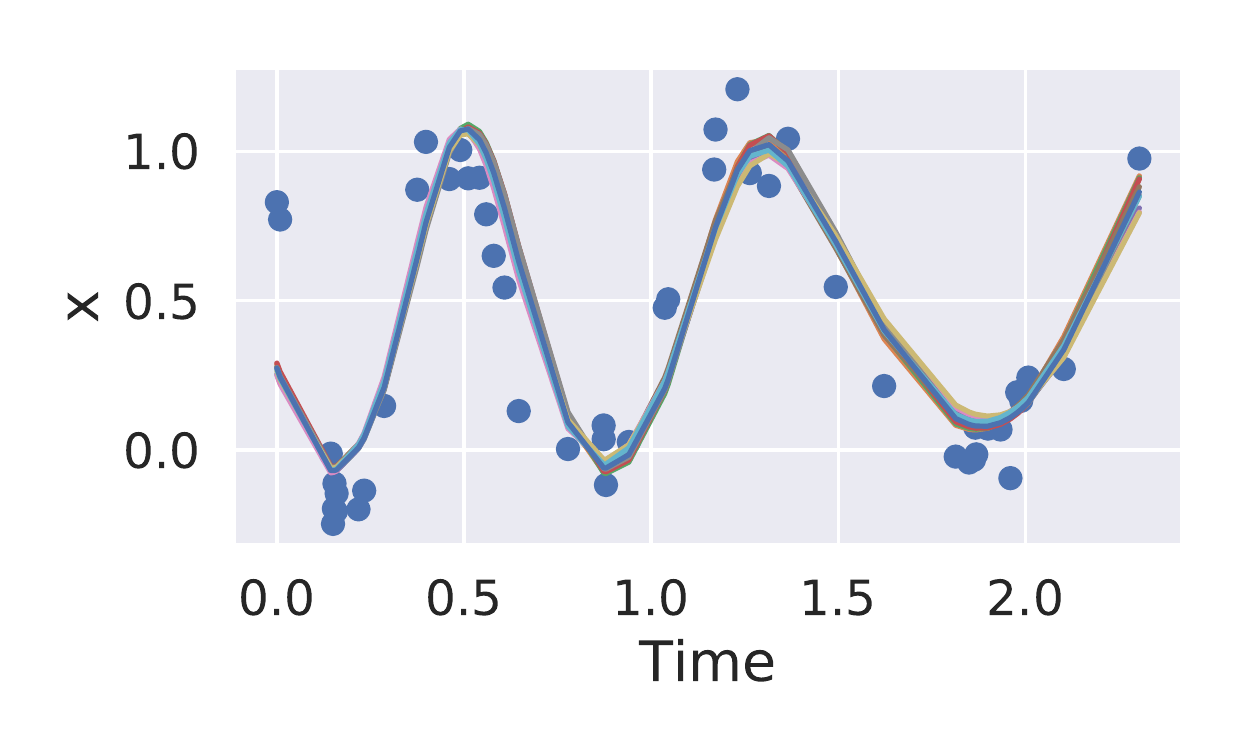}
         \caption{Latent ODE (RNN~enc)}
    \end{subfigure}
    \begin{subfigure}[b]{0.24\linewidth}
    	\centering
        \includegraphics[width=\textwidth]{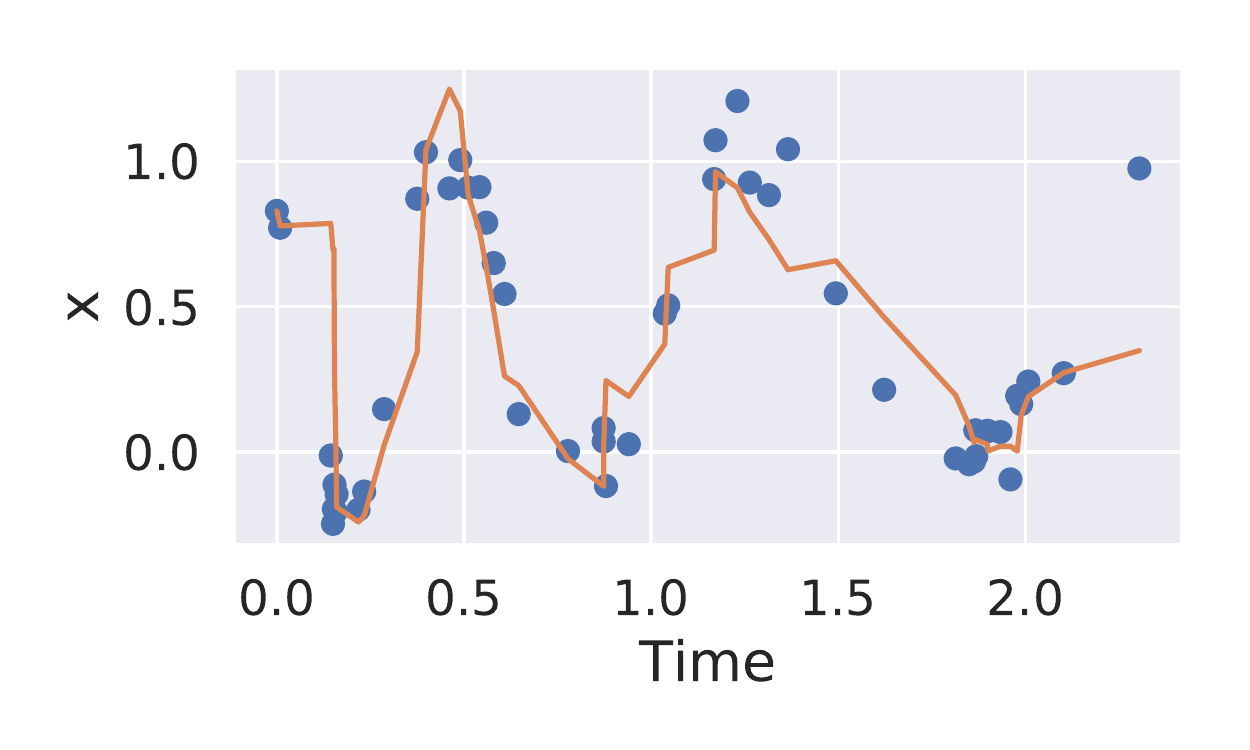}
       \caption{ODE-RNN}
    \end{subfigure}
    \begin{subfigure}[b]{0.24\linewidth}
    	\centering
        \includegraphics[width=\textwidth]{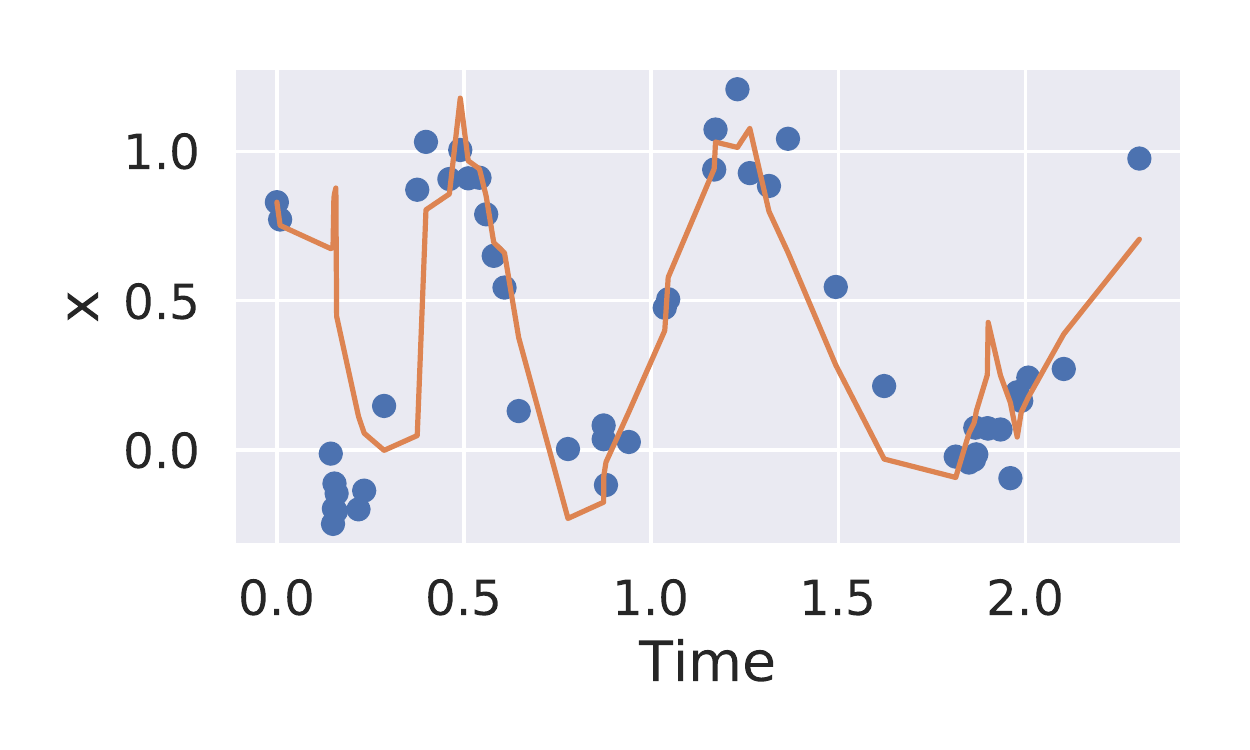}
       \caption{Standard RNN}
    \end{subfigure}
     \caption{Reconstructions on toy dataset}
    \label{fig:1d_data}
\end{figure*}

\begin{figure*}
	\centering
	\begin{subfigure}[b]{0.49\columnwidth}
		\centering
    	\includegraphics[width=0.5\textwidth]{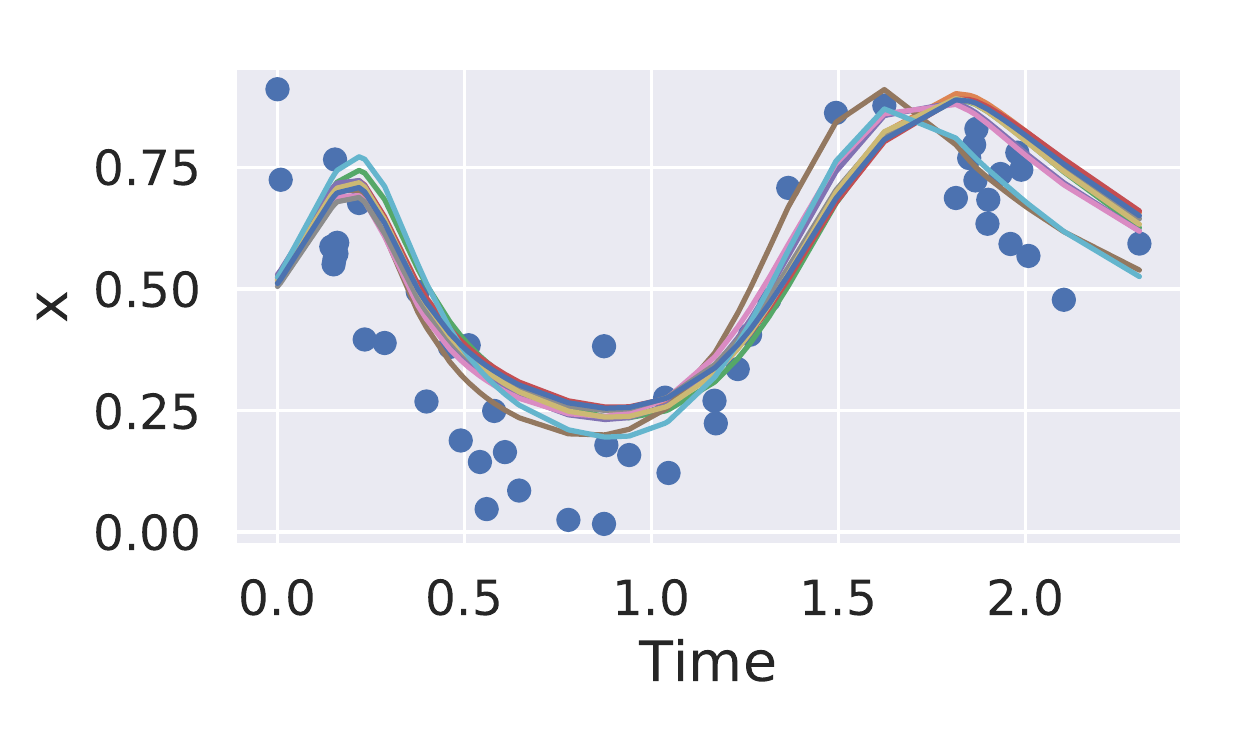}
    \end{subfigure}
	\begin{subfigure}[b]{0.49\columnwidth}
		\centering
    	\includegraphics[width=0.5\textwidth]{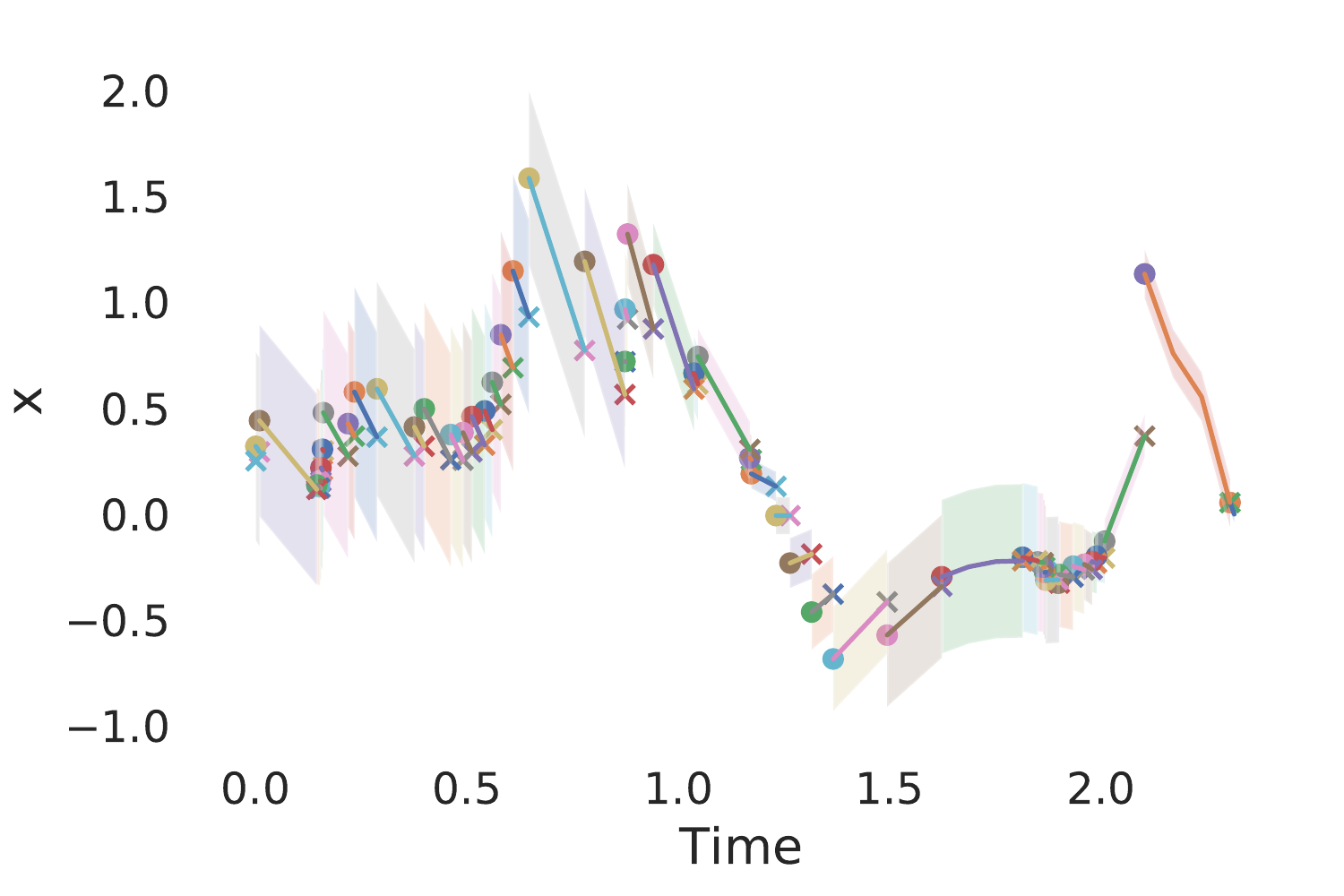}
    \end{subfigure}
\begin{subfigure}[b]{0.49\columnwidth}
		\centering
    	\includegraphics[width=0.5\textwidth]{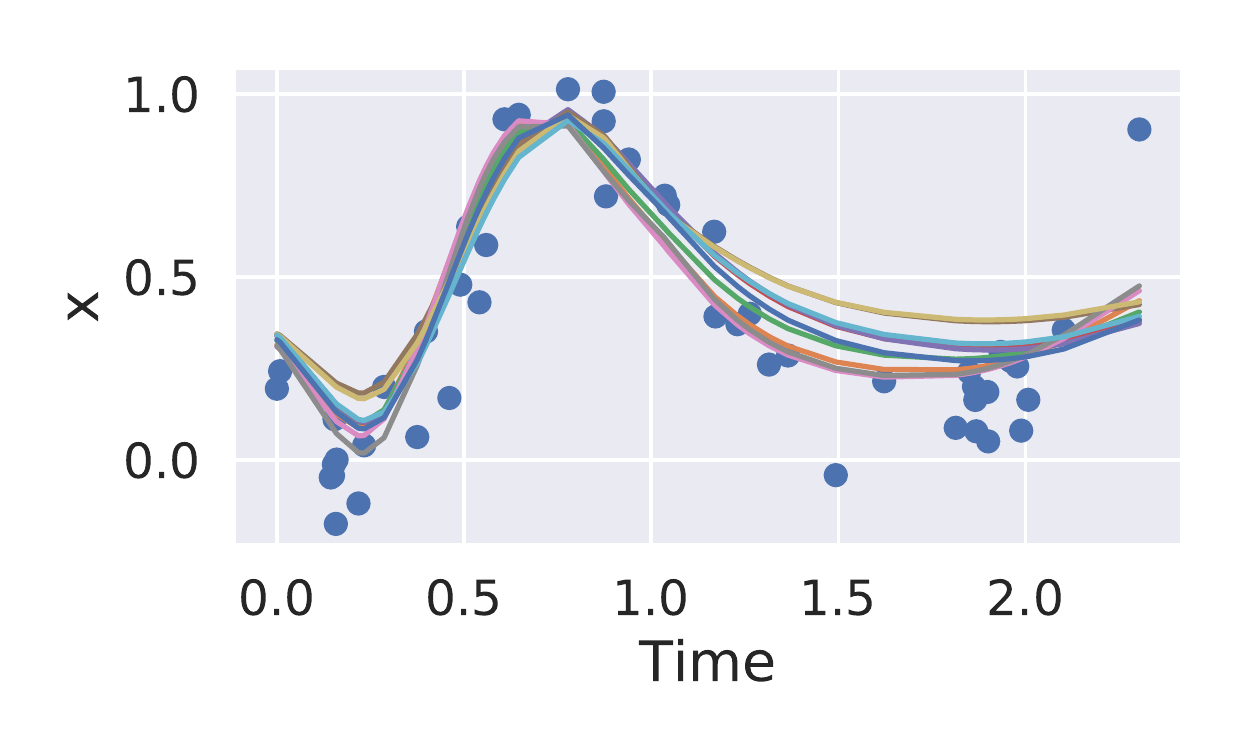}
    \end{subfigure}
	\begin{subfigure}[b]{0.49\columnwidth}
		\centering
    	\includegraphics[width=0.5\textwidth]{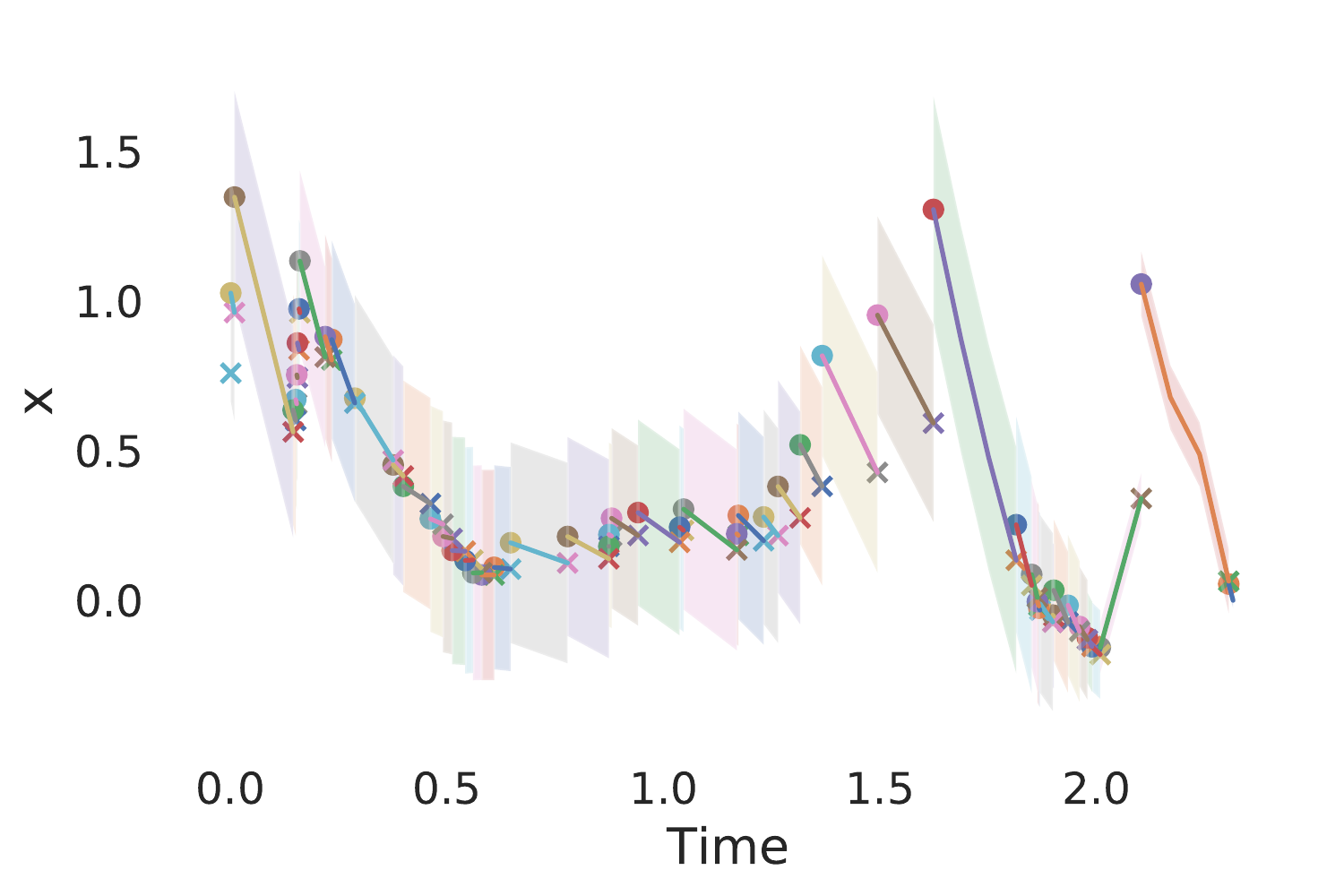}
    \end{subfigure}
	\begin{subfigure}[b]{0.49\columnwidth}
		\centering
    	\includegraphics[width=0.5\textwidth]{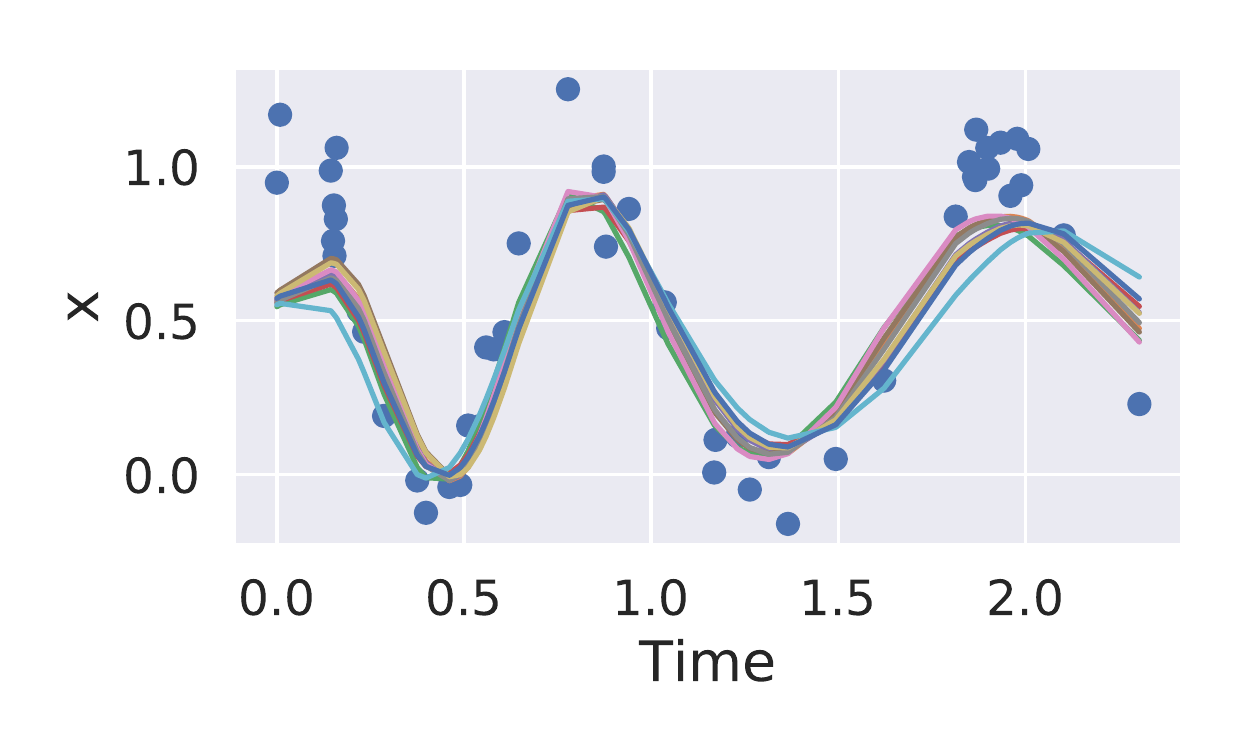}
    \end{subfigure}
	\begin{subfigure}[b]{0.49\columnwidth}
		\centering
    	\includegraphics[width=0.5\textwidth]{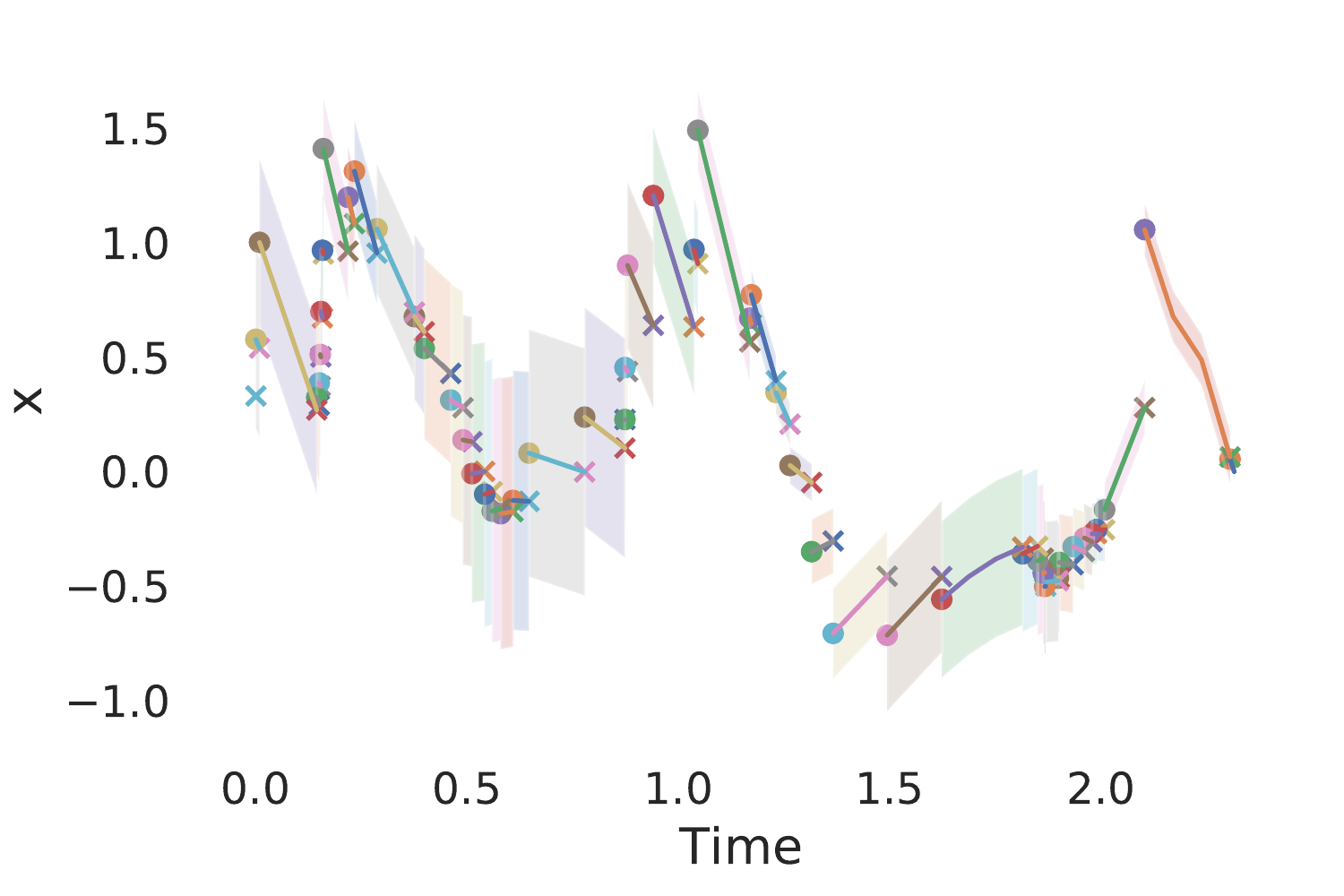}
    \end{subfigure}
    \label{fig:1d_data}
	\begin{subfigure}[b]{0.49\columnwidth}
		\centering
    	\includegraphics[width=0.5\textwidth]{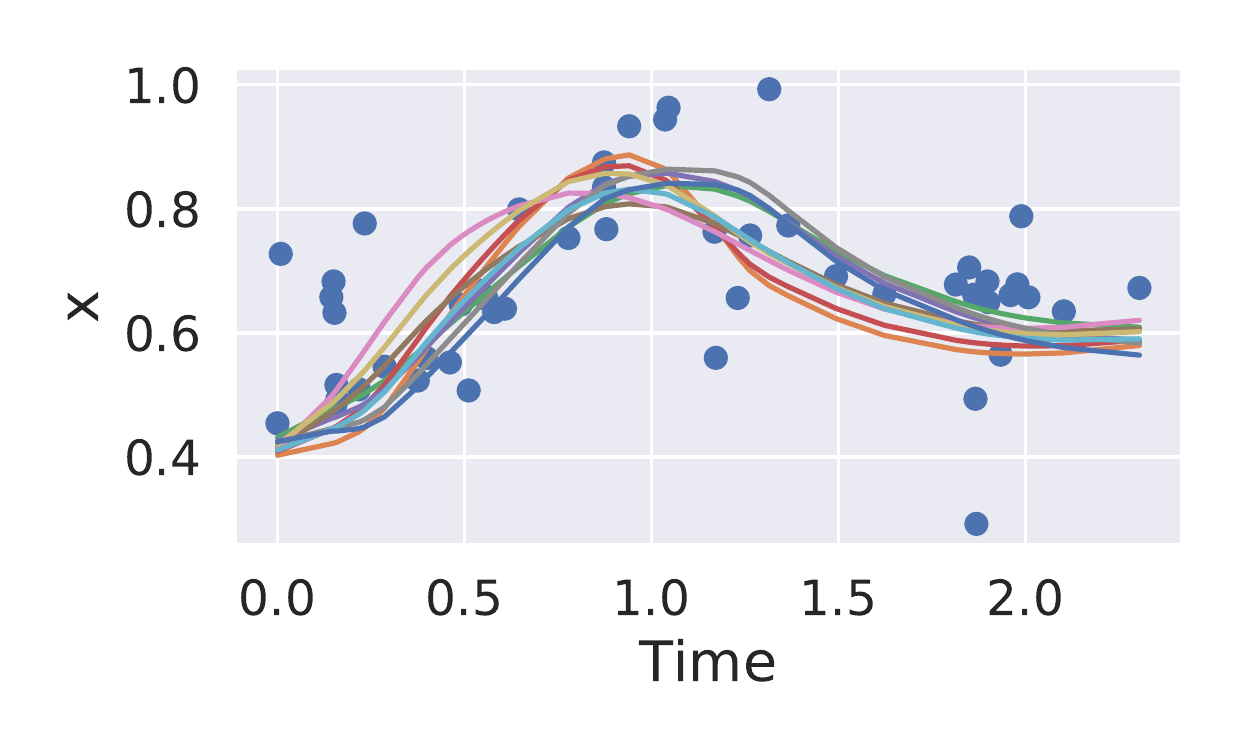}
    	\caption{}
    \end{subfigure}
	\begin{subfigure}[b]{0.49\columnwidth}
		\centering
    	\includegraphics[width=0.5\textwidth]{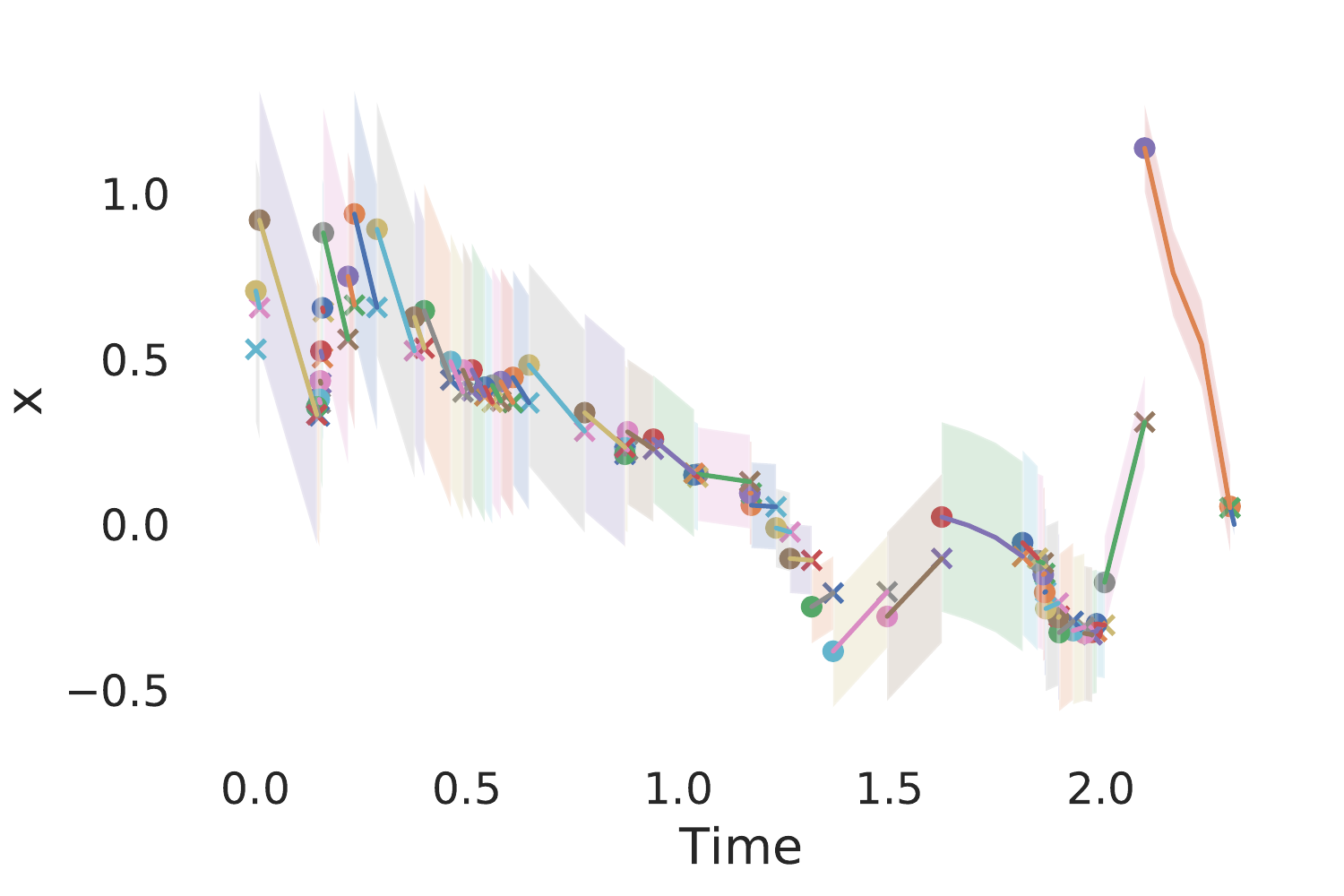}
    	\caption{}
    \end{subfigure}
     \caption{\textbf{(a)} Reconstructions on toy dataset from Latent ODE. Observations are shown as points. Lines are reconstructions for different samples of $z_0$ in Latent ODE model. 
     \textbf{(b)} Corresponding latent state in the recognition model (first dimension). The recognition model encodes the data backwards in time (from right to left). The lines show latent ODE path in-between the encoded data points. The discontinuities between the paths show the update of the latent state using the observation at that time point. The end of the end of the ODE path from the previous observation is shown as a small circle. The updated state (and the start a new ODE path) is shown as cross. 
     The shaded area shows the predicted standard deviation for the initial latent state $z_0$.
     Notice that the right-most point is similar for all four trajectories -- only one data point was encoded, which does not contain much information about the trajectory. As the encoding progresses (from right to left), the latent updated generally become smaller.}
    \label{fig:1d_data}
\end{figure*}

\newcommand{\postwidth}{0.143\linewidth}%
\begin{figure}[ht]
	\centering
	 \begin{subfigure}[b]{0.1\columnwidth}
    	\centering
        without poisson
        \vspace{6mm}
    \end{subfigure}
    \begin{subfigure}[b]{\postwidth}
    	\centering
        \includegraphics[width=0.97\textwidth]{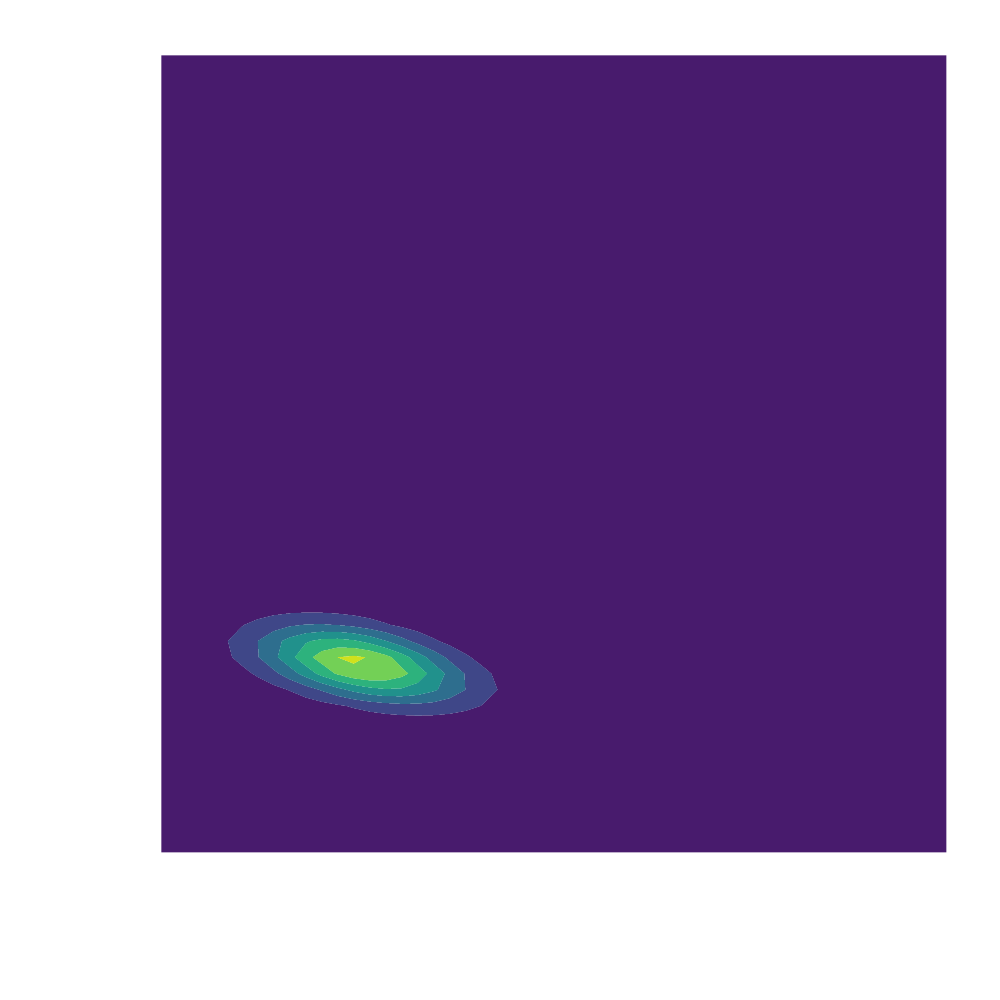}
    \end{subfigure}
    \begin{subfigure}[b]{\postwidth}
    	\centering
        \includegraphics[width=0.97\textwidth]{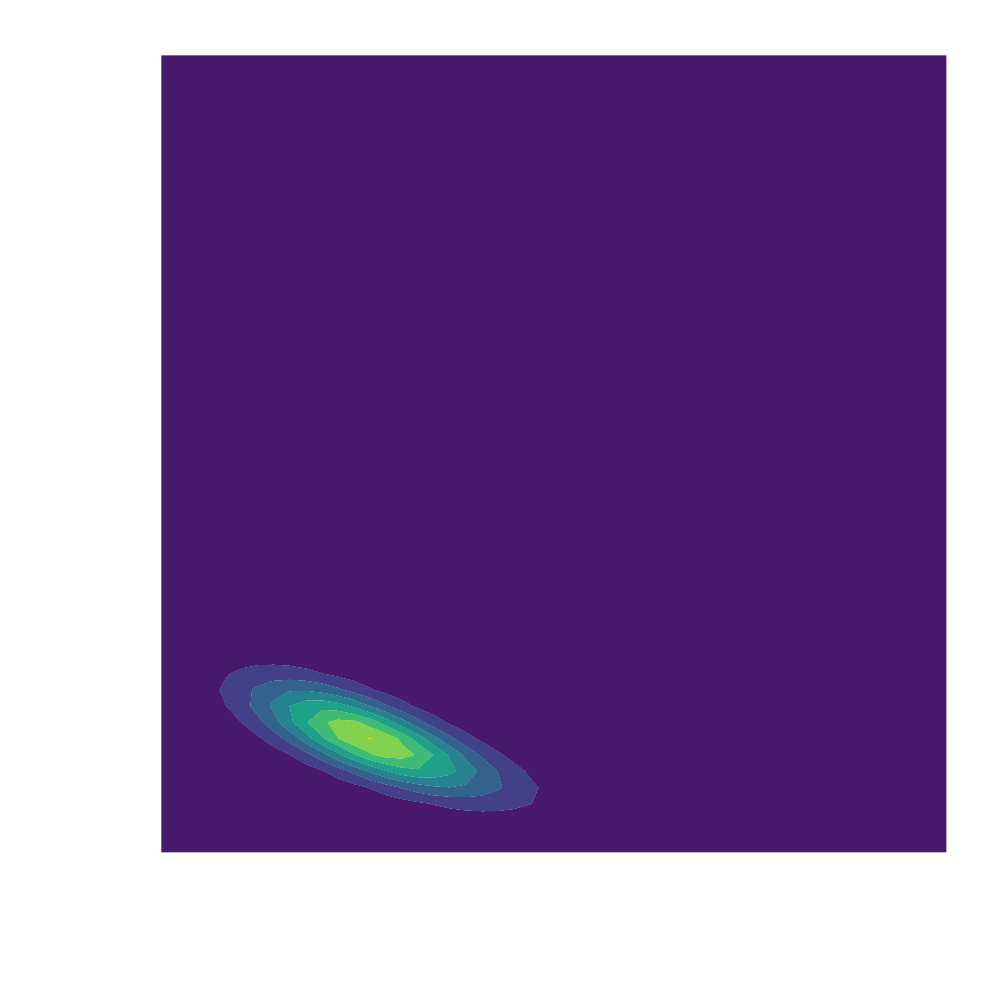}
    \end{subfigure}
    \begin{subfigure}[b]{\postwidth}
    	\centering
        \includegraphics[width=0.97\textwidth]{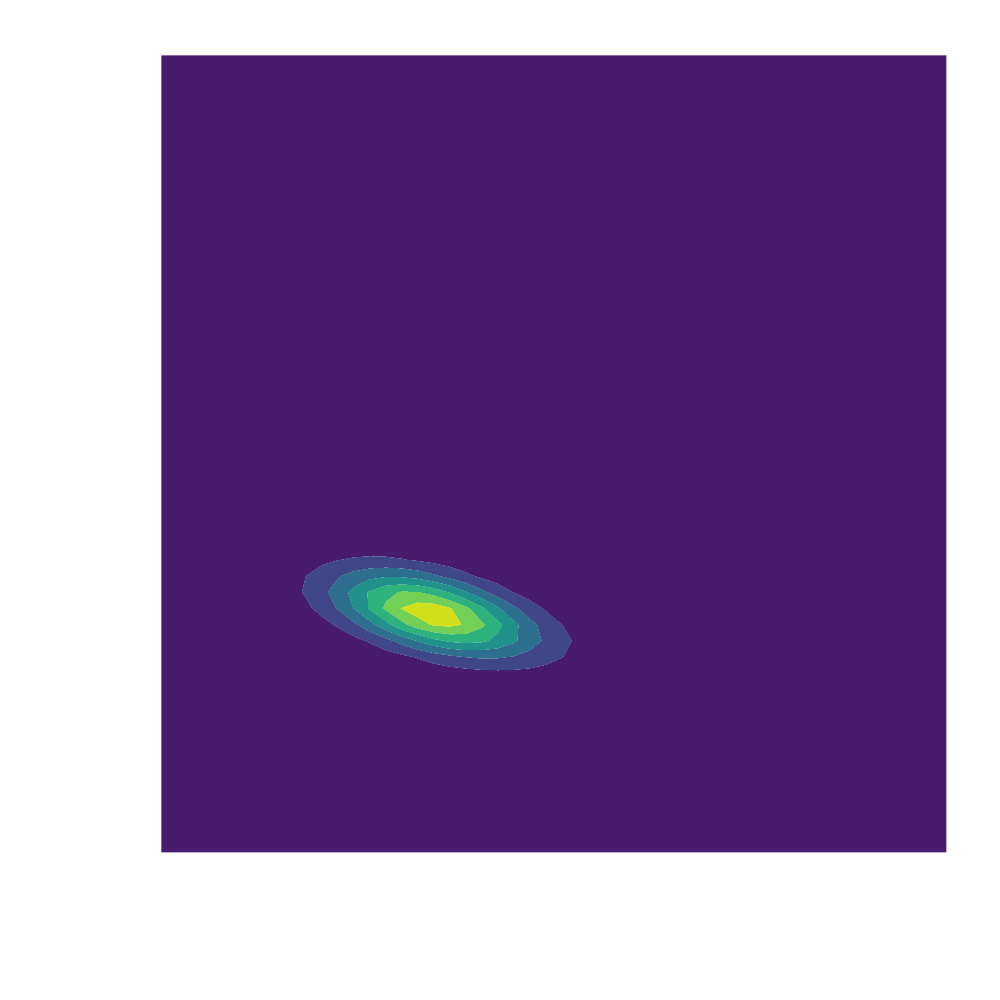}
    \end{subfigure}
    \begin{subfigure}[b]{\postwidth}
    	\centering
        \includegraphics[width=0.97\textwidth]{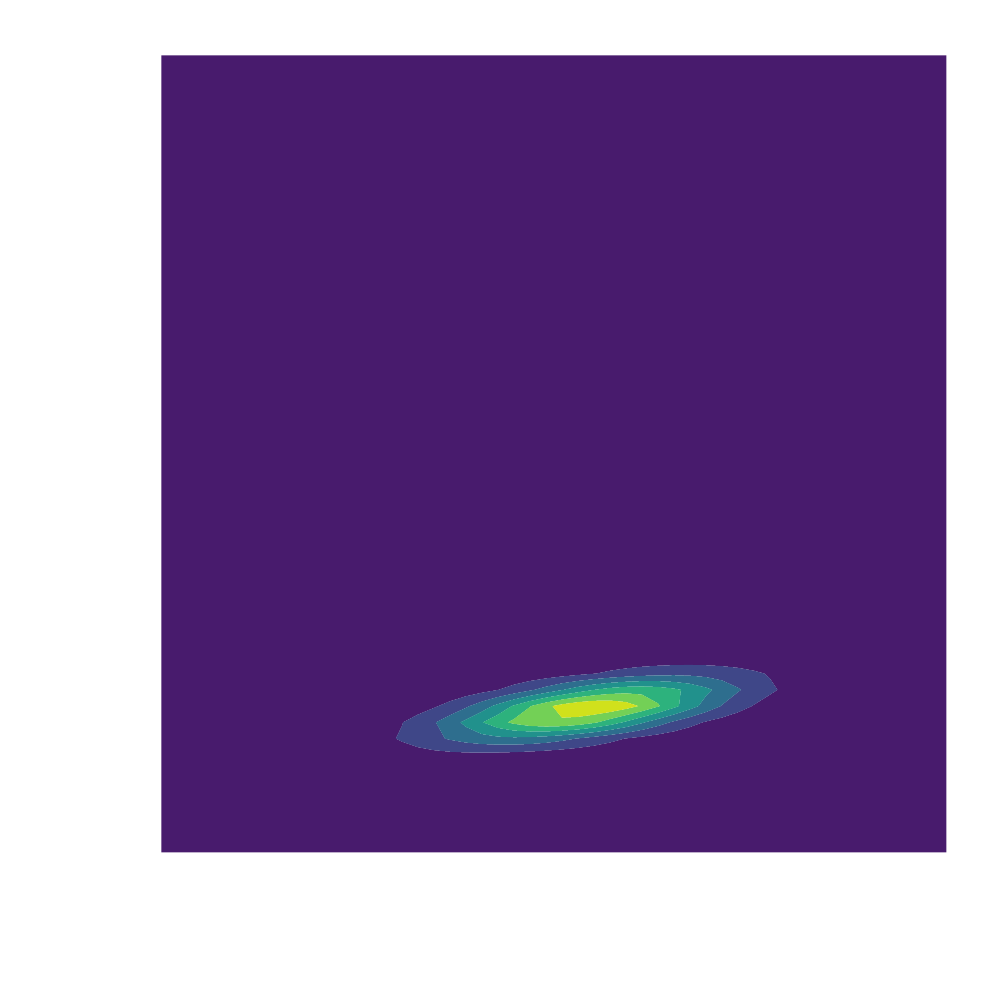}
    \end{subfigure}
    \begin{subfigure}[b]{\postwidth}
    	\centering
        \includegraphics[width=0.97\textwidth]{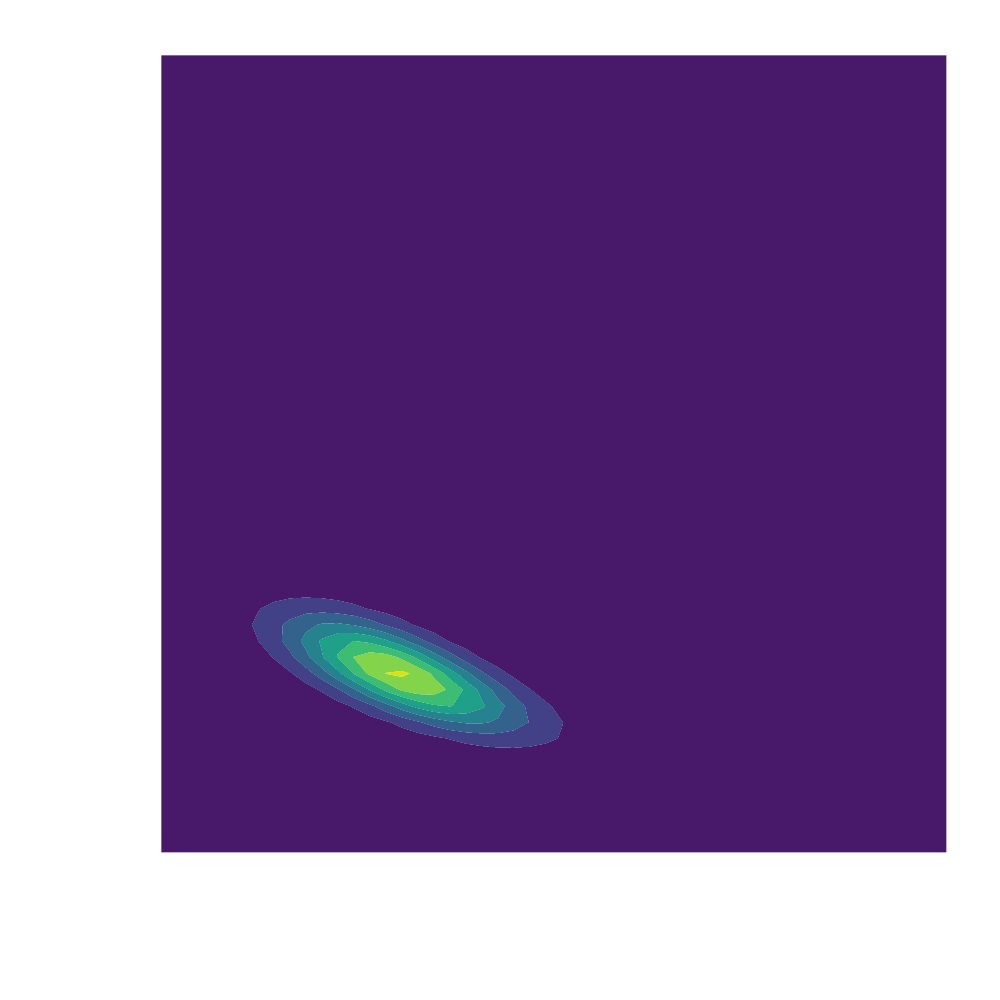}
    \end{subfigure}
    \begin{subfigure}[b]{\postwidth}
    	\centering
        \includegraphics[width=0.97\textwidth]{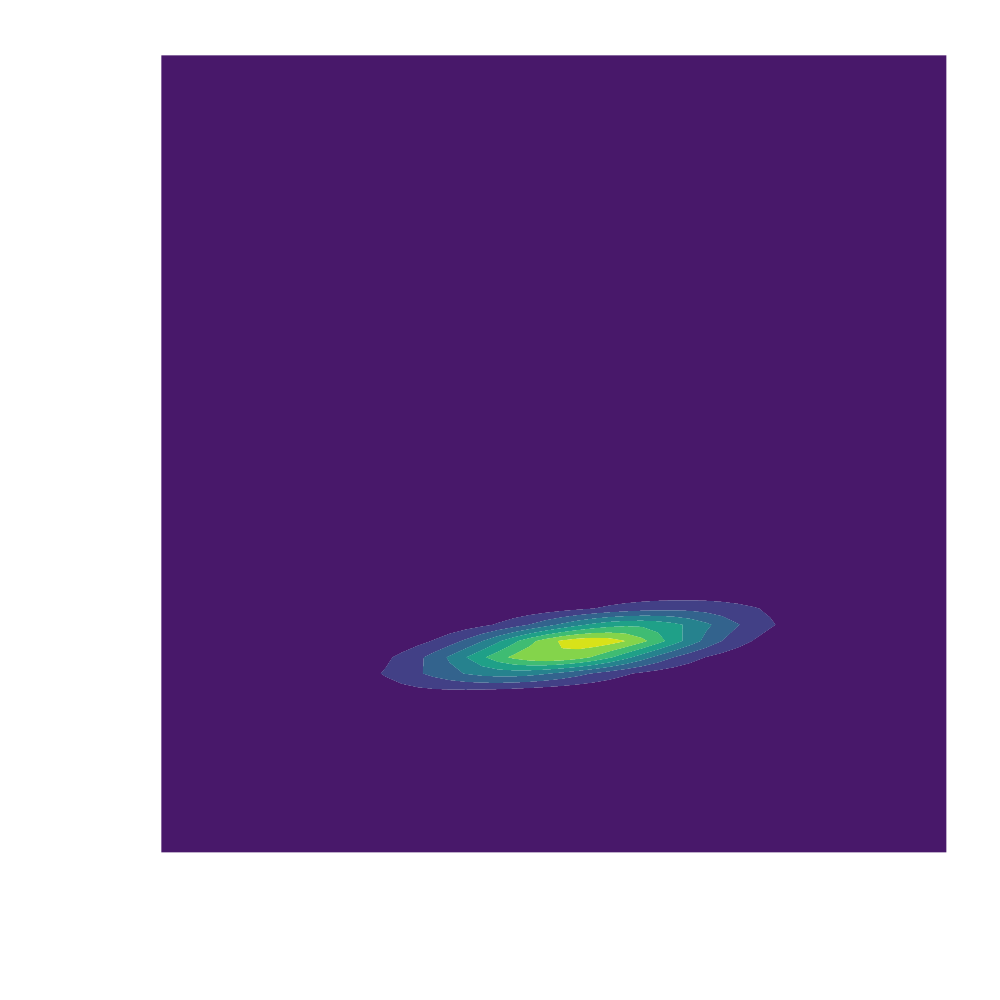}
    \end{subfigure}
	\begin{subfigure}[b]{0.1\linewidth}
    	\centering
        with poisson
        \vspace{6mm}
    \end{subfigure}
    \begin{subfigure}[b]{\postwidth}
    	\centering
        \includegraphics[width=0.97\textwidth]{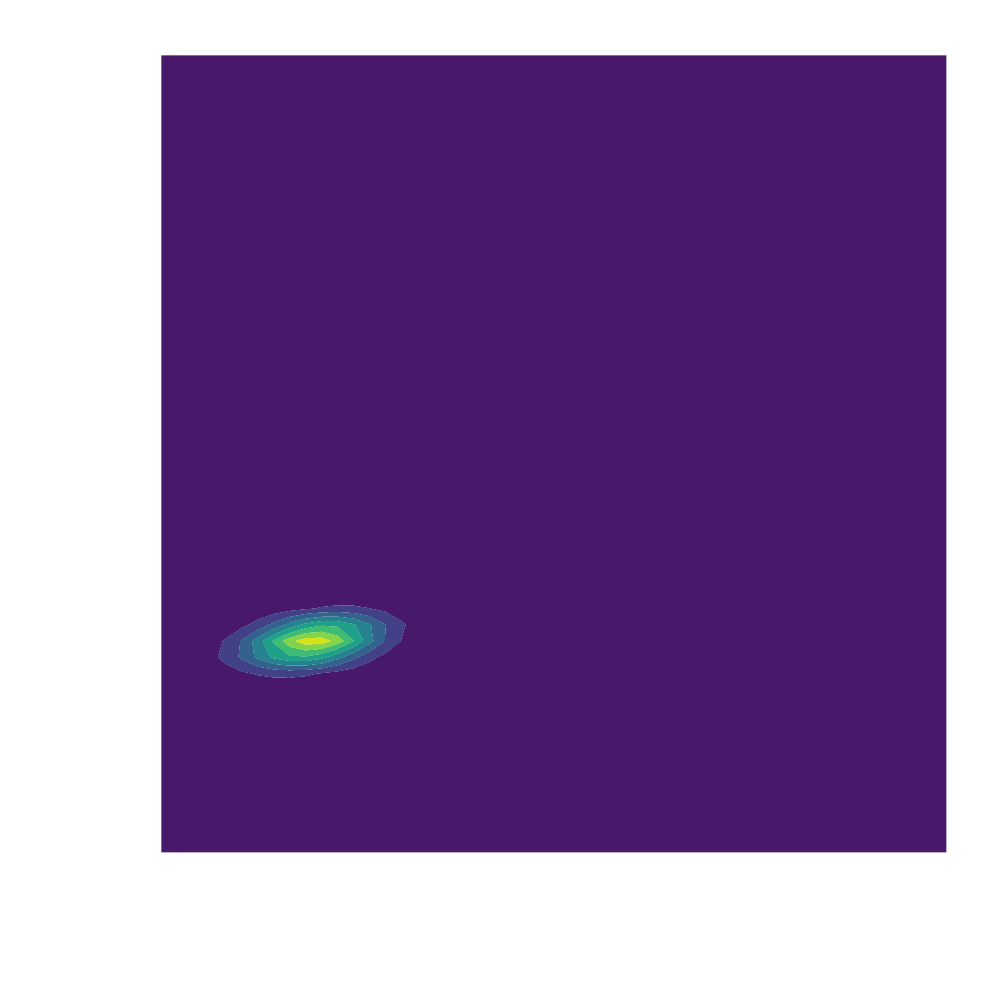}
    \end{subfigure}
    \begin{subfigure}[b]{\postwidth}
    	\centering
        \includegraphics[width=0.97\textwidth]{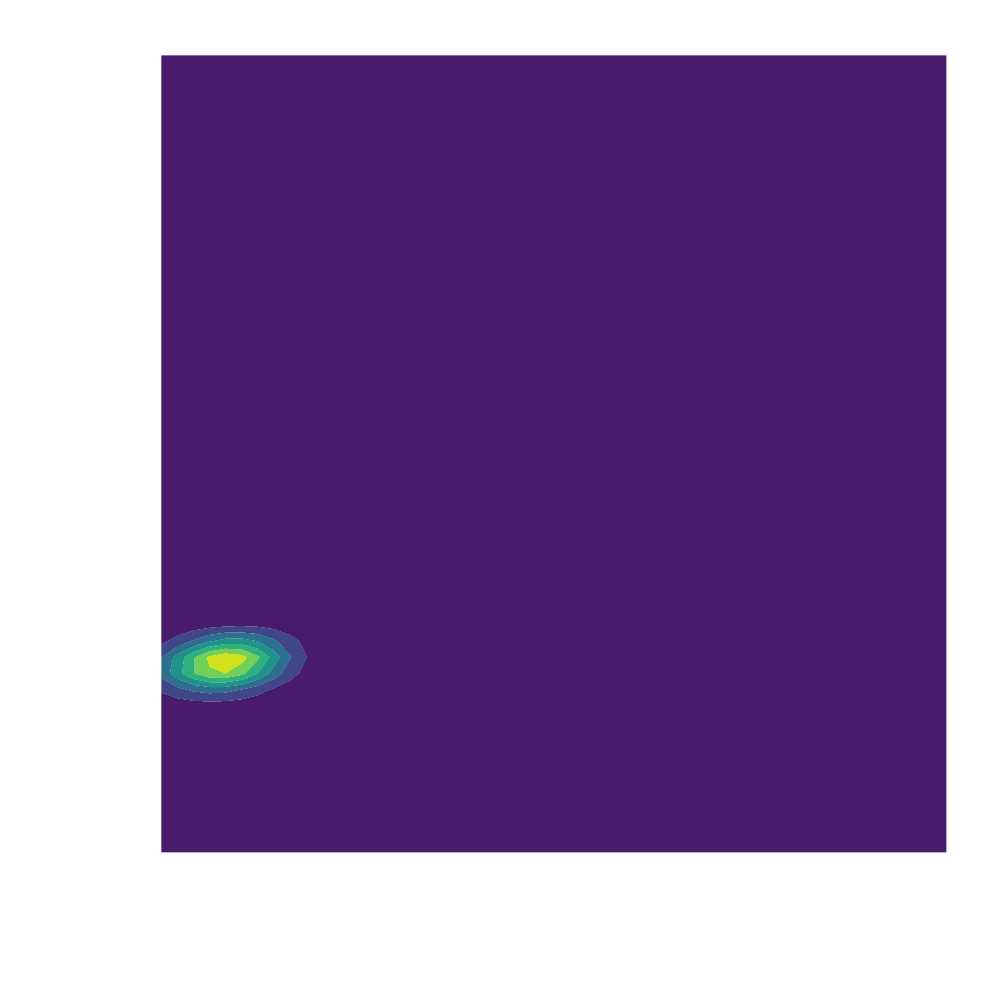}
    \end{subfigure}
    \begin{subfigure}[b]{\postwidth}
    	\centering
        \includegraphics[width=0.97\textwidth]{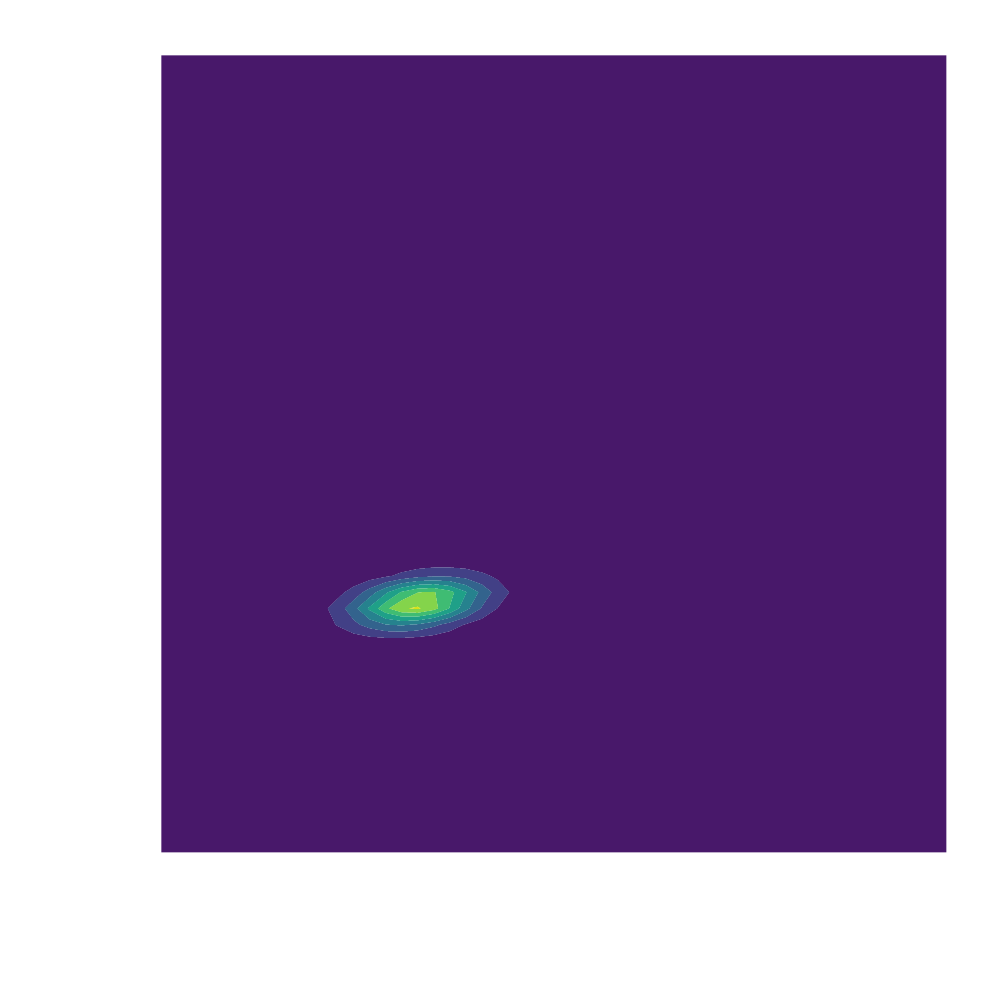}
    \end{subfigure}
    \begin{subfigure}[b]{\postwidth}
    	\centering
        \includegraphics[width=0.97\textwidth]{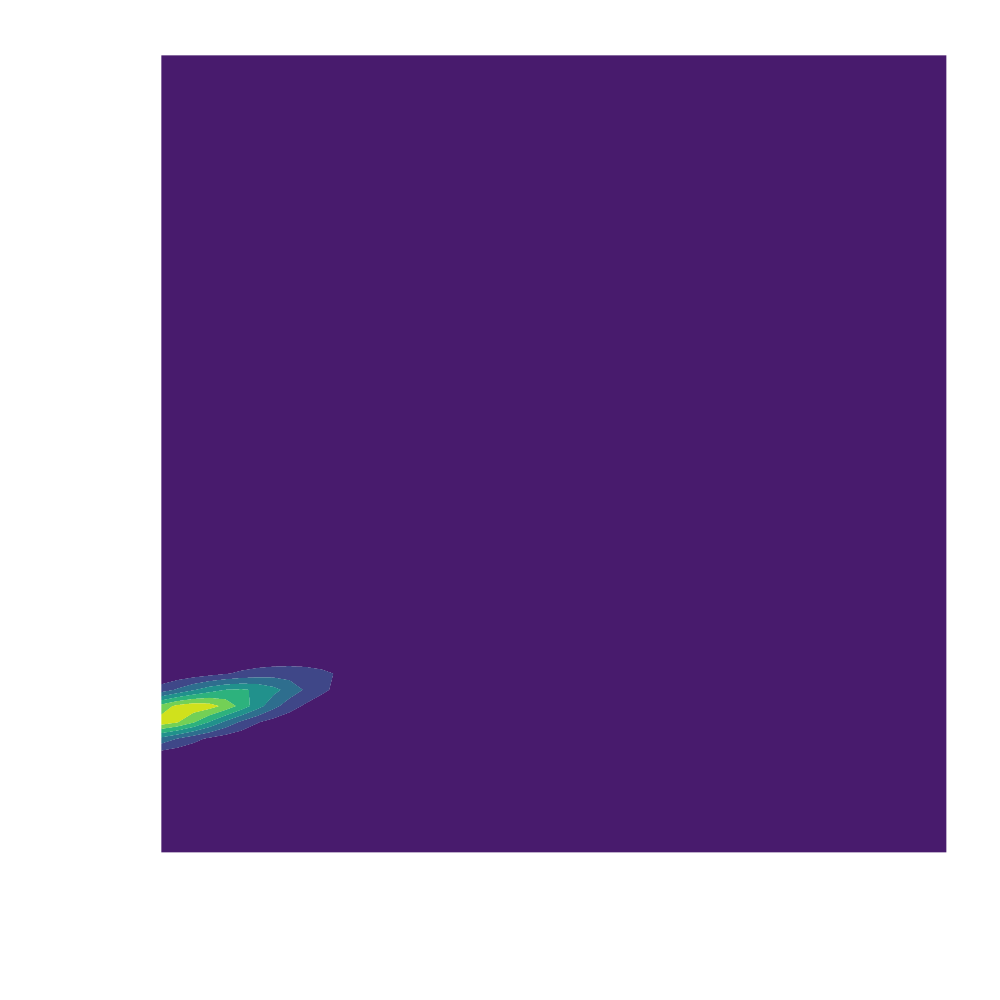}
    \end{subfigure}
    \begin{subfigure}[b]{\postwidth}
    	\centering
        \includegraphics[width=0.97\textwidth]{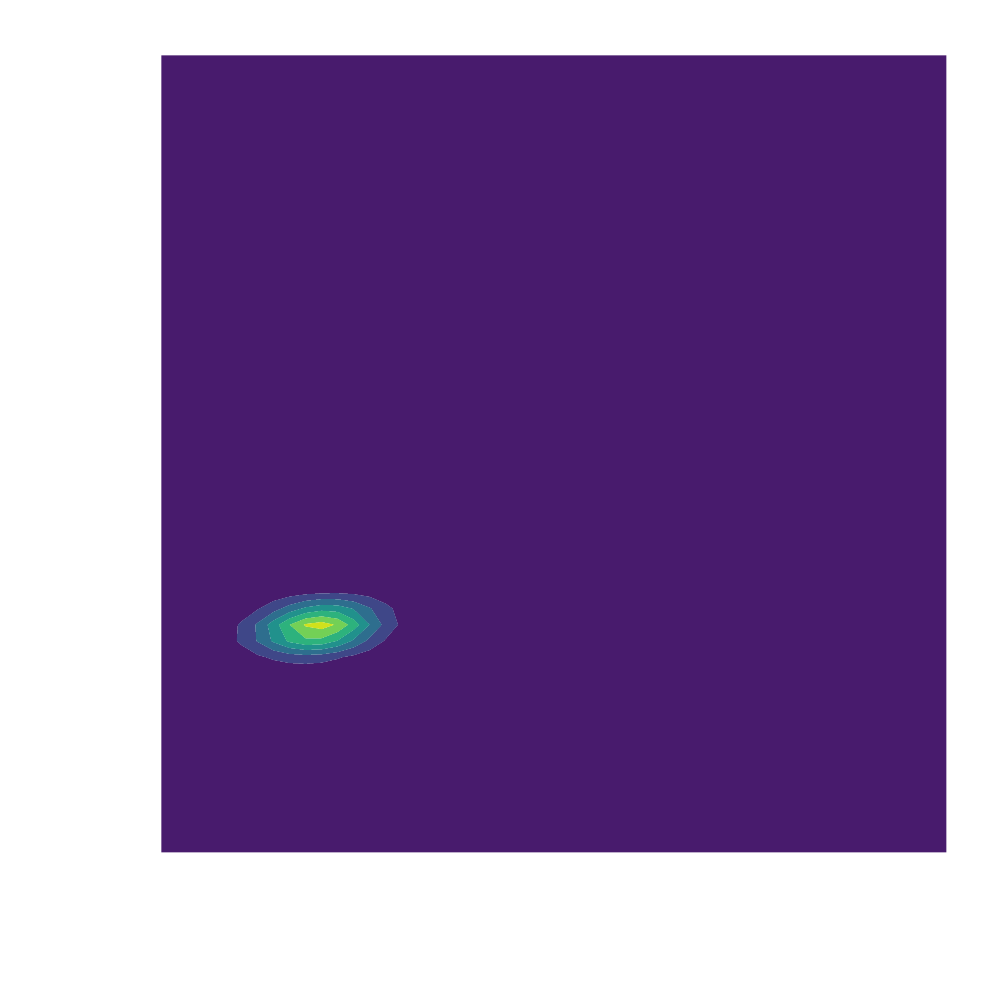}
    \end{subfigure}
    \begin{subfigure}[b]{\postwidth}
    	\centering
        \includegraphics[width=0.97\textwidth]{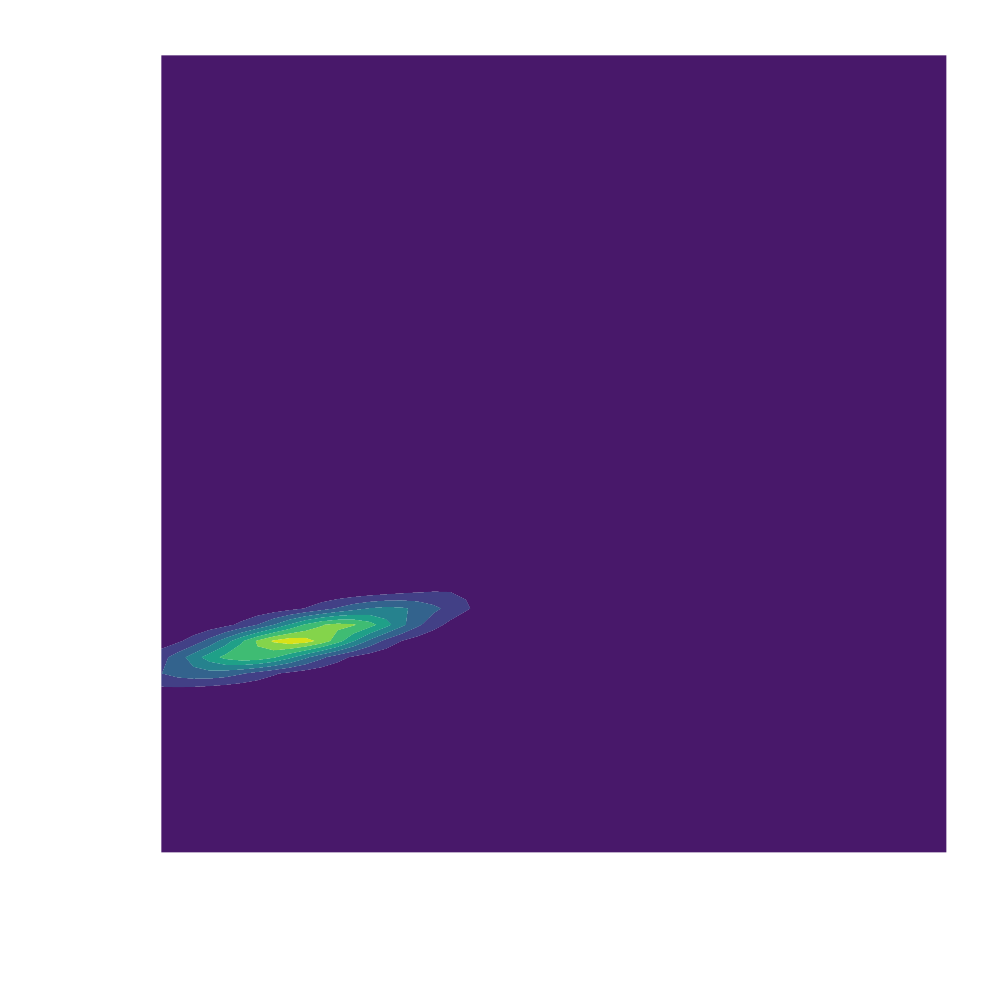}
    \end{subfigure}
    \caption{True posterior $p(z_0 | x_1 .. x_N)$ of Latent ODE trained on Physionet dataset with five features and two-dimensional latent space with and without poisson process likelihood. We train the model with 2-dimensional latent space on a subset of first five attributes. We compute the unnormalized density of the true posterior using the Bayes rule: $p(z_0 | x_1 .. x_N) \propto p(z_0) p(x_1 .. x_N | z_0)$. Similarly, we train the model with poisson process likelihood in the same manner. The posterior distribution is clearly more narrow if trained without poisson}
    \label{fig:true_posterior}
\end{figure}

\newcommand{\nfwidth}{0.15\textwidth}%
\begin{figure*}
	\begin{subfigure}[b]{0.15\linewidth}
		Truth
		\vspace{5mm}
	\end{subfigure}
	\begin{subfigure}[b]{0.8\linewidth}
		\centering
		\includegraphics[width=\nfwidth]{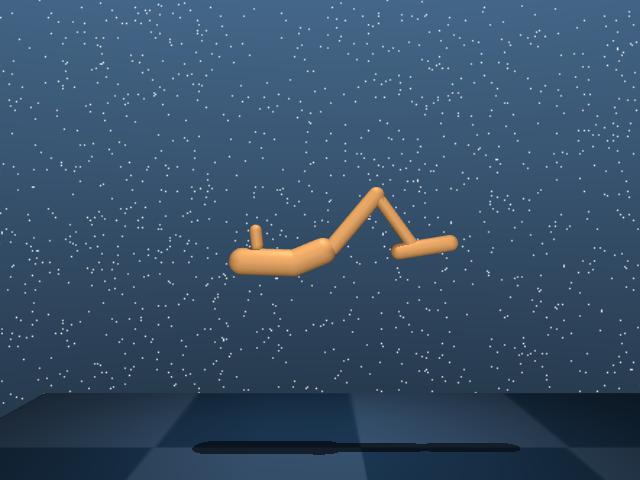}
		\includegraphics[width=\nfwidth]{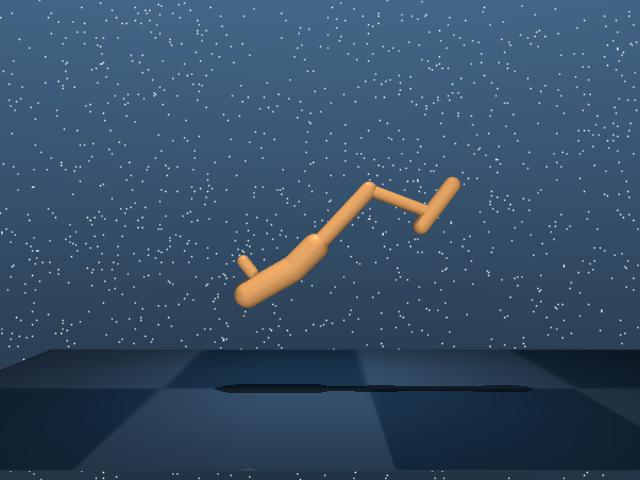}
		\includegraphics[width=\nfwidth]{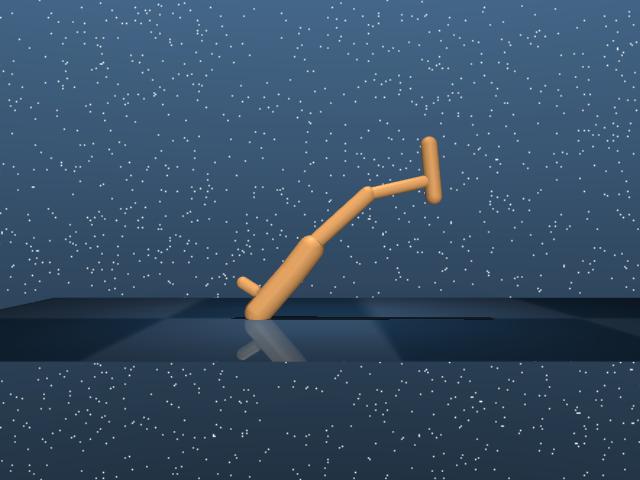}
		\includegraphics[width=\nfwidth]{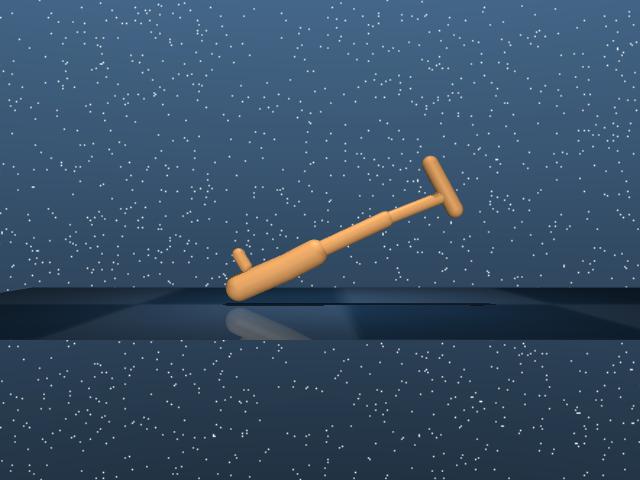}
		\includegraphics[width=\nfwidth]{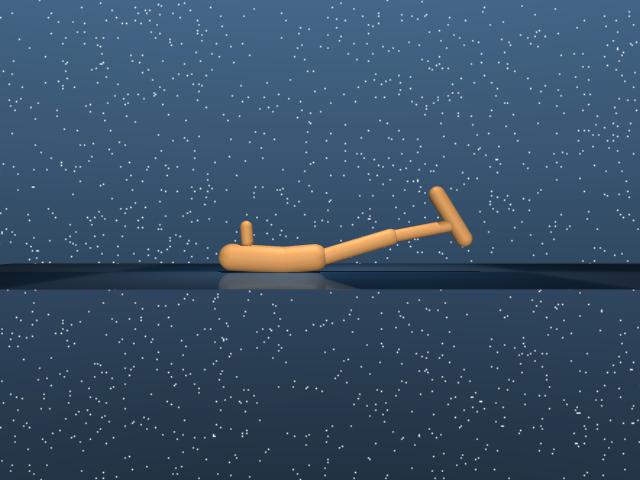}
	\end{subfigure}
	%
	\begin{subfigure}[b]{0.15\linewidth}
		\centering
	   Standard RNN
	    \vspace{5mm}
	\end{subfigure}
	\begin{subfigure}[b]{0.8\linewidth}
		\centering
		\includegraphics[width=\nfwidth]{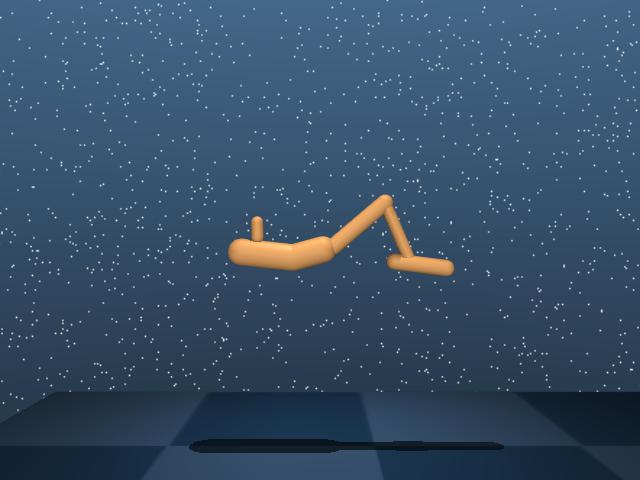}
		\includegraphics[width=\nfwidth]{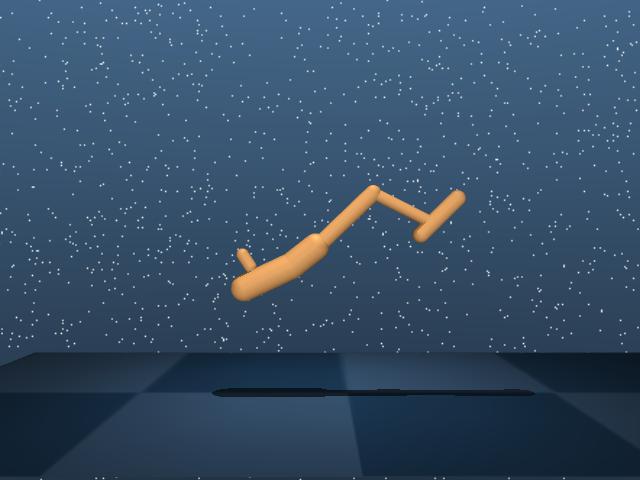}
		\includegraphics[width=\nfwidth]{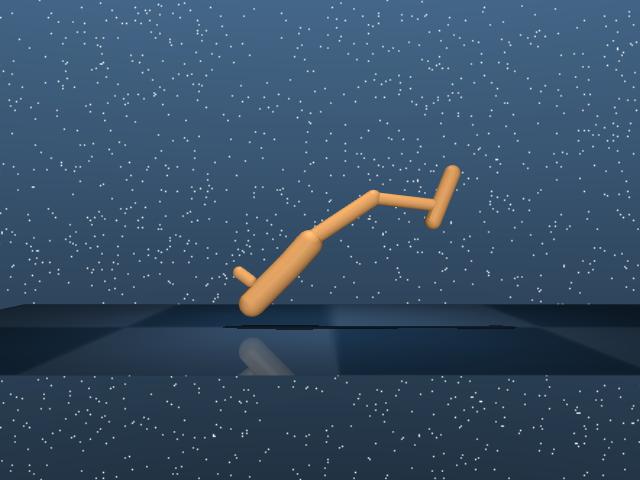}
		\includegraphics[width=\nfwidth]{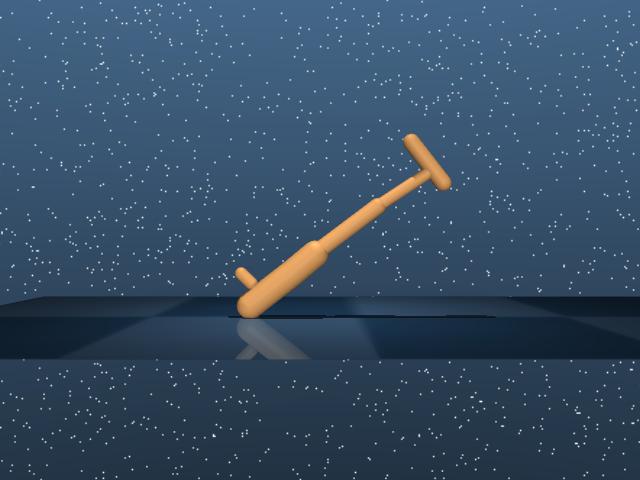}
		\includegraphics[width=\nfwidth]{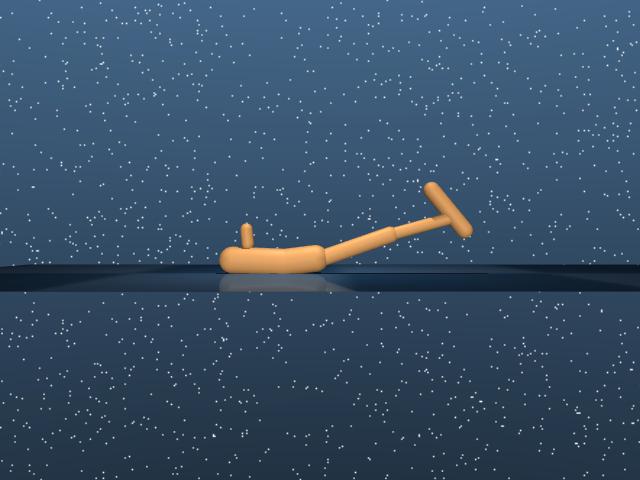}
	\end{subfigure}
	%
	\begin{subfigure}[b]{0.15\linewidth}
		\centering
		ODE-GRU
		\vspace{7mm}
	\end{subfigure}
	\begin{subfigure}[b]{0.8\linewidth}
		\centering
		\includegraphics[width=\nfwidth]{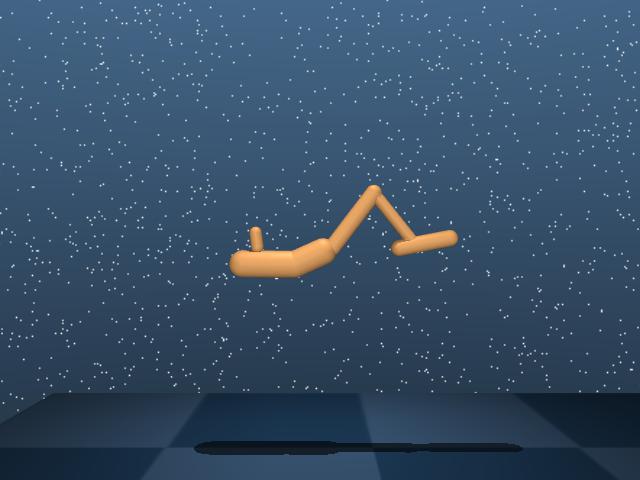}
		\includegraphics[width=\nfwidth]{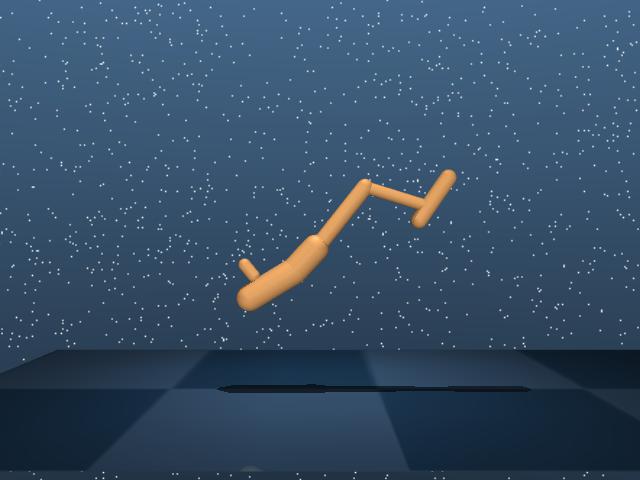}
		\includegraphics[width=\nfwidth]{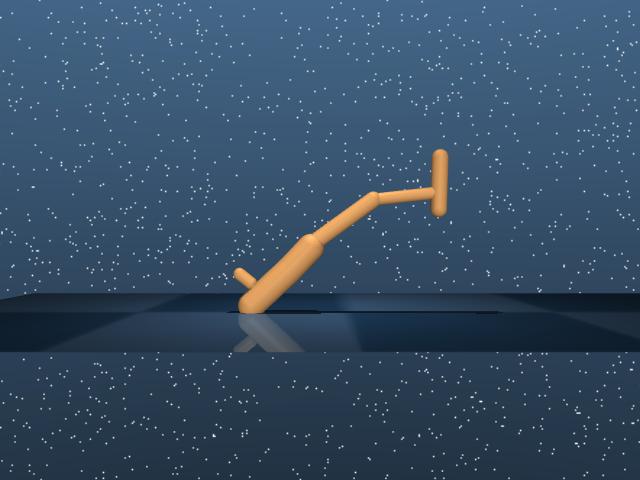}
		\includegraphics[width=\nfwidth]{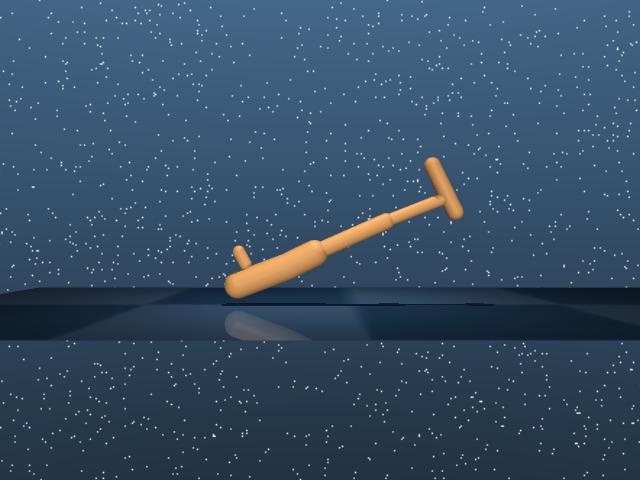}
		\includegraphics[width=\nfwidth]{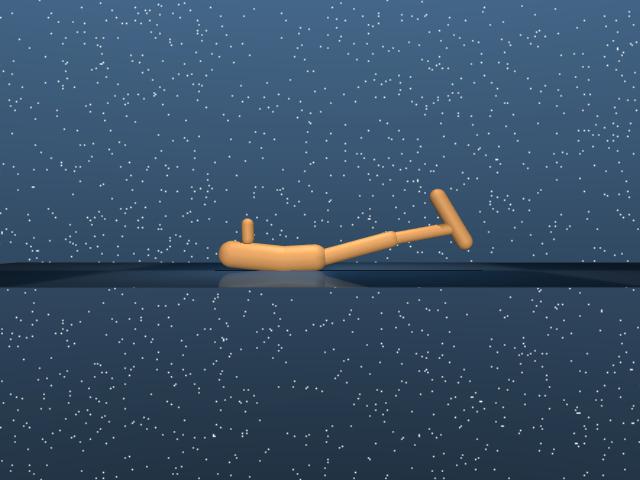}
	\end{subfigure}
	%
	\begin{subfigure}[b]{0.15\linewidth}
		\centering
		RNN-VAE
		\vspace{7mm}
	\end{subfigure}
	\begin{subfigure}[b]{0.8\linewidth}
		\centering
		\includegraphics[width=\nfwidth]{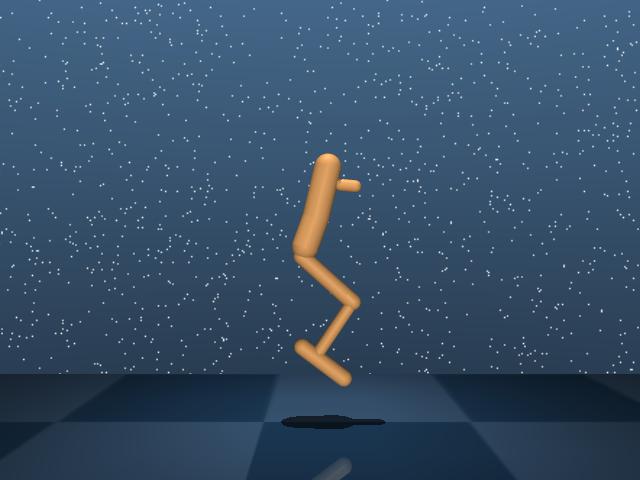}
		\includegraphics[width=\nfwidth]{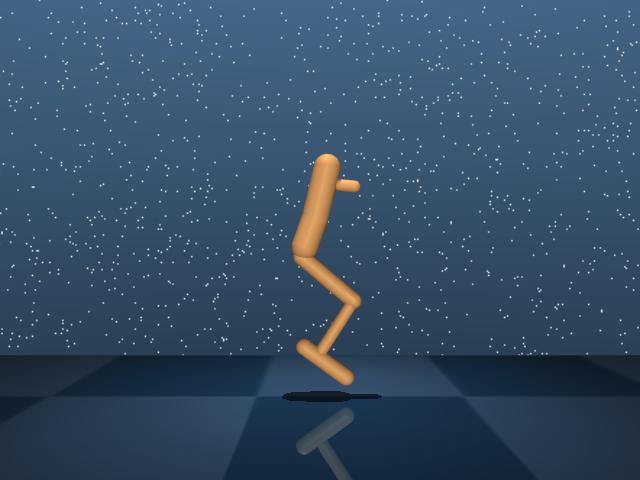}
		\includegraphics[width=\nfwidth]{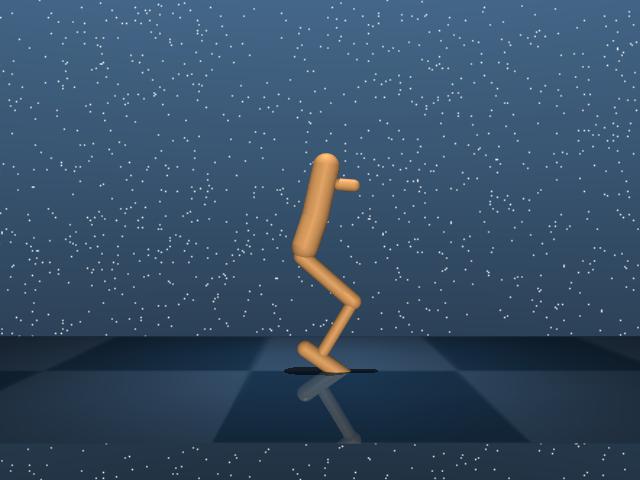}
		\includegraphics[width=\nfwidth]{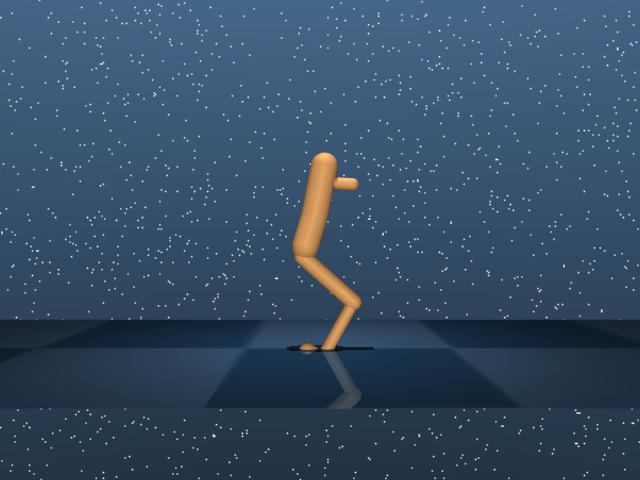}
		\includegraphics[width=\nfwidth]{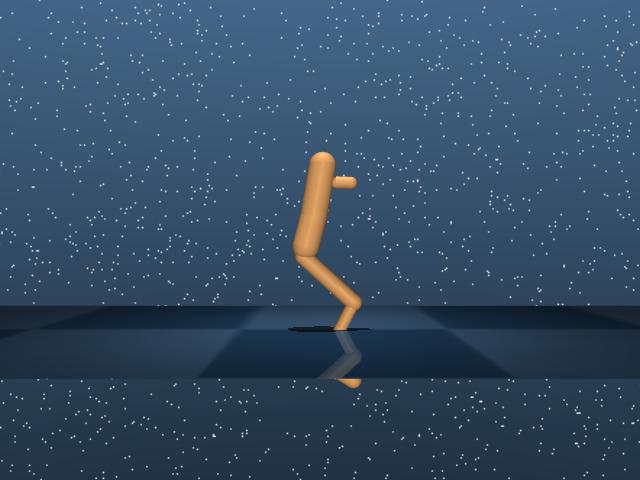}
	\end{subfigure}
	%
	\begin{subfigure}[b]{0.15\linewidth}
		\centering
		Latent ODE (RNN enc)
		\vspace{7mm}
	\end{subfigure}
	\begin{subfigure}[b]{0.8\linewidth}
		\centering
		\includegraphics[width=\nfwidth]{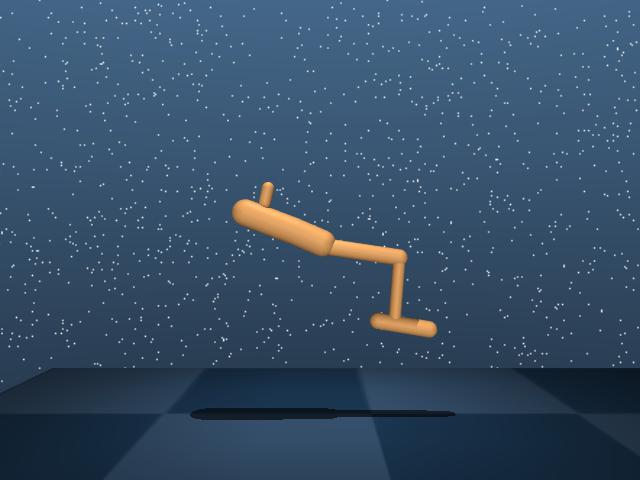}
		\includegraphics[width=\nfwidth]{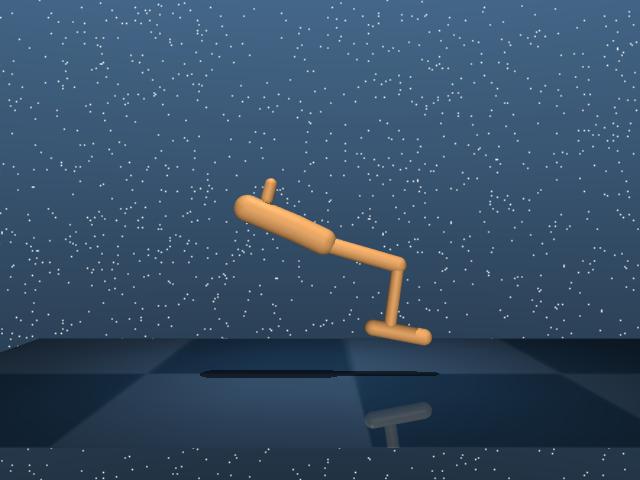}
		\includegraphics[width=\nfwidth]{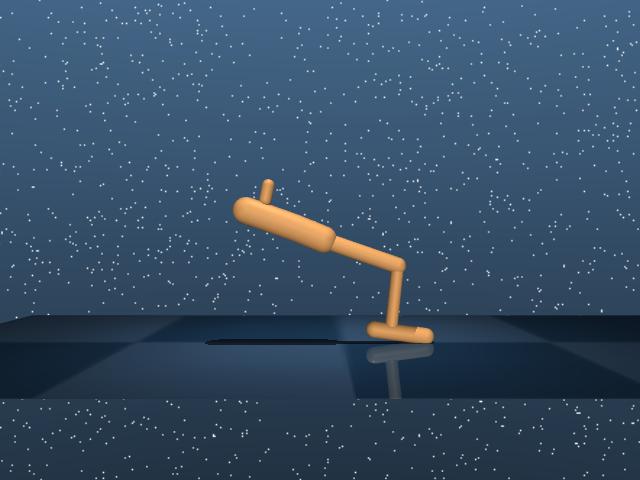}
		\includegraphics[width=\nfwidth]{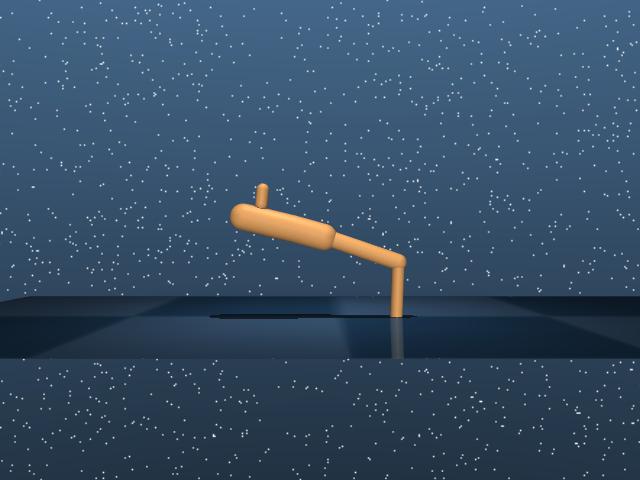}
		\includegraphics[width=\nfwidth]{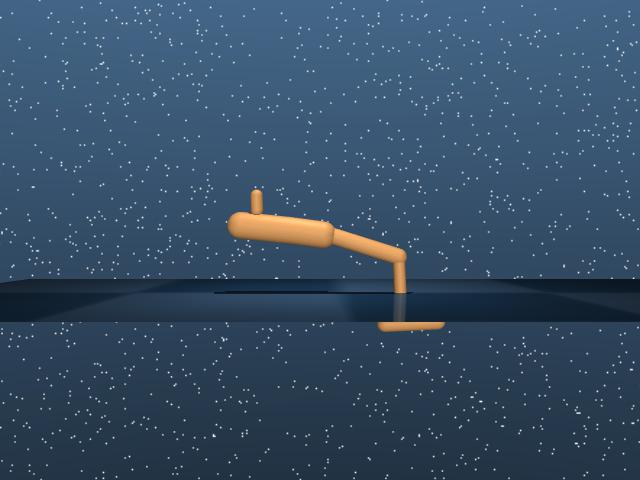}
	\end{subfigure}
	%
	\begin{subfigure}[b]{0.15\linewidth}
		\centering
		Latent ODE (ODE enc)
		\vspace{7mm}
	\end{subfigure}
	\hfill
	\begin{subfigure}[b]{0.8\linewidth}
		\centering
		\includegraphics[width=\nfwidth]{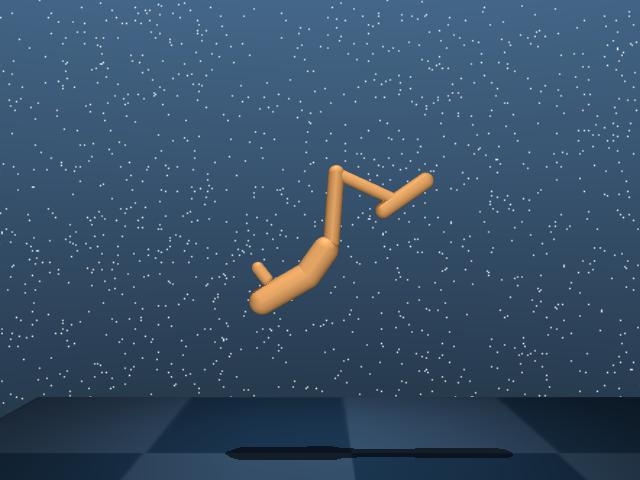}
		\includegraphics[width=\nfwidth]{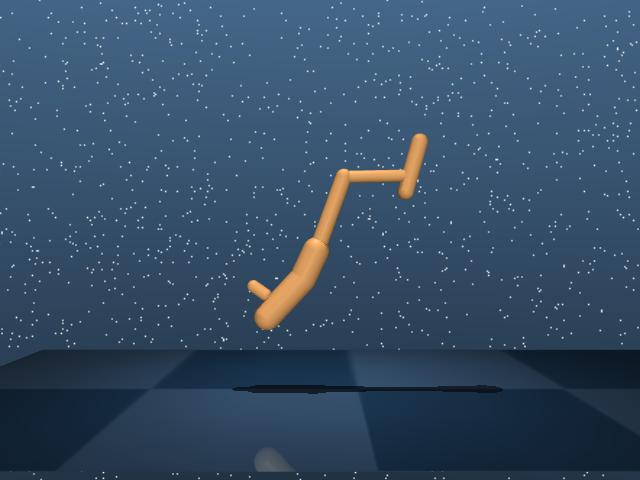}
		\includegraphics[width=\nfwidth]{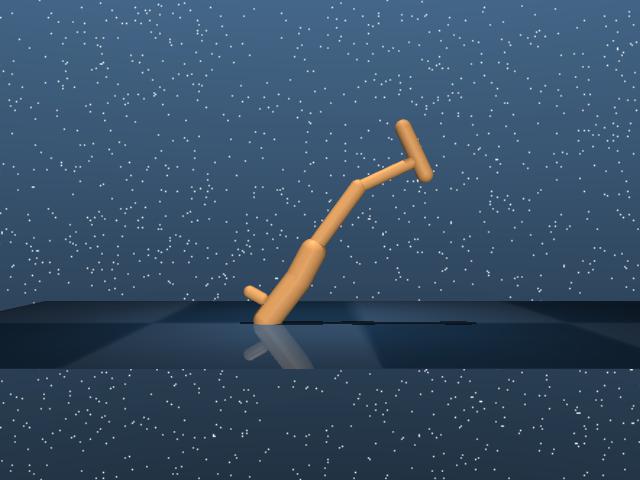}
		\includegraphics[width=\nfwidth]{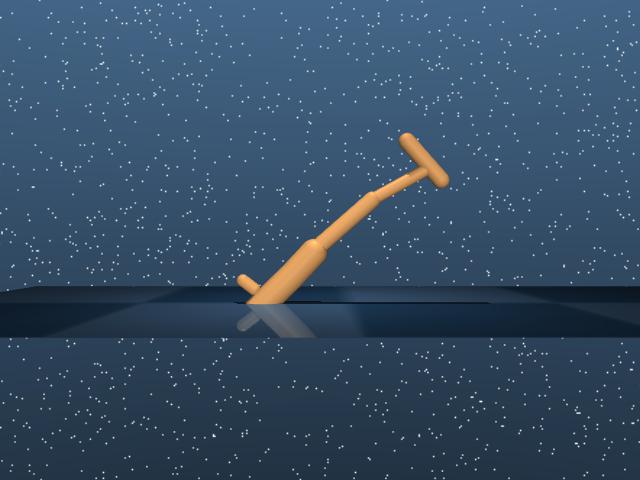}
		\includegraphics[width=\nfwidth]{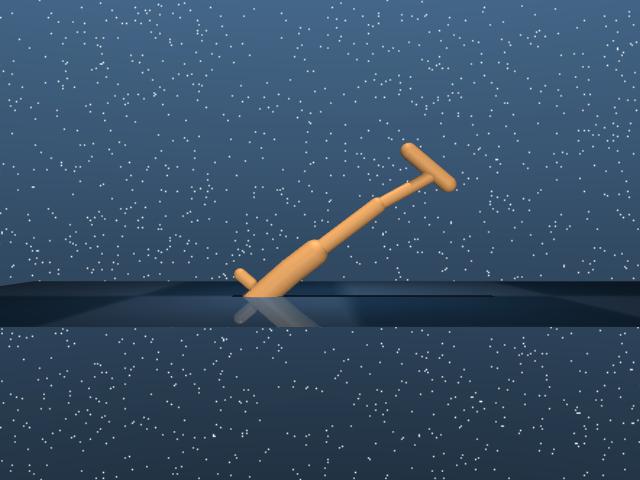}
	\end{subfigure}
	\caption{Reconstructed trajectories on Mujoco dataset}
\label{fig:mujoco}
\end{figure*}

\newcommand{\fgwidth}{0.235\columnwidth}%
\begin{figure*}
	\centering
    \begin{subfigure}[b]{0.02\columnwidth}
	\centering
    \hfill
    \end{subfigure}
    \begin{subfigure}[b]{\fgwidth}
    	\centering
        Patient 1
    \end{subfigure}
    \begin{subfigure}[b]{\fgwidth}
    	\centering
        Patient 2
    \end{subfigure}
    \begin{subfigure}[b]{\fgwidth}
    	\centering
        Patient 3
    \end{subfigure}
    \begin{subfigure}[b]{\fgwidth}
    	\centering
        Patient 4
    \end{subfigure}
    \begin{subfigure}[b]{0.02\columnwidth}
	\centering
	\small 
	\rotatebox{90}{MAP}
	\vspace{8mm}
    \end{subfigure}
    \begin{subfigure}[b]{\fgwidth}
    	\centering
        \includegraphics[width=\textwidth]{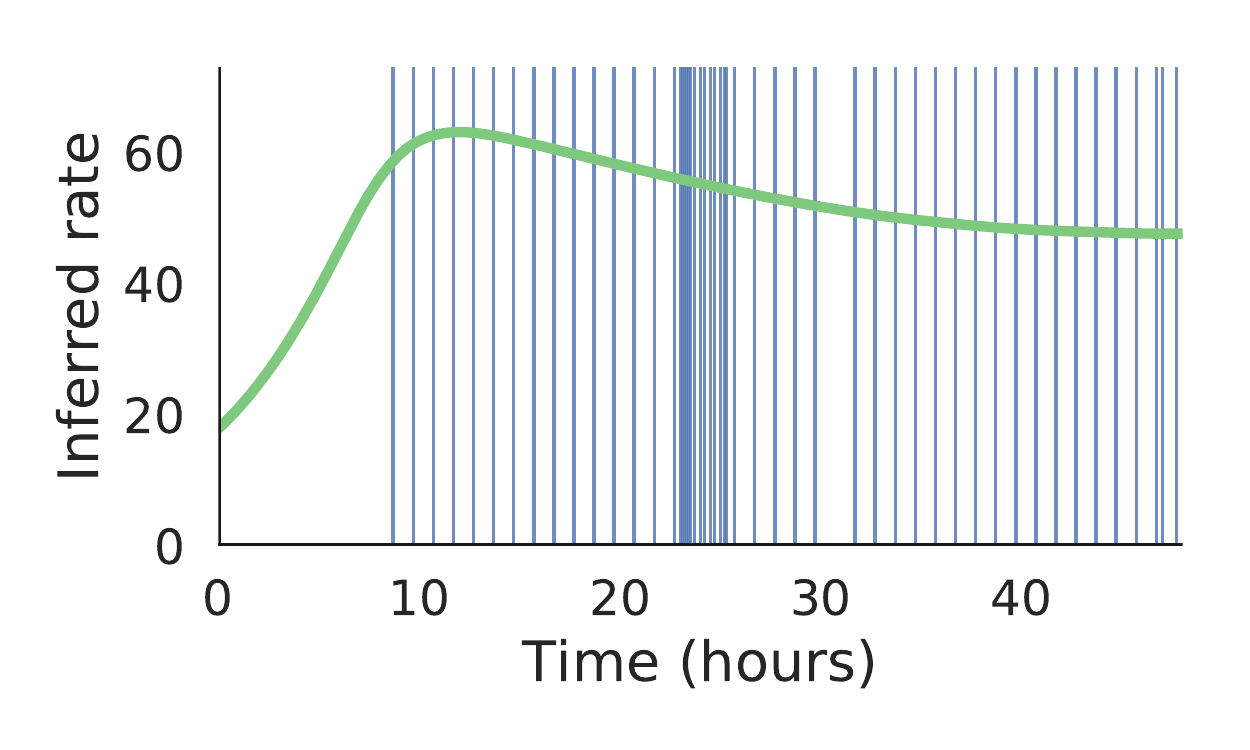}
    \end{subfigure}
    \begin{subfigure}[b]{\fgwidth}
    	\centering
        \includegraphics[width=\textwidth]{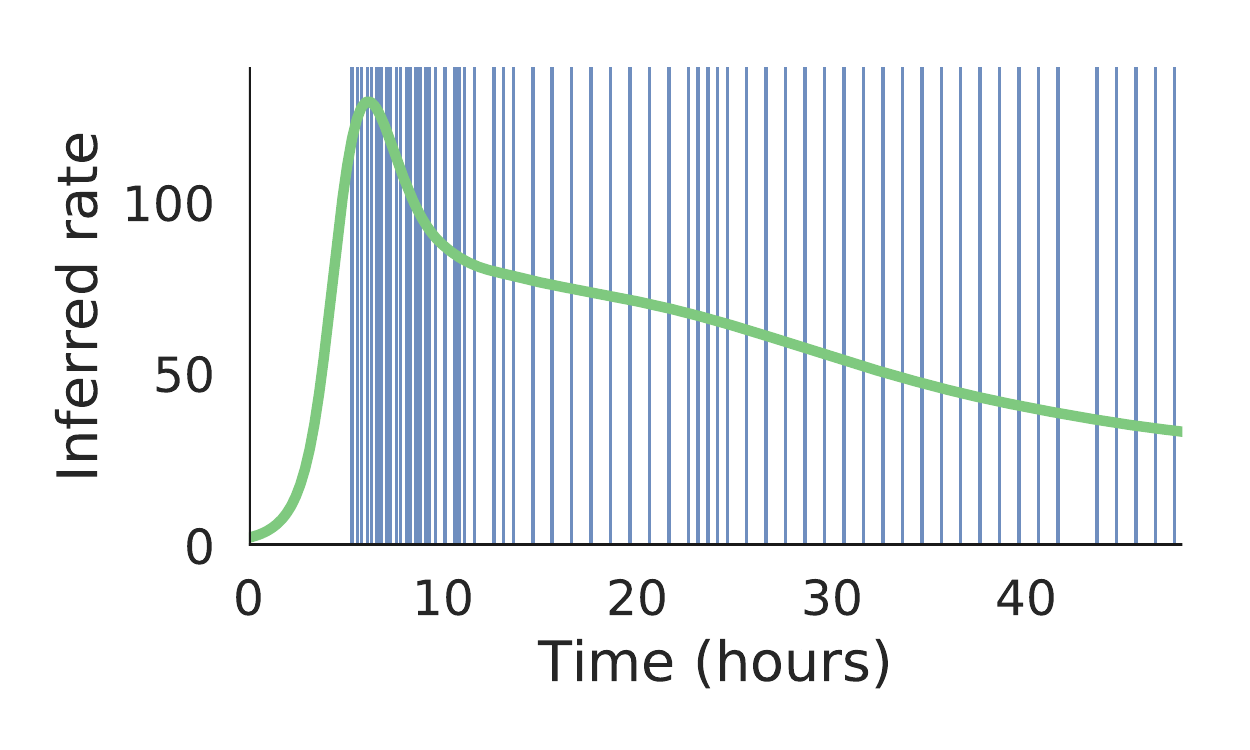}
    \end{subfigure}
    \begin{subfigure}[b]{\fgwidth}
    	\centering
        \includegraphics[width=\textwidth]{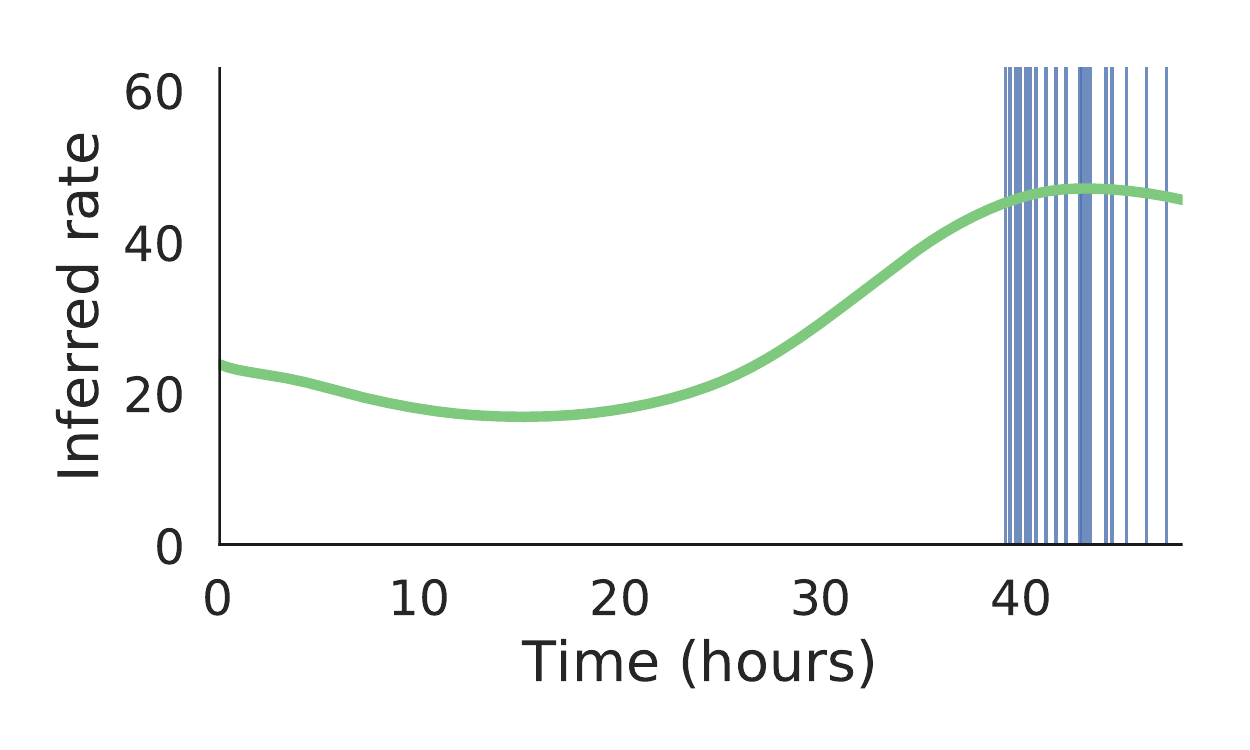}
    \end{subfigure}
    \begin{subfigure}[b]{\fgwidth}
    	\centering
        \includegraphics[width=\textwidth]{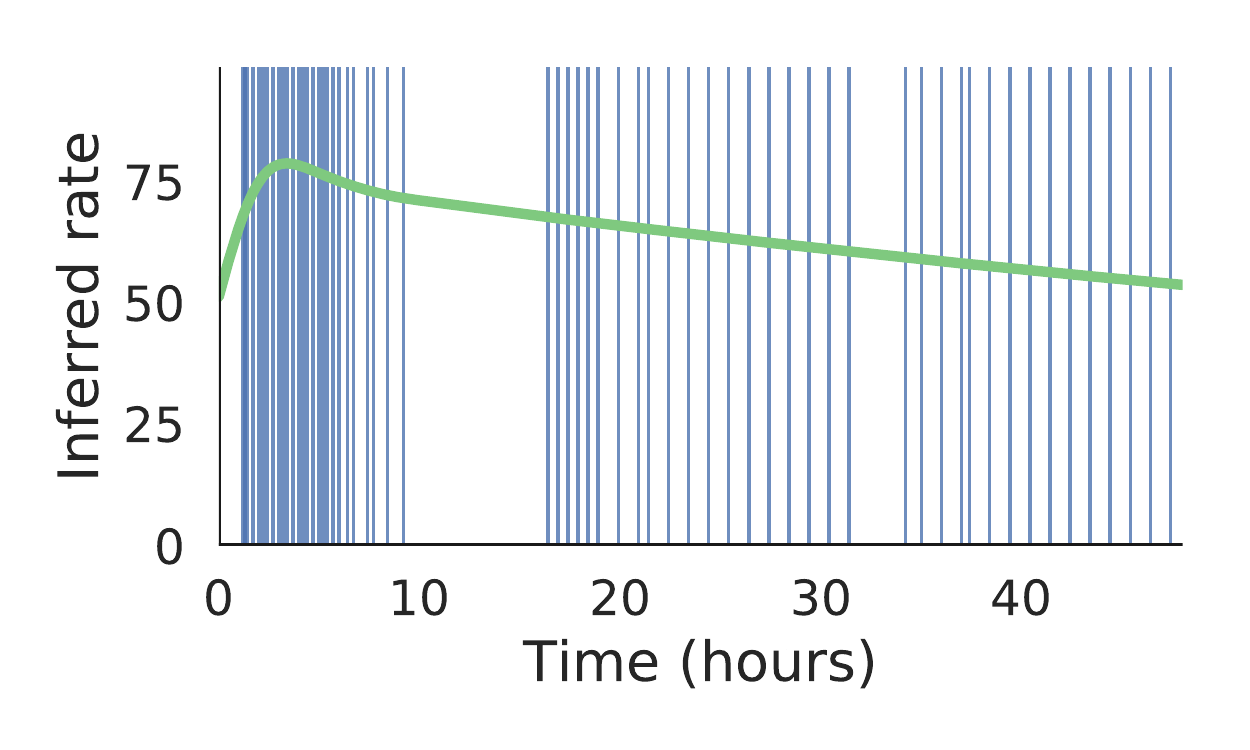}
    \end{subfigure}
    \begin{subfigure}[b]{0.02\columnwidth}
	\centering
	\small
	\rotatebox{90}{NIDiasABP}
	\vspace{3mm}
    \end{subfigure}
    \begin{subfigure}[b]{\fgwidth}
    	\centering
        \includegraphics[width=\textwidth]{plots/poisson/poisson_traj_0_NIDiasABP_24622_test}
    \end{subfigure}
    \begin{subfigure}[b]{\fgwidth}
    	\centering
        No data
        \vspace{1cm}
    \end{subfigure}
    \begin{subfigure}[b]{\fgwidth}
    	\centering
        \includegraphics[width=\textwidth]{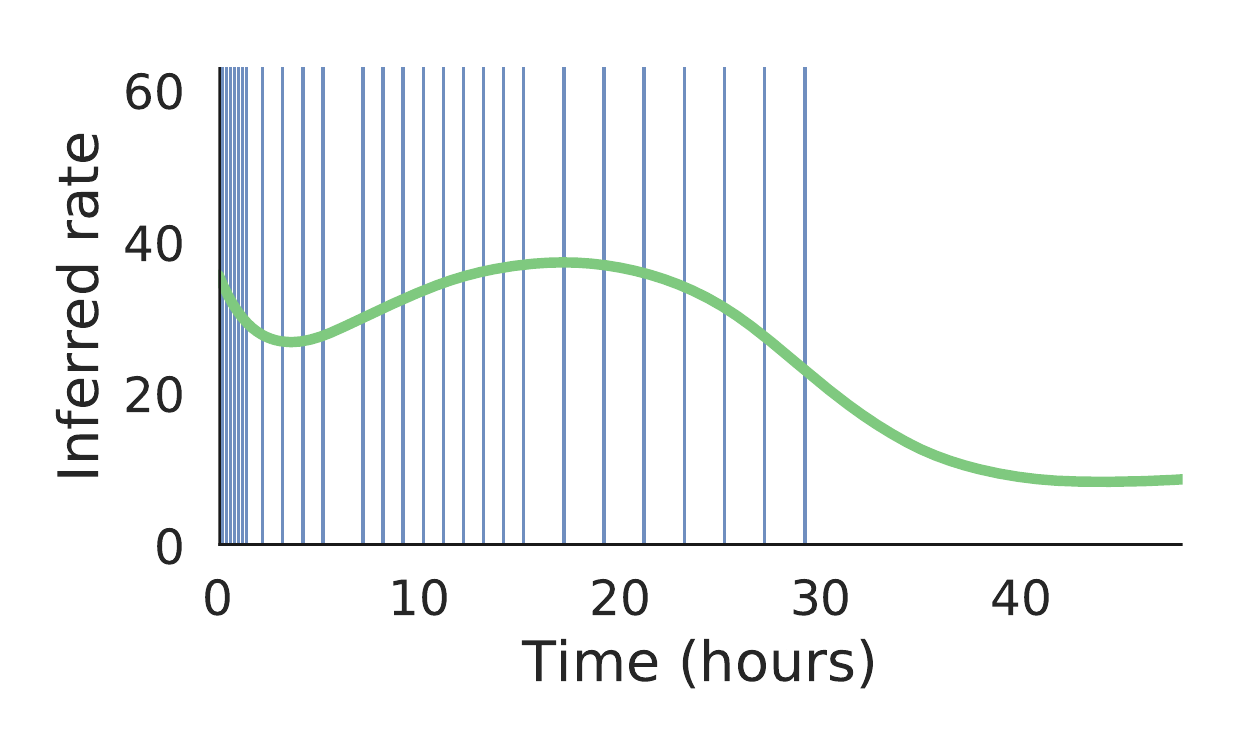}
    \end{subfigure}
    \begin{subfigure}[b]{\fgwidth}
    	\centering
        \includegraphics[width=\textwidth]{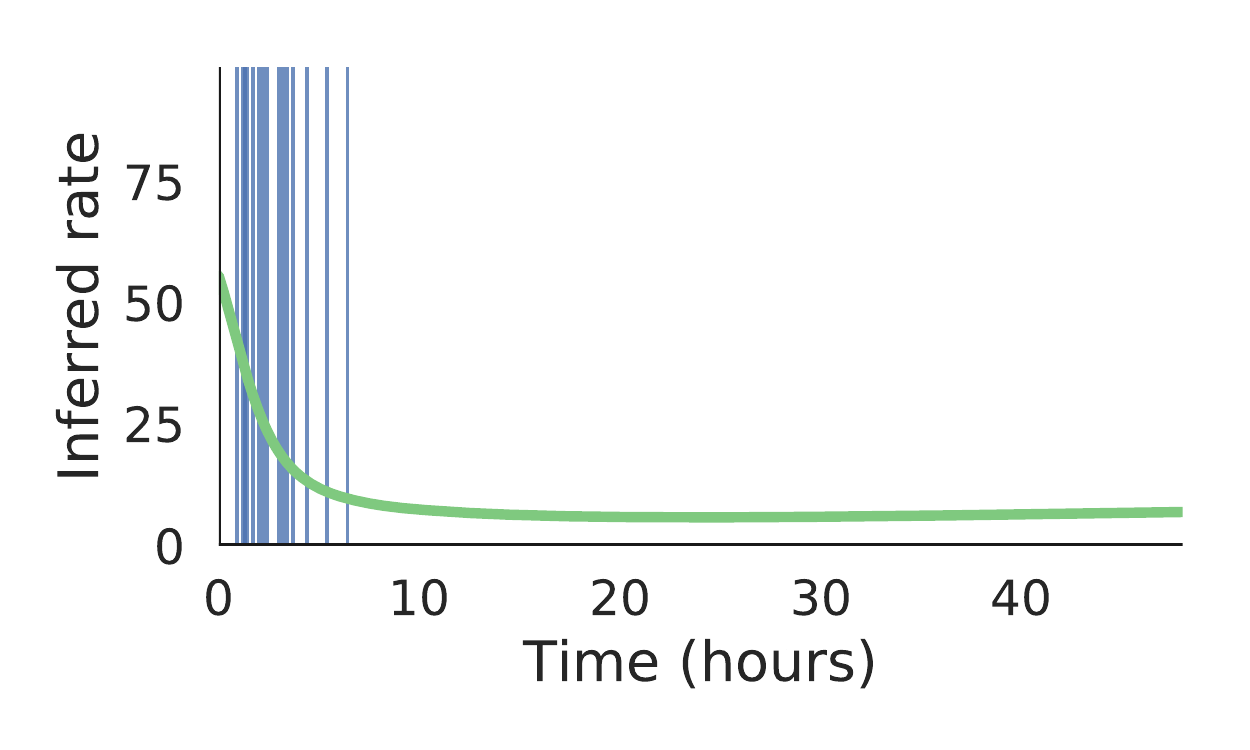}
    \end{subfigure}
    \begin{subfigure}[b]{0.02\columnwidth}
	\centering
	\small 
	\rotatebox{90}{PaO2}
	\vspace{8mm}
    \end{subfigure}
    \begin{subfigure}[b]{\fgwidth}
    	\centering
        \includegraphics[width=\textwidth]{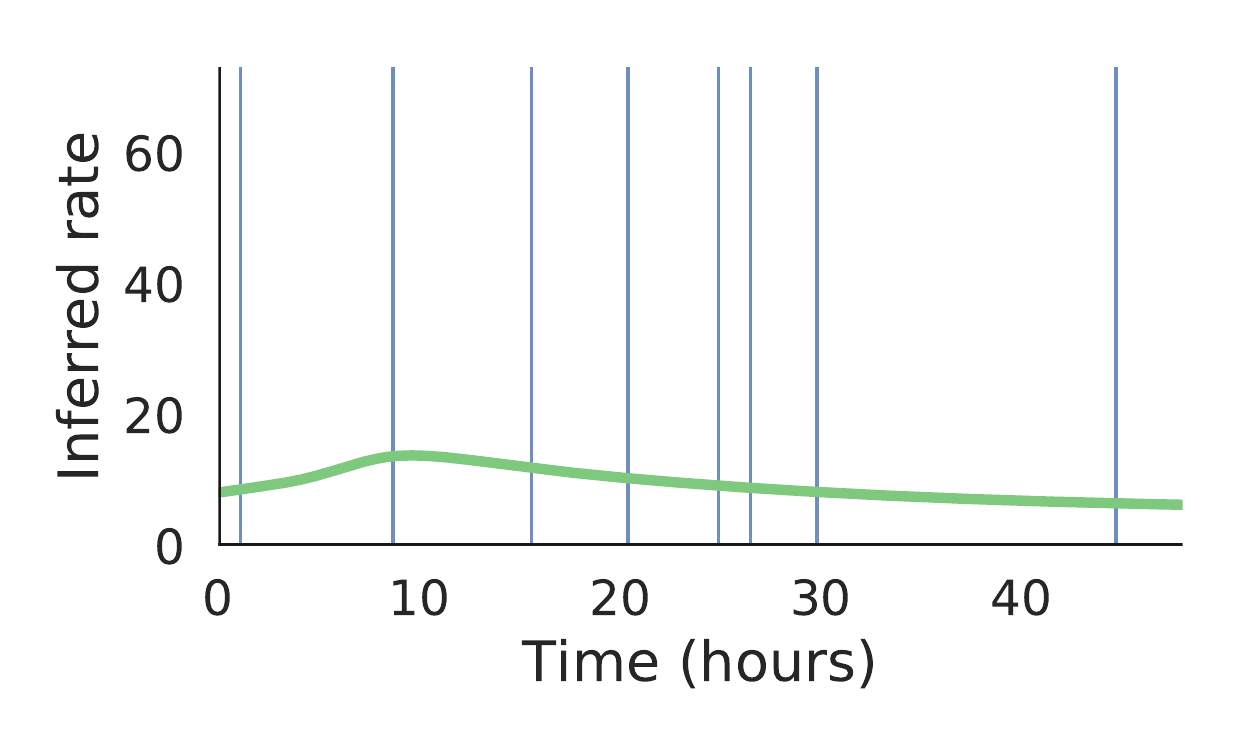}
    \end{subfigure}
    \begin{subfigure}[b]{\fgwidth}
    	\centering
        \includegraphics[width=\textwidth]{plots/poisson/poisson_traj_16_PaO2_24622_test}
    \end{subfigure}
    \begin{subfigure}[b]{\fgwidth}
    	\centering
        \includegraphics[width=\textwidth]{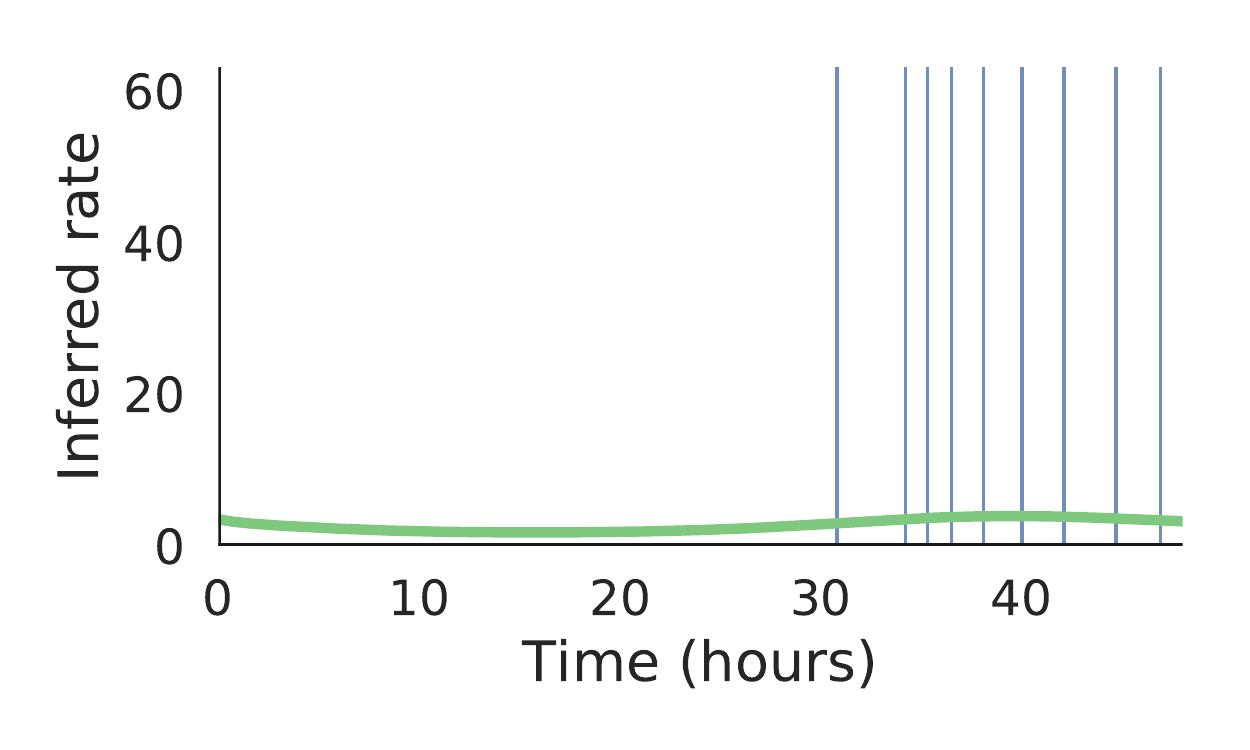}
    \end{subfigure}
    \begin{subfigure}[b]{\fgwidth}
    	\centering
        \includegraphics[width=\textwidth]{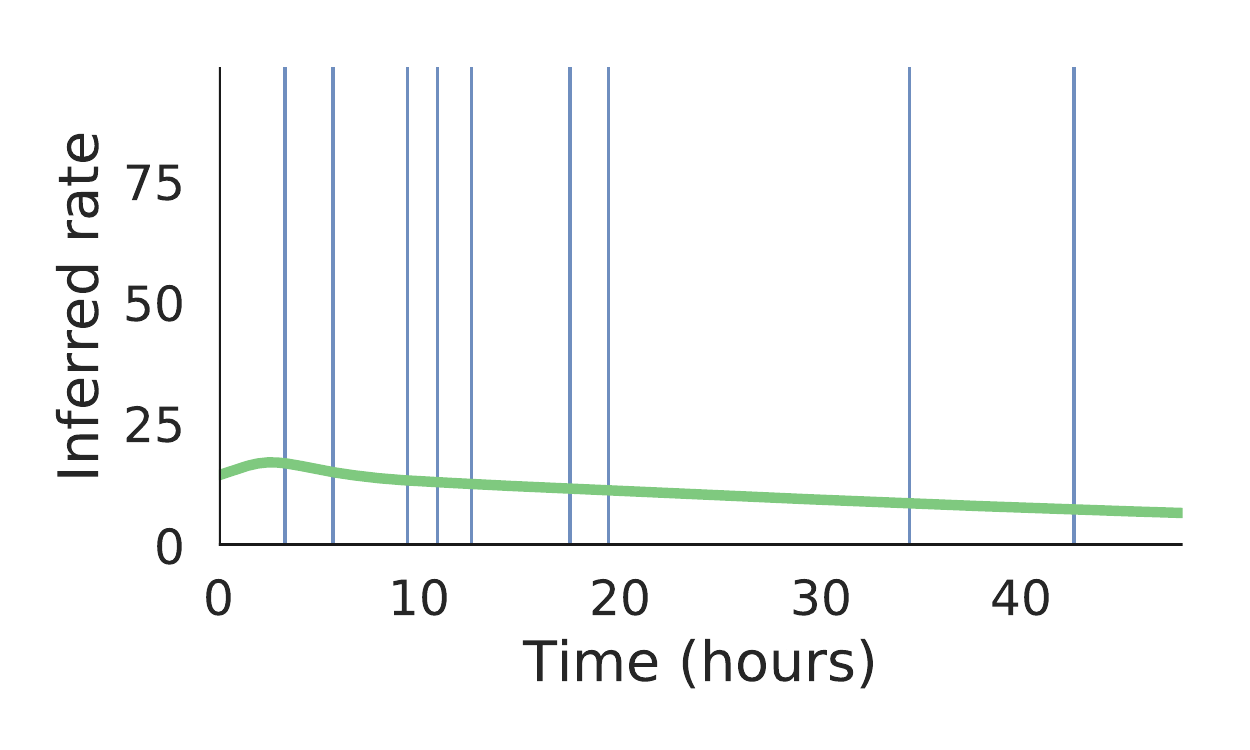}
    \end{subfigure}
    \begin{subfigure}[b]{0.02\columnwidth}
	\centering
	\small
	\rotatebox{90}{HR}
	\vspace{11mm}
    \end{subfigure}
    \begin{subfigure}[b]{\fgwidth}
    	\centering
        \includegraphics[width=\textwidth]{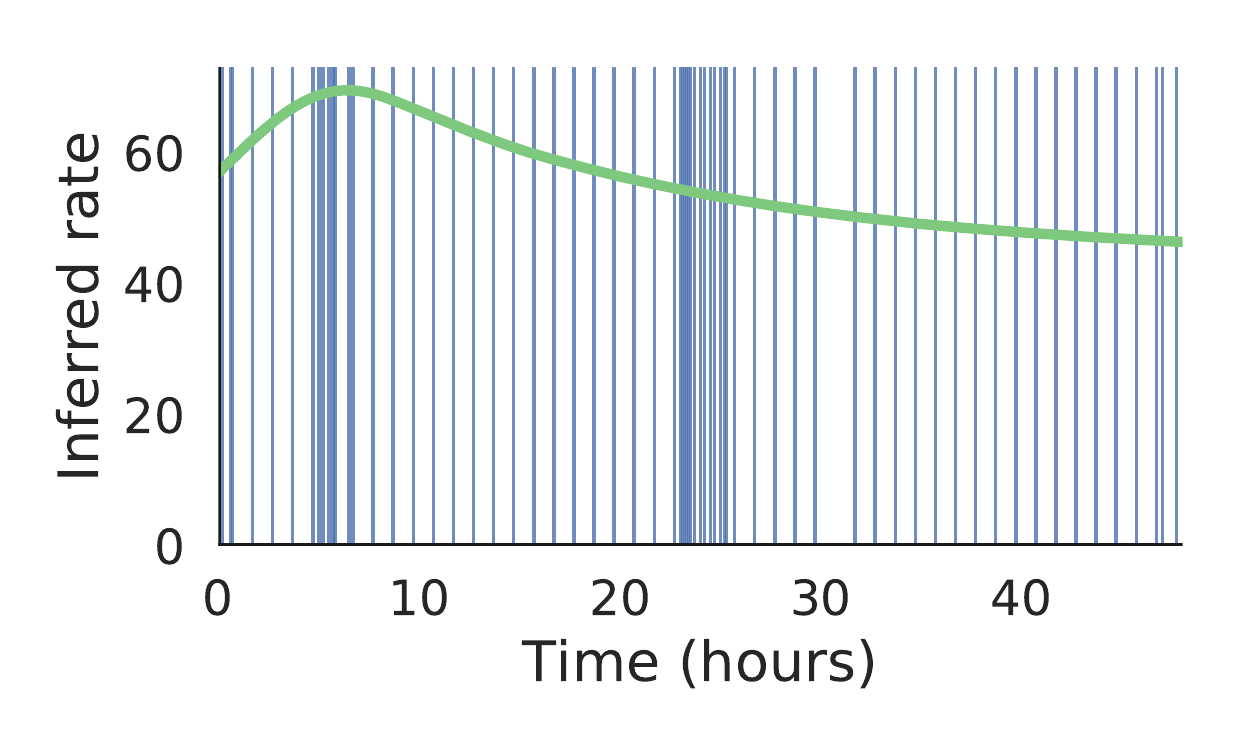}
    \end{subfigure}
    \begin{subfigure}[b]{\fgwidth}
    	\centering
        \includegraphics[width=\textwidth]{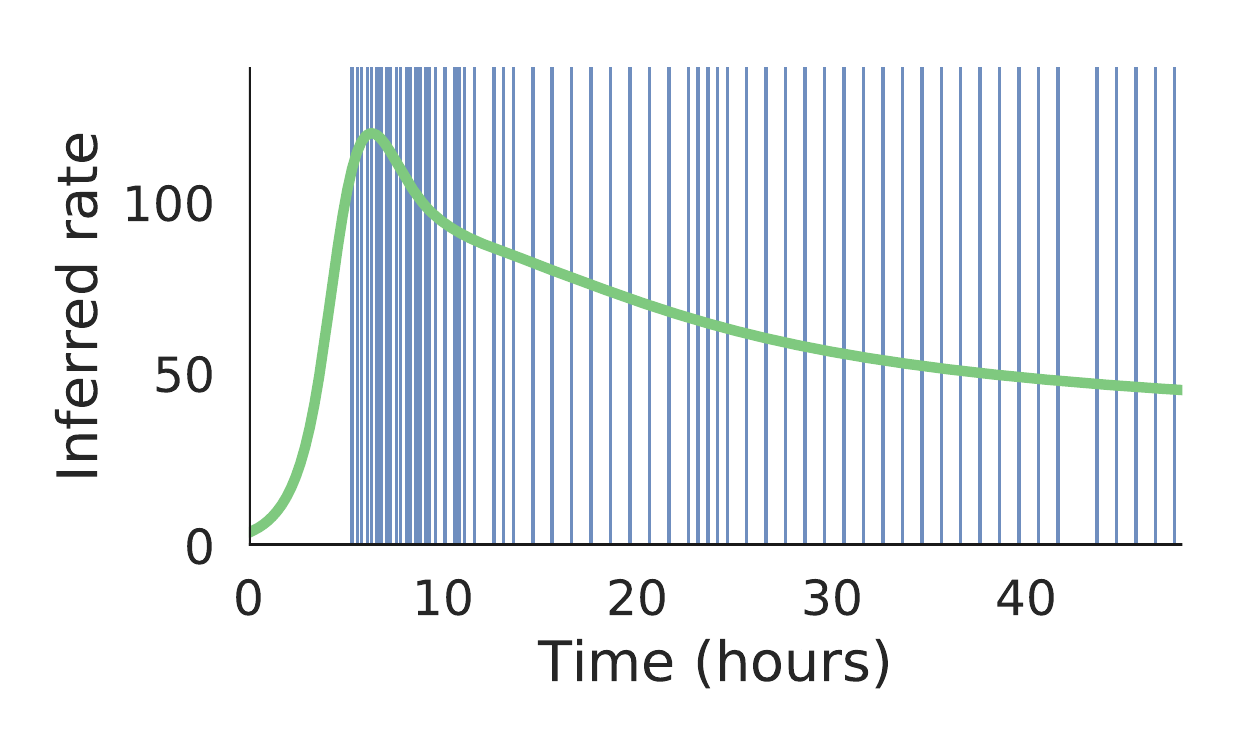}
    \end{subfigure}
    \begin{subfigure}[b]{\fgwidth}
    	\centering
        \includegraphics[width=\textwidth]{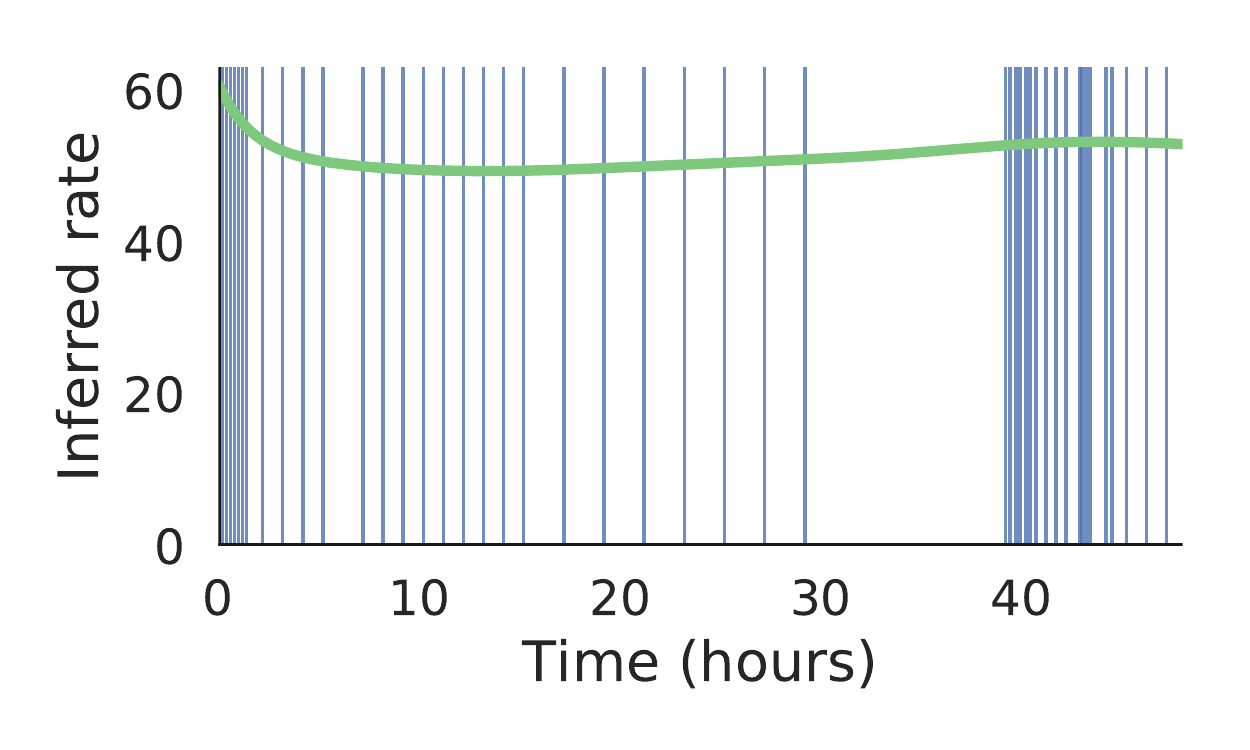}
    \end{subfigure}
    \begin{subfigure}[b]{\fgwidth}
    	\centering
        \includegraphics[width=\textwidth]{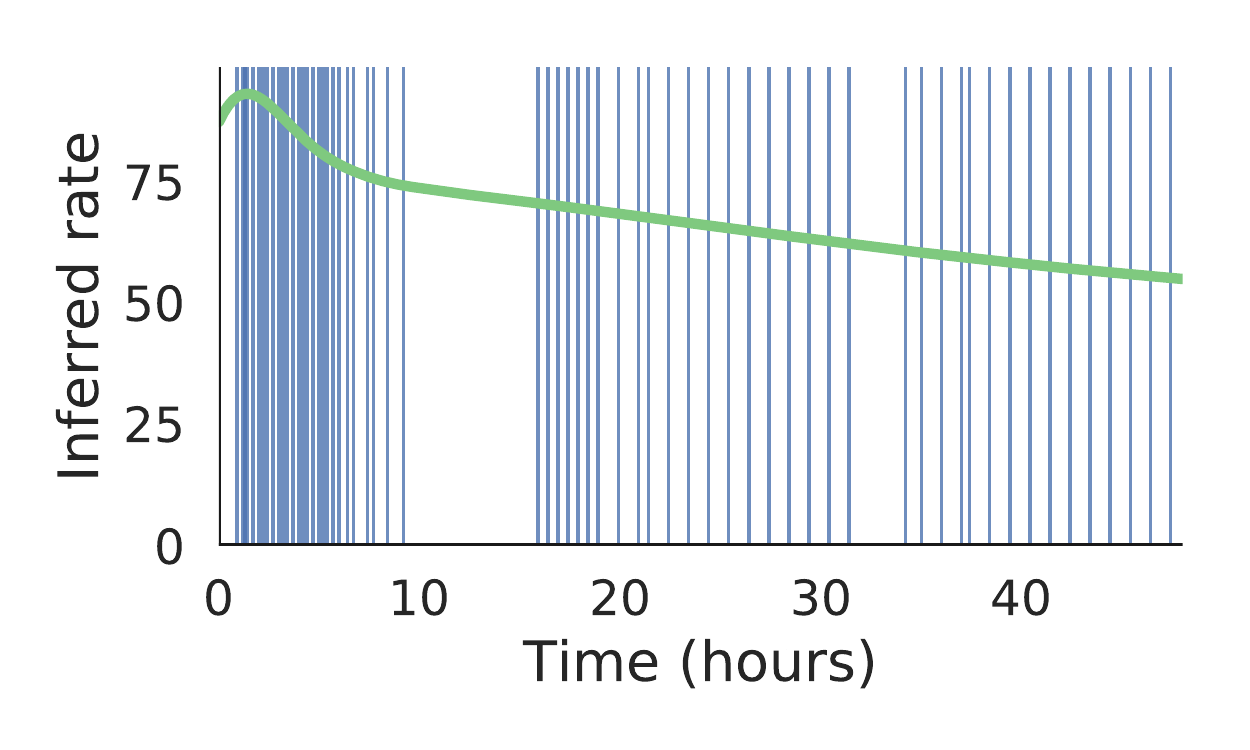}
    \end{subfigure}
    \caption{Examples of the inferred Poisson rate $\lambda(t)$ (green line) for two selected features of different patients from the Physionet dataset. Vertical lines mark observation times. HR: heart rate; PaO2: partial pressure of arterial O2; NIDiasABP: noninvasive diastolic arterial blood pressure; MAP: invasive mean arterial blood pressure}
    \label{fig:poisson}
\end{figure*}

\begin{table}
    \small
	\caption{Mean squared error on the toy dataset}
	\label{tab:1dsim}
	\centering
	\begin{tabular}{cl|cccc|cccc}
	    \centering
    	 & & \multicolumn{4}{c}{ Interpolation } & \multicolumn{4}{c}{ Extrapolation } \\
        & \% observed points & 10 & 20 & 30 & 50 & 10 & 20 & 30 & 50 \\
        \hline
        \multirow{5}{*}{\rotatebox{90}{Autoreg}} & RNN $\Delta t$ & 0.06081 & 0.04680 & 0.05822 & 0.04116 & 0.06172 & 0.06115 & 0.06891 & 0.05617 \\
        & RNN-imputed & 0.08558 & 0.06043 & 0.03922 & 0.04116 & \textbf{0.06095} & 0.07212 & \textbf{0.06541} & \textbf{0.05049} \\
        & RNN-exp & 1.65891 & 0.05344 & 0.04974 & 0.03275 & 0.06172 & 0.06115 & 0.06891 & 0.05617 \\
        & RNN GRU-D & 2.35628 & 0.05997 & 0.04832 & 0.04116 & \textbf{0.06095} & 0.07212 & \textbf{0.06541} & \textbf{0.05049} \\
        & ODE-RNN & \textbf{0.05150} & \textbf{0.03211} & \textbf{0.02643} & \textbf{0.01666} & 0.06592 & \textbf{0.04774} & 0.10940 & 0.08000 \\
        \hline
        \multirow{3}{*}{\rotatebox{90}{VAE}} & RNN-VAE & 0.07352 & 0.07346 & 0.07323 & 0.07304 & 0.20107 & 0.03710 & 0.07281 & 0.02871 \\
        & Latent ODE (RNN enc) & \textbf{0.06860} & 0.06764 & \textbf{0.02754} & 0.05721 & \textbf{0.04920} & 0.04807 & \textbf{0.01788} & 0.02703 \\
        & Latent ODE (ODE enc) & 0.07133 & \textbf{0.03144} & 0.05354 & \textbf{0.01717} & 0.05313 & \textbf{0.04427} & 0.03572 & \textbf{0.01388} \\
		\hline
	\end{tabular}
\end{table}